\documentclass[lettersize,journal]{IEEEtran}
\usepackage{amsmath,amsfonts,amssymb}
\usepackage{algorithmic}
\usepackage{algorithm}
\usepackage{array}
\usepackage{textcomp}
\usepackage{stfloats}
\usepackage{url}
\usepackage{verbatim}
\usepackage{graphicx}
\usepackage{bm}
\usepackage{cite}
\usepackage{booktabs}
\usepackage{tablefootnote}
\usepackage{multirow}
\usepackage{colortbl}
\usepackage{diagbox}
\usepackage{threeparttable}
\usepackage{mathtools}
\usepackage{subfigure}
\usepackage{hyperref}
\hypersetup{hidelinks} 
\usepackage{color}

\newcommand \transpose {\mathsf{T}} 
\newcommand \wc {\mathcal{W}} 
\newcommand \rc {\mathcal{R}} 

\begin{document}

\title{Global Model Learning for Large Deformation Control of Elastic Deformable Linear Objects: An Efficient and Adaptive Approach}

\author{Mingrui Yu,~\IEEEmembership{Student Member,~IEEE},
Kangchen Lv, 
Hanzhong Zhong, 
Shiji Song,~\IEEEmembership{Senior Member,~IEEE}
and Xiang Li,~\IEEEmembership{Member,~IEEE}
\thanks{Mingrui Yu, Kangchen Lv, Hanzhong Zhong, Shiji Song and Xiang Li are with the Department of Automation, Tsinghua University, Beijing, China. Xiang Li is also with the CUHK Shenzhen Research Institute, Shenzhen, China. Emails: ymr20@mails.tsinghua.edu.cn; lkc21@mails.tsinghua.edu.cn; zhonghz18@mails.tsinghua.edu.cn;
shijis@mail.tsinghua.edu.cn;
xiangli@tsinghua.edu.cn. (\textit{Corresponding author: Xiang Li}).}
\thanks{This work was supported in part by the Science and Technology Innovation 2030-Key Project under Grant 2021ZD0201404, in part by the National Natural Science Foundation of China under Grant U21A20517 and 52075290, in part by the Science and Technology Innovation Council of Shenzhen under Grant JCYJ20170816164748290, and in part by the Institute for Guo Qiang, Tsinghua University.}
}

\markboth{IEEE Transactions on Robotics}{Mingrui Yu \MakeLowercase{\textit{et al.}}: A Sample Article Using IEEEtran.cls for IEEE Journals}

\IEEEpubid{0000--0000/00\$00.00~\copyright~2022 IEEE}

\maketitle

\begin{abstract}
Robotic manipulation of deformable linear objects (DLOs) has broad application prospects in many fields. However, a key issue is to obtain the exact deformation models (i.e., how robot motion affects DLO deformation), which are hard to theoretically calculate and vary among different DLOs. Thus, shape control of DLOs is challenging, especially for large deformation control which requires global and more accurate models. In this paper, we propose a coupled offline and online data-driven method for efficiently learning a global deformation model, allowing for both accurate modeling through offline learning and further updating for new DLOs via online adaptation. Specifically, the model approximated by a neural network is first trained offline on random data, then seamlessly migrated to the online phase, and further updated online during actual manipulation. Several strategies are introduced to improve the model's efficiency and generalization ability. We propose a convex-optimization-based controller and analyze the system's stability using the Lyapunov method. Detailed simulations and real-world experiments demonstrate that our method can efficiently and precisely estimate the deformation model, and achieve large deformation control of untrained DLOs in 2D and 3D dual-arm manipulation tasks better than the existing methods. It accomplishes all 24 tasks with different desired shapes on different DLOs in the real world, using only simulation data for the offline learning.
\end{abstract}

\begin{IEEEkeywords}
Deformable linear objects (DLOs), shape control, robotic manipulation, model learning
\end{IEEEkeywords}

\section{Introduction}
\IEEEPARstart{D}{eformable}
linear objects (DLOs) refer to deformable objects in one dimension, such as ropes, elastic rods, wires, cables, etc. The demand for manipulating DLOs is reflected in many applications, and a significant amount of research efforts have been devoted to the robotic solutions to these applications \cite{jose2018robotic, yin2021modeling, nadon2018multi,herguedas2019survey}. For example, wires are manipulated to assemble devices in 3C manufacturing \cite{8460694}; belts are manipulated in assemblies of belt drive units \cite{jin2021trajectory}; and in surgery, sutures are manipulated to hold tissue together \cite{cao2020sewing}. 

\IEEEpubidadjcol

The manipulation tasks of DLOs can be divided into two categories \cite{rita2021reform}. In the first category, the goals are not about the exact shapes of DLOs; rather, they concern high-level conditions such as tangling or untangling knots \cite{wakamatsu2006knotting,saha2007manipulation}, obstacle-avoidance \cite{mcconachie2020manipulating,mitrano2021learning}, following and insertion \cite{she2021cable},  etc. The second category concerns manipulating DLOs to desired shapes, where a key challenge is to estimate the unknown deformation models (i.e., how the robot motion affects the DLO shapes) with sufficient accuracy  \cite{zhu2021challenges}. This work focuses on the shape control tasks. 

\begin{figure}[tb]
    \centering
    \includegraphics[width=8.7cm]{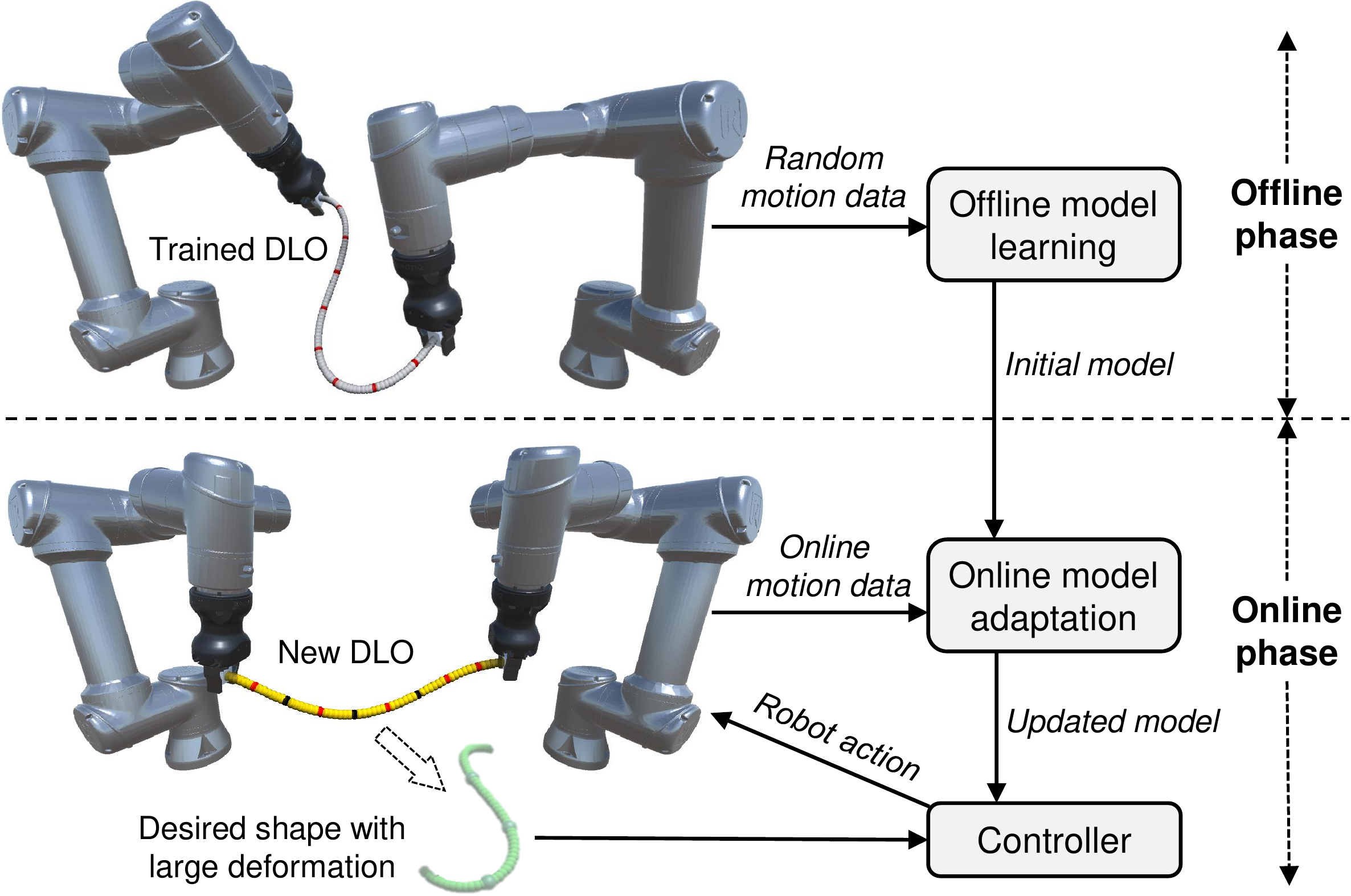}
    \caption{Overview of the proposed scheme for DLO shape control. In the offline phase, an initial estimation of the deformation model is learned based on randomly collected data. Then, in the online phase, the shape control task is executed, while the model migrated from the offline phase is further updated to compensate for offline modeling errors and adapt to new DLOs.}
    \label{fig:overview}
\end{figure}

It is very challenging to obtain the exact deformation models of DLOs. First, they are hard to calculate theoretically. Some analytical modeling methods can be used to model DLOs, such as mass-spring systems, position-based dynamics, and finite element methods \cite{yin2021modeling, arriola2020modeling}. However, all are approximate models and require accurate parameters of DLOs which are difficult to acquire in the real world. As a result, data-driven approaches have been widely applied to learn the deformation models without studying the complex physical dynamics.
Second, the deformation model is nonlinear with respect to the DLO configuration, making it usually data-inefficient to learn an accurate global model effective for any DLO shape. Note that such a global model is essential for large deformation control of DLOs, where the initial and final shapes can be very different.
Third, the deformation models may vary significantly among different DLOs, owing to different lengths, thicknesses, and materials. 
It is impractical to pay a huge time cost to collect new data of the new DLO before every manipulation task. Thus, the adaptiveness to different DLOs must be considered.

The existing data-driven methods to obtain the deformation model can be mainly divided into purely offline and purely online methods. 
For the purely offline methods, the most common one is to first learn a forward kinematics model (FKM) offline, and then use the model predictive control (MPC) in manipulations \cite{yan2020learning, wenbo2021deformable,yan_self-supervised_2020,yang2021learning,lee2022sample}. Reinforcement learning methods have also been studied \cite{lin2020softgym,rita2021learning}. Although they can achieve large deformation control of a well-trained DLO, they usually require a large amount of data for model learning, and may have trouble manipulating a different untrained DLO.
Apart from these offline approaches, some studies have used purely online methods to estimate a local linear deformation model (Jacobian matrix) of the manipulated DLO \cite{david2013modelfree,navarro2016Automatic,zhu_dual-arm_2018,jin2019robust,lagneau_automatic_2020,zhu2021vision}.
Since they are executed purely online, they can be applied to any new DLO. However, because only a small amount of local data can be utilized, these estimated models are less accurate and only effective in local configurations, making them only able to handle tasks with local and small deformation. 
While both offline learning and online learning have advantages, finding a solution to utilize them to complement each other effectively is not trivial.

In this paper, we consider the problem of large deformation control of elastic DLOs, where the initial configurations of DLOs are far from the desired shapes. 
The ``large" here is relative to the existing works.
To achieve it, we propose a coupled offline learning and online adaptation method for efficiently learning the global deformation model, as illustrated in Fig. \ref{fig:overview}.
This complementary scheme allows more accurate modeling through offline learning and further updating for new DLOs via online adaptation.
Specifically, we use a radial-basis-function neural network (RBFN) to globally model the nonlinear mapping from the current state to the current local linear deformation model (a locally effective Jacobian matrix). 
The RBFN is first offline trained on collected random data, then migrated to the online phase as an initial estimation, and finally further updated to adapt to the manipulated DLO during actual manipulation.
Hence, both the advantages of offline and online learning are well explored and seamlessly incorporated.
In addition, we introduce several strategies to improve the model's generalization ability and training efficiency, such as scale normalization and rotation data augmentation.
We also propose a convex-optimization-based feedback control law, which considers the singularity of the Jacobian and constrains the robots not to overstretch the DLO.
The stability of the closed-loop system and the convergence of task errors are analyzed using the Lyapunov method. 
Exhaustive simulation and real-world experiments are carried out to validate the proposed method. 
The video and code are available at the project website\footnote{\url{https://mingrui-yu.github.io/shape_control_DLO_2}}.

Our key contributions are highlighted as follows:
\begin{enumerate}
    \item We prove that the deformation model of DLOs can be globally described by a nonlinear mapping from the DLO configuration to a local Jacobian matrix in quasi-static manipulations. Such models can be learned more data-efficiently than the forward kinematics models.
    \item We propose a coupled offline and online method to efficiently learn the global deformation model, which firstly achieves both stable large deformation control and effective adaptation to new DLOs.
    \item We conduct detailed simulation and real-world experiments to demonstrate the outperformance of the proposed method over the existing works.
\end{enumerate}

This work is an extension of our previous work presented in conference paper \cite{yu2022shape} which proposed a preliminary coupled offline and online model learning method for DLO shape control. The improvements include: 
1) proposing new model modifications and training strategies to improve the model's generalization ability and data efficiency; 
2) proposing a new controller based on convex optimization, which considers the singularity of the Jacobian and constrains the robots not to overstretch the DLO; 
3) proposing a new robust online model updating law with detailed rigorous stability analysis; 
4) conducting more detailed simulation studies, and real-world 2D and 3D dual-arm manipulation experiments.

\section{Related Work}

\subsection{DLO Shape Control Tasks}
The shape control tasks of DLOs can be further divided into two types. The first type concerns manipulating soft ropes placed on tables, where the ropes are so soft that their deformation shapes can be held by friction of tables without being grasped by robots (i.e., such deformation can be considered as plastic deformation). Therefore, the robots can move the DLO to the desired shape by executing a series of pick-and-place actions at any point on the DLO \cite{yan2020learning, wenbo2021deformable,yan_self-supervised_2020,yang2021learning,lin2020softgym,lee2022sample}.
The second type is about manipulating stiffer DLOs such as flexible rods and cables, in which their deformation under forces from robots is mainly elastic deformation 
\cite{david2013modelfree,navarro2016Automatic,zhu_dual-arm_2018,jin2019robust,lagneau_automatic_2020,zhu2021vision,rita2021learning}.
The robots grasp only the ends of DLOs to control the internal shapes, so the deformation model is essential. 
Moreover, the robot degree of freedoms (DoFs) can be up to 12 in 3D dual-arm manipulations, making modeling and control more challenging. 
This work focuses on the second type of shape control tasks: manipulating elastic DLOs. While the existing works usually consider local and small deformation, this work deals with much larger deformation. 

\subsection{Existing Approaches}
\subsubsection{With Analytical Models}
Analytical modeling of DLOs has been researched over the past several decades \cite{yin2021modeling}. Some works on shape control were based on analytical models. In \cite{bretl2014quasi,roussel2015manipulation}, the static equilibrium configurations of elastic rods were analyzed using a geometric model, and the simulated manipulation was based on planning a proper path through the set of equilibrium shapes. In \cite{lv2022dynamic}, an energy-based elastic rod model was utilized for dynamic simulation of DLOs, and a heuristic model-free controller was proposed for DLO shaping.
Finite element model (FEM) simulation of DLOs was used for open-loop shape control in \cite{duenser2018interactive}, and a reduced FEM was used for closed-loop shape control in \cite{koessler2021anefficient}. 
The application of these methods is limited because the models usually require a large amount of computation and accurate DLO parameters (such as the cross-section area, Young modulus, shear modulus, etc.), which are hard and cumbersome to obtain in the real world.

\subsubsection{With Demonstrations or Reinforcement Learning}
Recently, data-driven approaches have been applied to shape control of DLOs, dispensing with analytical modeling. In \cite{rambow2012autonomous,nair2017combining,tang2018aframework}, the shaping of DLOs was addressed by learning from human demonstrations. Robots could reproduce human actions for specific tasks. 
Reinforcement learning (RL) has also been applied to learn control policies in an end-to-end manner.
A simulated benchmark of RL algorithms for deformable object manipulation was presented in \cite{lin2020softgym}, in which rope straightening and shaping tasks were studied. RL policies for shape control of elastoplastic DLOs in simulation were learned in \cite{rita2021learning}, and a simulation sandbox for DLO manipulation was introduced in \cite{rita2021reform}. 
RL-based methods for DLO manipulation are in the early stages of research. Like other RL applications, these methods suffer from considerably high training expenses and challenging transfer from simulation to real-world scenarios. As a result, up to now, RL-based methods have been primarily studied in simulation only and are difficult to apply in the real world.

\subsubsection{With Offline-learned Forward Kinematics Models}
Different from the end-to-end methods, many works first learn neural-network-based forward kinematics models (FKM) of DLOs offline, and then use the MPC to control the shape. A forward kinematics model predicts the shape at the next time step based on the current shape and input action. In \cite{lee2022sample}, an FKM in the image space was learned, and random-sampling-based planning was applied for control. In \cite{yan2020learning, wenbo2021deformable}, an encoder from the image space to the latent space, and an FKM in the latent space, were jointly trained. A more robust and data-efficient approach is to estimate the DLO state first and then learn the FKM in the physical state space \cite{yan_self-supervised_2020,yang2021learning}. 
Owing to the nonlinear DLO kinematics and complex neural network architectures, the training of these models often requires tens of hours of data. Moreover, their generalization to different untrained DLOs cannot be guaranteed.

\subsubsection{With Online-estimated Local Models}
To control the shape of unknown objects, a series of methods tackle the shape control problem based on purely online estimation of local linear deformation models of DLOs, in which a small change of the DLO is linearly related to a small movement of the manipulator by a locally effective estimated Jacobian matrix. The control input is directly calculated using the inverse of the Jacobian. The estimated Jacobian matrix was updated online using the Broyden update rule \cite{david2013modelfree}, the gradient descent method \cite{navarro2016Automatic}, or the (weighted) least square estimation on recent data in the current sliding window \cite{zhu_dual-arm_2018,jin2019robust,lagneau_automatic_2020,zhu2021vision}.
Compared with the offline models, these online estimated models are only effective in local configurations, and less accurate because only limited local data are utilized. Thus, they mostly handle tasks with local and small deformation. In addition, their estimated local models cannot be used for multi-step predictions or even reused for new tasks. 

\subsubsection{With Offline + Online Model Learning}
To leverage both the advantages of the offline and online learning, we proposed a preliminary coupled offline and online model learning method in our previous work \cite{yu2022shape}. Later, Wang et al. proposed another scheme to combine the two phases \cite{wang2022offline}. They first trained a nonlinear FKM in the offline phase, and then used another local Jacobian model to compensate for the residual error of the FKM in the online phase. In contrast, in our method, by reformulating the deformation model as a global Jacobian model, the shortcomings of the FKM and local Jacobian model are avoided, and the offline learning and online adaptation can be executed on the same model with seamless migration. Moreover, \cite{wang2022offline} was dealing with local deformation in 2D scenarios, while our method is validated on 3D large deformation control tasks.

\section{Preliminaries}

\subsection{Problem Formulation}
\begin{figure}[tb]
    \centering
    \includegraphics[width=8.7cm]{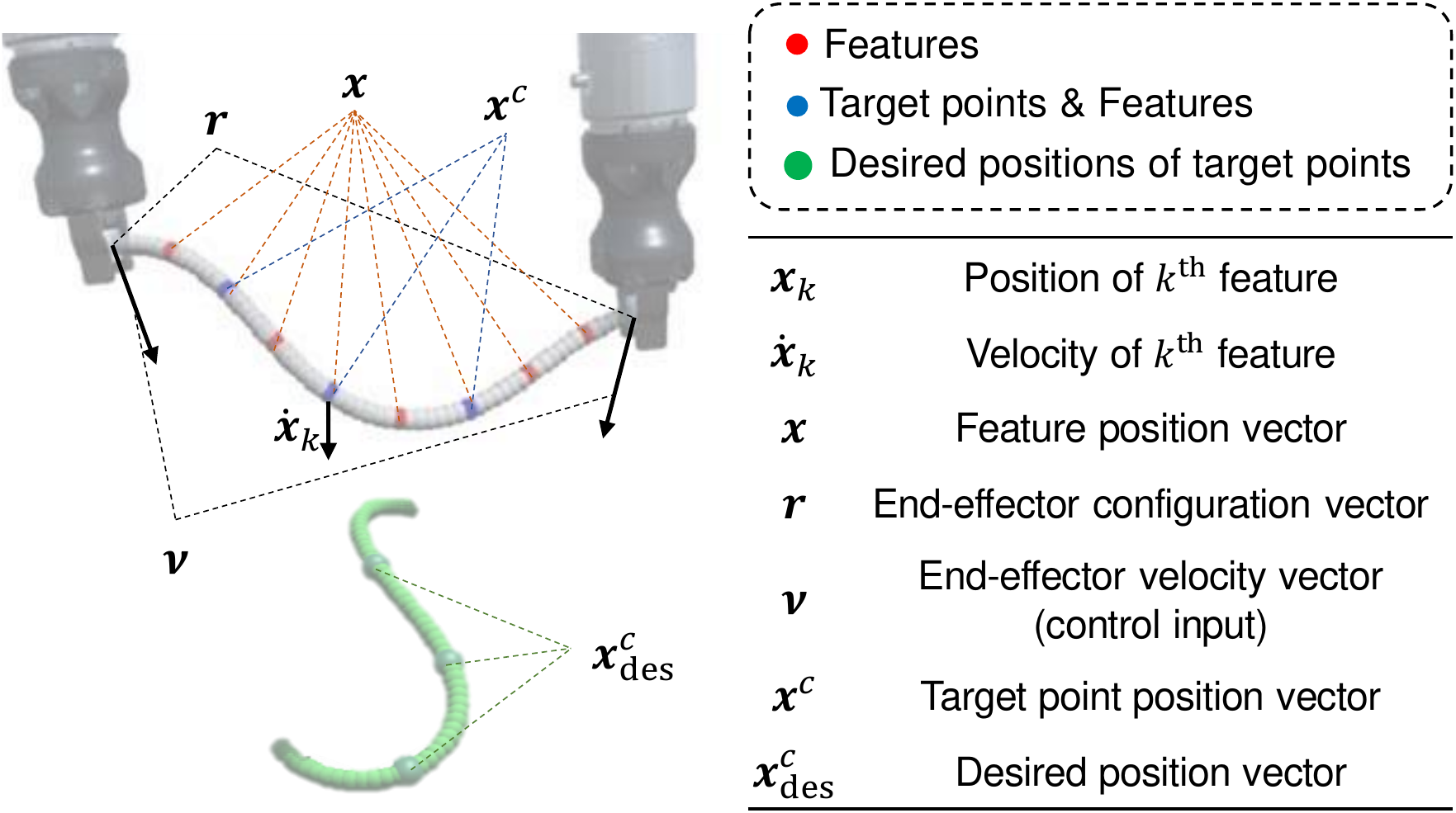}
    \caption{Problem formulation of the considered DLO shape control task and the notations for the main variables. The robots grasp the ends of the DLO. The DLO shape is represented by the positions of multiple features. Some of the features are chosen as target points, and the task is defined as controlling robots to manipulate the target points to their desired positions.}
    \label{fig:variables}
\end{figure}

This paper considers the quasi-static shape control of elastic DLOs. As illustrated in Fig. \ref{fig:variables}, the robots grasp the ends of the DLO and manipulate it to the desired shape. The overall shape of the DLO is represented by the positions of multiple \textit{features} uniformly distributed along the DLO. The \textit{target points} are chosen from the features, and the task is defined as moving the target points on the DLO to their corresponding \textit{desired positions} by controlling the velocities of the end-effectors. The specific choice of the target points depends on the task needs.

\noindent \textbf{Assumptions}: 
The following assumptions are made for the DLO manipulation:
\begin{enumerate}
    \item Only elastic deformation of the DLO will happen during manipulation.
    
    \item The manipulation process is quasi-static, meaning the shape of the DLO is determined by only its potential energy and no inertial effects during manipulation \cite{navarro2016Automatic}.
    
    \item The stiffness matrix of the DLO is positive and full-rank around the equilibrium point \cite{navarro2016Automatic}.
    
    \item The ends of the DLO have been rigidly grasped by the robot end-effectors. The velocities of the end-effectors can be kinematically controlled.
\end{enumerate}
These assumptions are commonly used in the research of deformable object manipulation.

For generalization and simplicity of writing, we formulate the problem as a 3D dual-arm manipulation task in the following text. Note that our method can also be applied to other specific settings, such as 2D or single-arm manipulations.

\subsection{Notations}
Some frequently used notations are described as follows:
\begin{itemize}
    \item $\bm I$: the identity matrix.
    \item {\spaceskip=0.2em\relax $[\bm a; \bm b]$: the vertical concatenation of column vector $\bm a$ and $\bm b$.}
    \item $\bm a \cdot \bm b$: the dot product of vector $\bm a$ and $\bm b$.
    \item $\|\bm a\|_2$: the Euclidean norm of vector $\bm a$.
    \item $x(t)$: the value of time-varying variable $x$ at time $t$. Time $(t)$ is omitted in most of the text, except in cases where variable values at different times appear in the same equation. 
\end{itemize}

The notations for the main variables are shown in Fig. \ref{fig:variables}. The position of the $k^{\rm th}$ feature is represented as $\bm x_k \in \Re^3$.  The overall shape vector of the DLO is represented as $\bm{x} := [\bm{x}_1; \cdots; \bm{x}_m] \in \Re^{3m}$ , where $m$ is the number of the features.
The configuration vector of the end-effectors (also the grasped ends) is denoted as
    $\bm r := \left[ \bm p_1 ; \bm q_1; \bm p_2; \bm q_2 \right] \in \Re^{14}$
, where $\bm p_1, \bm p_2 \in \Re^3$ is the position of the left and right end respectively; $\bm q_1, \bm q_2 \in \Re^4$ is the orientation of the left and right end respectively, which are represented by quaternions.
For simplicity of writing, we use $\bm s := [\bm{x}; \bm r]$ to represent the state containing both the DLO shape and the configuration of the end-effectors.
We denote the velocity vector of the end-effectors as
    $\bm \nu :=  \left[ \bm v_1 ; \bm \omega_1; \bm v_2; \bm \omega_2 \right]
    \in \Re^{12}$
, where $\bm v_1, \bm v_2 \in \Re^3$ is the linear velocity of the left and right end respectively; $\bm \omega_1, \bm \omega_2 \in \Re^3$ is the angular velocity of the left and right end respectively. The velocity vector of the robot end-effectors $\bm \nu$ is the control input.

\subsection{Mathematical Expressions of the Existing Methods}
We provide the mathematical expressions of the existing methods to facilitate the readers' understanding.

The forward kinematics model (FKM) in the offline methods is described as
\begin{equation}
    \bm x(t+\delta t) = \bm f\left( \bm x(t), \bm r(t), \bm\nu(t)  \right)
\end{equation}
where $\delta t$ is the step interval. It learns a nonlinear mapping $\bm f(\cdot)$ from the current configuration and input action to the next DLO shape.

The Jacobian model in the online methods is described as
\begin{equation} \label{equation:jacobian0}
    \dot{\bm x} = \bm J \, \bm\nu
\end{equation}
where the velocity vector of the DLO features can be locally linearly related to the velocity vector of the end-effectors using a Jacobian matrix. Note that the Jacobian matrix is varying during moving and needs to be constantly updated.
Most of the previous works derive such models by assuming the DLO shape can be determined by the configuration of the end-effectors using a function like $\bm x = \bm h(\bm r)$, and differentiating it with respect to time \cite{berenson2013manipulation,david2013modelfree,zhu2021vision}. However, in a global sense, the DLO shape cannot be uniquely determined by the configuration of the end-effectors \cite{bretl2014quasi}. 
In the next section, we derive the Jacobian-based model in another way, using the quasi-static assumption.

\section{Deformation Model}

\subsection{Global Jacobian-based Deformation Model}

One key problem of DLO shape control is studying the mapping from the motion of the end-effectors to the motion of the DLO features, which is essential for model-based control. We also use the Jacobian to describe the local relationship, but in a global way. That is, we discover that the Jacobian matrix can be fully determined by the current state, so our model learns the mapping from the state $(\bm x, \bm r)$ to the corresponding Jacobian matrix, which is specified as
\begin{equation} \label{equation:jacobian1}
    \dot{\bm x} = \bm J(\bm x, \bm r) \, \bm\nu
\end{equation}
Note that here $\bm J(\cdot)$ is a globally effective function.

\noindent \textbf{Theorem 1}: Under the quasi-static assumption, the velocity vector of the features on the elastic DLO can be related to the velocity vector of the end-effectors as (\ref{equation:jacobian1}).

\noindent \textbf{Proof}: Denote the potential energy of the elastic DLO as $E$, which is assumed to be fully determined by $\bm x$ and $\bm r$. 
In the quasi-static assumption, internal equilibrium holds at any state during the manipulation, where the DLO's internal shape $\bm x$ locally minimizes the potential energy $E$ \cite{bretl2014quasi}. That is, $\partial E / \partial \bm x = \bm 0$ at any state.
Consider the DLO is moved from state $(\bar{\bm x}, \bar{\bm r})$ to  state $(\bar{\bm x} + \delta \bm x, \bar{\bm r} + \delta \bm r)$ where $\delta \bm x$ and $\delta \bm r$ are small movements of the features and grasped ends.
Denote
\begin{equation}
    \bm g(\bm x, \bm r) = \frac{\partial E}{\partial \bm x}, 
    \bm A(\bm x, \bm r) = \frac{\partial^2 E}{\partial\bm x \partial\bm x},
    \bm B(\bm x, \bm r) = \frac{\partial^2 E}{\partial\bm x \partial\bm r}
\end{equation}
Using the Taylor expansion and neglecting the higher-order terms, we have
\begin{equation}
    \bm g(\bar{\bm x} + \delta \bm x, \bar{\bm r} + \delta \bm r) \approx \bm g(\bar{\bm x}, \bar{\bm r}) + \bm A(\bar{\bm x}, \bar{\bm r}) \delta {\bm x} + \bm B(\bar{\bm x}, \bar{\bm r}) \delta {\bm r}
\end{equation}
With the assumption, $\bm g(\bm x, \bm r) \equiv \bm 0$, so $\bm g(\bar{\bm x} + \delta \bm x, \bar{\bm r} + \delta \bm r) = \bm g(\bar{\bm x}, \bar{\bm r}) = \bm 0$. In addition, $\bm A$ and $\bm B$ physically represent the stiffness matrices; assuming the DLO has a positive and full-rank stiffness matrix around the equilibrium point, matrix $\bm A$ is invertible \cite{navarro2016Automatic}. Then, it can be obtained that
\begin{equation}
    \delta {\bm x} \approx -  \left(\bm A(\bar{\bm x}, \bar{\bm r})\right)^{-1} \bm B(\bar{\bm x}, \bar{\bm r}) \delta {\bm r}
\end{equation}
Note that this equation holds for arbitrary state $(\bar{\bm x}, \bar{\bm r})$. In addition, in slow manipulations, $\dot{\bm x} \approx \delta \bm x / \delta t$ and $\dot{\bm r} \approx \delta \bm r / \delta t$ with small $\delta t$. Thus, we have
\begin{equation}
    \dot{\bm x} \approx -  \left(\bm A(\bm x, \bm r)\right)^{-1} \bm B(\bm x, \bm r) \, \dot{\bm r}
\end{equation}
where $\dot{\bm r}$ is the derivative of the configuration of the grasped ends with respect to time, which can be related to the velocity vector $\bm \nu$ by a matrix as:
\begin{equation}
\begin{aligned}
    \dot{\bm r} = 
    \left[ \begin{array}{c}
         \dot{\bm p_1} \\
         \dot{\bm q_1} \\
         \dot{\bm p_2} \\
         \dot{\bm q_2}
    \end{array} \right]
    &=
    \underbrace{
    \left[ \begin{array}{cccc}
         \bm I_{3\times3} & & & \\
         & \bm M(\bm q_1) & & \\
         & & \bm I_{3\times3} & \\
         & & & \bm M(\bm q_2)
    \end{array} \right]
    }_{\bm C(\bm r)}
    \left[ \begin{array}{c}
         \bm v_1 \\
         \bm \omega_1 \\
         \bm v_2 \\
         \bm \omega_2
    \end{array} \right]
\end{aligned}
\end{equation}
where $\bm M$ is the matrix relating the derivative of the quaternion to the angular-velocity vector, which is determined by the quaternion.
Then, denoting $-(\bm A(\bm x, \bm r))^{-1} \bm B(\bm x, \bm r) \bm C(\bm r)$ as $\bm J(\bm x, \bm r)$, we derive (\ref{equation:jacobian1}) and prove Theorem 1.

\,

Note that (\ref{equation:jacobian1}) can be rewritten as
\begin{equation} \label{equation:jaco2}
    \dot{\bm x} = \left[ \begin{array}{c}
      \dot{\bm{x}}_1   \\ \vdots \\ \dot{\bm{x}}_m \end{array} \right]
    = \left[ \begin{array}{c}
      \bm{J}_1(\bm x, \bm r)   \\ \vdots \\ \bm{J}_m(\bm x, \bm r) \end{array} \right]
      \bm \nu
\end{equation}
where $\bm{J}_k(\bm x, \bm r)$ is the $(3(k-1)+1)^{\rm th}$ to $(3k)^{\rm th}$ rows of $\bm J(\bm x, \bm r)$. Thus, it can be obtained that
\begin{equation} \label{equation:jaco3}
    \dot{\bm x}_k = \bm J_k(\bm x, \bm r) \, \bm\nu, \quad k =  1,\cdots , m
\end{equation}
where $m$ is the number of the features. It indicates that different features correspond to different Jacobian functions. This formulation makes it convenient when choosing any subset of features as the target points in the manipulation tasks.

We emphasize that this Jacobian-based model is global because it is effective for any DLO state, which is essential for large deformation control. We then estimate it using a data-driven method based on a neural network (NN), where the input is the current state, and the output is the Jacobian matrix. 

\subsection{Model Modifications to Improve Generalization Ability} \label{section:model_modification}

We make the following modifications to improve the model's generalization ability on different DLOs in large deformation control tasks.

First, it can be noticed that the Jacobian is translation-invariant. That is, the translation of the DLO without changes of the internal shape will not alter the Jacobian matrix. Thus, for the input of the NN, we represent the position of each feature by its relative position to its left adjacent feature (or the left end for the leftmost feature). It seems that the Jacobian is also rotation-invariant, but it is only valid for rotations around the vertical axis because of gravity. We consider it by using a rotation data augmentation introduced in the next section.

\begin{figure} [tb]
  \centering 
  \subfigure[For translations of the ends]{ 
    \label{fig:scale_invariance_translation}
    \includegraphics[width=7.5cm]{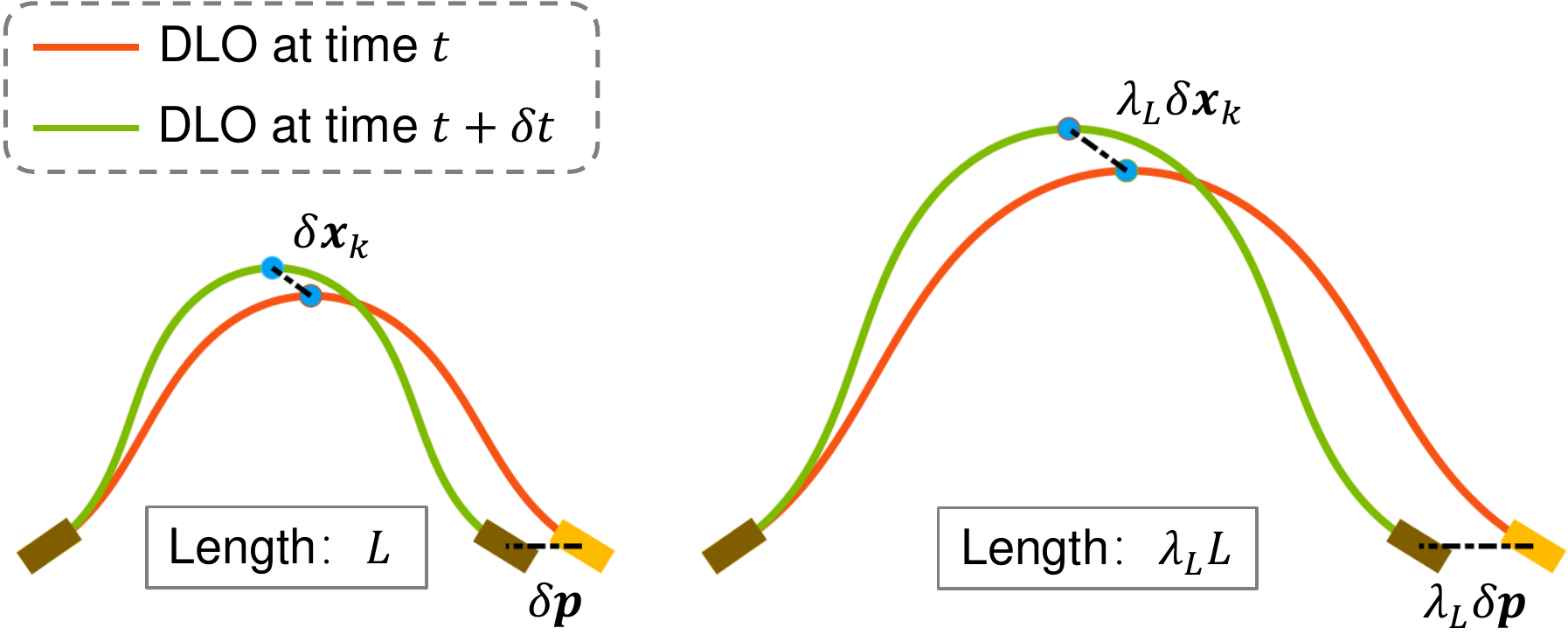} 
  } 
  \subfigure[For rotations of the ends]{ 
    \label{fig:scale_invariance_rotation}
    \includegraphics[width=7.5cm]{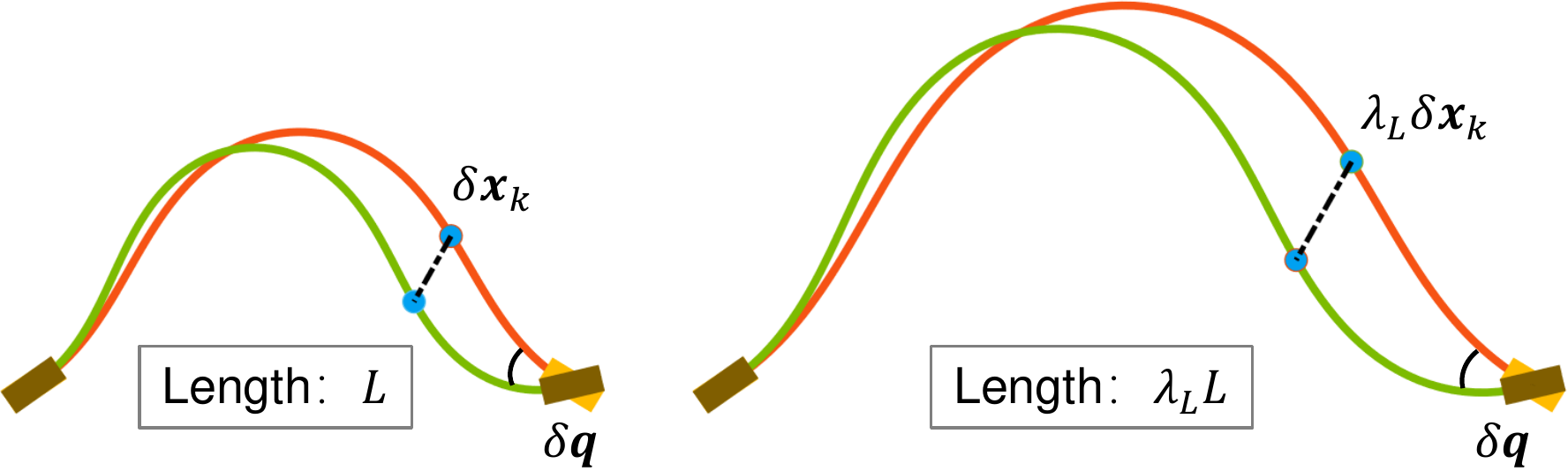} 
  }
  \caption{Illustration of the approximate scale-invariance. Two DLOs of different lengths are moved from one identical shape to another identical shape (ignoring the scale). In these ideal cases, the proportional relationship between the translation of the end and the movement of the feature is uncorrelated to the scale, while that between the rotation of the end and the movement of the feature is proportional to the scale.}
  \label{fig:scale_invariance}
\end{figure}

If the model trained on one DLO is applied on another DLO of a very different length, the changed value range of the state input will make the model (NN) almost completely fail. Considering the adaptiveness on different DLOs, we propose the \textit{scale normalization}. It is based on an approximation that there are similarities between the Jacobian matrices of DLOs with different lengths but similar shapes, which we call the \textit{approximate scale-invariance}. Fig. \ref{fig:scale_invariance} is an illustration of the ideal cases, where two DLOs are moved from one identical overall shape to another identical overall shape, and the long DLO is $\lambda_L$ times as long as the short DLO. First, consider the translation of the grasped ends, as shown in Fig. \ref{fig:scale_invariance_translation}. For the short DLO, the grasped end moves $\delta \bm p$, and the $k^{\rm th}$ feature moves $\delta \bm x_k$; for the long DLO, the grasped end moves $\lambda_L \delta \bm p$, and the feature moves $\lambda_L \delta \bm x_k$. Next, consider the rotation of the grasped ends, as shown in Fig. \ref{fig:scale_invariance_rotation}. For the short DLO, the grasped end rotates $\delta \bm q$, and the feature moves $\delta \bm x_k$; for the long DLO, the grasped end also rotates $\delta \bm q$, but the feature moves $\lambda_L \delta \bm x_k$. In these ideal cases, the proportional relationship between the translation of the end and the movement of the feature is independent of the scale, while that between the rotation of the end and the movement of the feature is proportional to the scale. Inspired by it, we define the approximate scale-invariance as follows: for DLOs of different lengths but similar overall shapes, the Jacobian matrices for the linear velocities of the ends are similar, and those for the angular velocities of the ends are approximately proportional to the lengths.

Considering these properties, we split the Jacobian into two parts: an unknown matrix uncorrelated to the scale and translation, and a constant scale matrix. The model (\ref{equation:jaco3}) is modified to
\begin{equation} \label{equation:jaco4}
    \dot{\bm x}_k 
    = \bm J_k(\bm s) \bm\nu
    = \left( \bm N_k(\tilde{\bm s}) \bm T \right) \bm\nu, 
    \quad k =  1,\cdots , m
\end{equation}
where $\bm s = [\bm{x}; \bm{r}]$, $\tilde{\bm s}$ is the relative state representation, and $\bm T$ is the scale matrix as 
\begin{equation}
    \bm T = \text{diag} \left[ \bm I_{3\times3}, L \bm I_{3\times3}, \bm I_{3\times3}, L \bm I_{3\times3}  \right]
\end{equation}
where $L$ is the length of the DLO. The relative state representation $\tilde{\bm s}$ is specifically defined as
\begin{equation} \label{equation:state_representation}
    \tilde{\bm s}
    := \left[ 
     \tilde{\bm x}_1 ; \cdots ; \tilde{\bm x}_m ; \,
     \tilde{\bm r}
    \right]
\end{equation}
where
\begin{equation}
\begin{aligned}
    \tilde{\bm x}_1 &= \frac{\bm x_1 - \bm p_1}{\|\bm x_1 - \bm p_1\|}, \tilde{\bm x}_k = \frac{\bm x_k - \bm x_{k-1}}{\|\bm x_k - \bm x_{k-1}\|}, k=2,\cdots,m
    \\
    \tilde{\bm r} &= \left[ \frac{\bm p_2 - \bm p_1}{\| \bm p_2 - \bm p_1 \|}; \bm q_1; \bm q_2 \right]
\end{aligned}
\end{equation}
where the scale is normalized by the normalization of the relative position vectors.

In (\ref{equation:jaco4}), only $\bm N_k(\tilde{\bm s})$ is unknown and will be approximated by an NN. The input of the NN is only related to the overall shape, ignoring the scale and translation. Therefore, it is much more data-efficient than using the absolute state representation $[\bm x; \bm r]$ which requires a larger NN and more training data to guarantee the generalization to different DLO lengths and large translations. 

\noindent \textbf{Remark 1}:  
The approximate scale-invariance is only an approximation and may cause modeling errors. In fact, DLOs of the same length may also have different properties because of different materials and thicknesses.
However, the approximation at least makes the offline-trained model able to work on DLOs of different lengths but not completely fail owing to the changed value range of the NN input. In the experiments, we demonstrate that this approximation is effective. Moreover, the remaining modeling errors can be compensated for by the online model adaptation on the specific DLO.

\section{Offline Model Learning}
Prior to the shape control tasks, an initial approximation of the model by an NN is learned based on offline-collected random motion data. This section introduces the NN model, data collection method, and training details.

\subsection{Neural Network Model}
We apply a radial-basis-function neural network (RBFN) to approximate the $\bm N_k(\tilde{\bm s})$ in the Jacobian model. The actual $\bm N_k(\tilde{\bm s})$ is represented as
\begin{equation} \label{vecJ=W*theta}
    {\rm vec} \left(\bm N_k(\tilde{\bm s})\right) = \bm W_k \bm{\theta}(\tilde{\bm s}), \quad k =  1,\cdots , m
\end{equation}
where $\bm W_k$ is the unknown actual RBFN weights for the $k^{\rm th}$ feature and $\bm{\theta}(\tilde{\bm s})=[\theta_1(\tilde{\bm s}), \theta_2(\tilde{\bm s}), \cdots, \theta_q(\tilde{\bm s})]^{\transpose} \in \Re^q$ is the vector of activation functions. We use the gaussian radial function as the activation function: 
\begin{equation} \label{singleNeuron}
\theta_i(\tilde{\bm s})={\rm e}^{\frac{-||\tilde{\bm s} -\bm \mu_i||_2^2}{\sigma_i^2}}, \quad i =  1,\cdots , q
\end{equation}
where $q$ is the number of the hidden neurons, and $\bm \mu_i$ and $\sigma_i$ are the center and width of the $i^{\rm th}$ hidden neuron.

Equation (\ref{vecJ=W*theta}) can be decomposed as
\begin{equation} \label{equation:J=Wtheta}
    \bm{N}_{ki}(\tilde{\bm s}) = \bm{W}_{ki} \bm{\theta}(\tilde{\bm s}), \quad i =  1,\cdots , 12
\end{equation}
where $\bm{N}_{ki}$ is the $i^{\rm th}$ column of $\bm N_k$, and $\bm{W}_{ki}$ is the ${(3(i-1)+1)}^{\rm th}$ to ${(3i)}^{\rm th}$ rows of $\bm{W}_k$. Subscribing (\ref{equation:J=Wtheta}) into (\ref{equation:jaco4}) yields
\begin{equation}
    \dot{\bm{x}}_k = \bm{N}_k(\tilde{\bm s}) \bm T \bm\nu 
    = \sum_{i=1}^{12}\bm{N}_{ki}(\tilde{\bm s}) T_i \nu_i 
    = \sum_{i=1}^{12}\bm{W}_{ki}\bm{\theta}(\tilde{\bm s}) T_i \nu_i
\end{equation}
where $\nu_i$ is the $i^{\rm th}$ element of $\bm \nu$ and $T_i$ is the $i^{\rm th}$ diagonal element of $\bm T$.

\begin{figure}[tb]
    \centering
    \includegraphics[width=8.7cm]{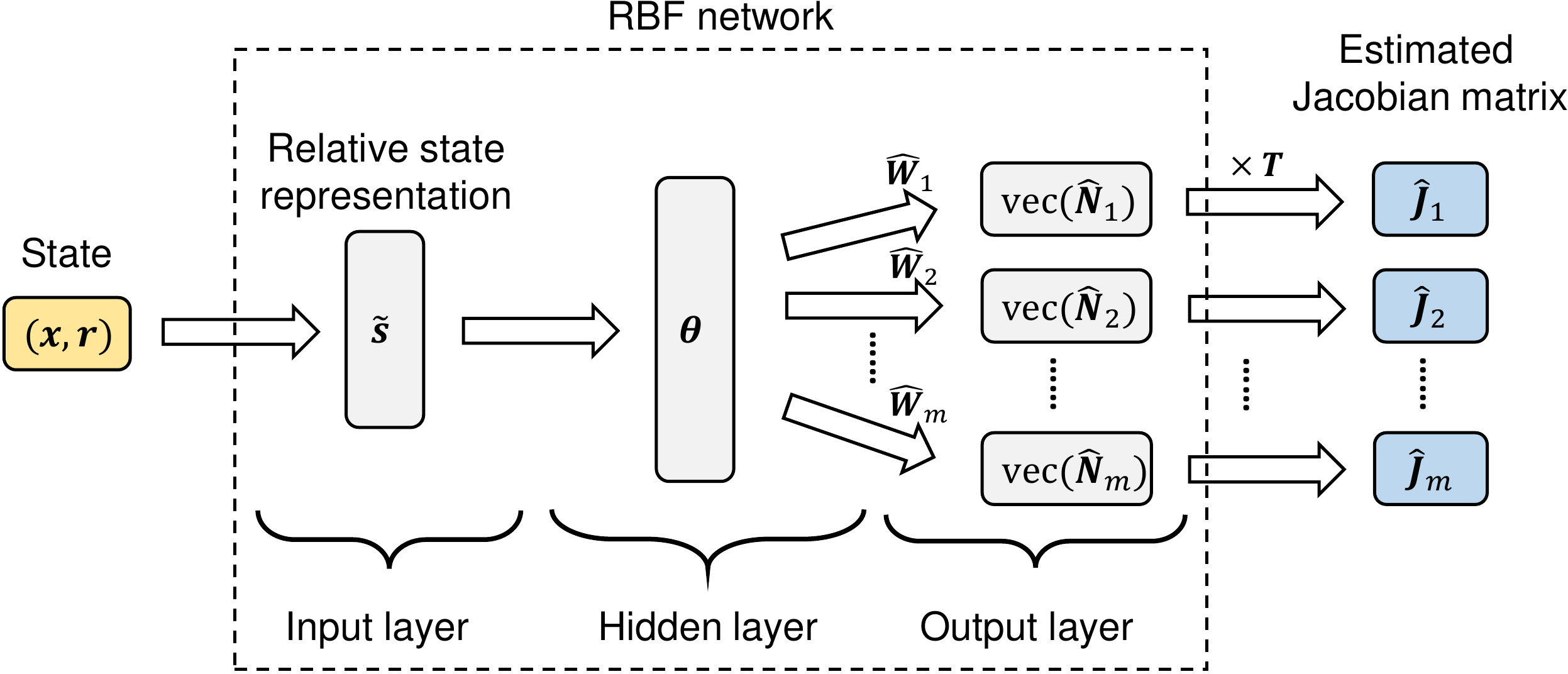}
    \caption{Architecture of our model to output the estimated Jacobian matrices for all DLO features.
    The RBFN takes the relative state representation (\ref{equation:state_representation}) as the input, and the product of its output and the scale matrix $\bm T$ is the estimated Jacobian matrix.}
    \label{fig:RBFN}
    \vspace{-2mm}
\end{figure}

As the actual Jacobian is unknown, the estimated Jacobian model is represented as
\begin{equation}
    \hat{\bm{J}}_k(\bm{s}) = \hat{\bm{N}}_k(\tilde{\bm s}) \bm T
\end{equation}
\begin{equation} \label{estimatedJ}
    {\rm{vec}} (\hat{\bm{N}}_k(\tilde{\bm s})) = \hat{\bm{W}}_k\bm{\theta}(\tilde{\bm s})
\end{equation}
where $\hat{\bm{W}}_k$ is the estimated RBFN weights. The architecture of our RBFN-based Jacobian model is shown in Fig. \ref{fig:RBFN}. Note that the learning or estimation for different features is carried out in parallel. 

Then, the prediction error of the estimated model for the $k^{\rm th}$ feature is obtained as
\begin{equation} \label{equation:ew}
\begin{aligned}
    \bm{e}_k 
    &= \dot{\bm{x}}_k - \hat{\bm{J}}_k(\bm s) \bm\nu
    = \dot{\bm{x}}_k - \hat{\bm{N}}_k(\tilde{\bm s}) \bm T \bm\nu
    \\
    &= \sum_{i=1}^{12}\bm{W}_{ki}\bm{\theta}(\tilde{\bm s}) T_i \nu_i - \sum_{i=1}^{12}\hat{\bm{W}}_{ki}\bm{\theta}(\tilde{\bm s}) T_i \nu_i
    \\
    &= \sum_{i=1}^{12}\Delta{\bm{W}}_{ki}\bm{\theta}(\tilde{\bm s}) T_i \nu_i
\end{aligned}
\end{equation}
where $\Delta{\bm{W}}_{ki}$ is the approximation error of the RBFN weights. 

It is known that the proper values of the centers and widths of the RBFN hidden neurons are important. To obtain good centers and widths, we use the k-means clustering on a subset of the training data to calculate initial values first, and then further update them during the offline NN training. 
Note that in the online phase, they are fixed, and only $\hat{\bm W}$ is updated.
Considering the noise and outliers in the data, we use the \textit{smooth L1 loss} \cite{girshick2015fast} of $\bm e_k$ for the offline training. 
All parameters are updated according to the mean loss on the dataset, using the \textit{Adam} optimizer \cite{kingma2014adam}.

We choose RBFN for its simple structure, robustness, and online learning ability \cite{yu2011advantages}. Though less expressive than some more complex NN architectures, it performs well enough in this work. 
The offline-trained RBFN can also be seamlessly migrated to the online adaptive control, to fully explore the advantages of both online and offline learning.

\subsection{Offline Data Collection} \label{section:offline_data_collection}
During the offline data collection, the DLO is moved randomly to collect data. Each data tuple contains the current positions and velocities of the features, and the current poses and velocities of the grasped ends, as $(\bm x, \dot{\bm x}, \bm r, \bm \nu)$.

\begin{figure} [tb]
  \centering 
  \subfigure[]{ 
    \includegraphics[width=2.82cm]{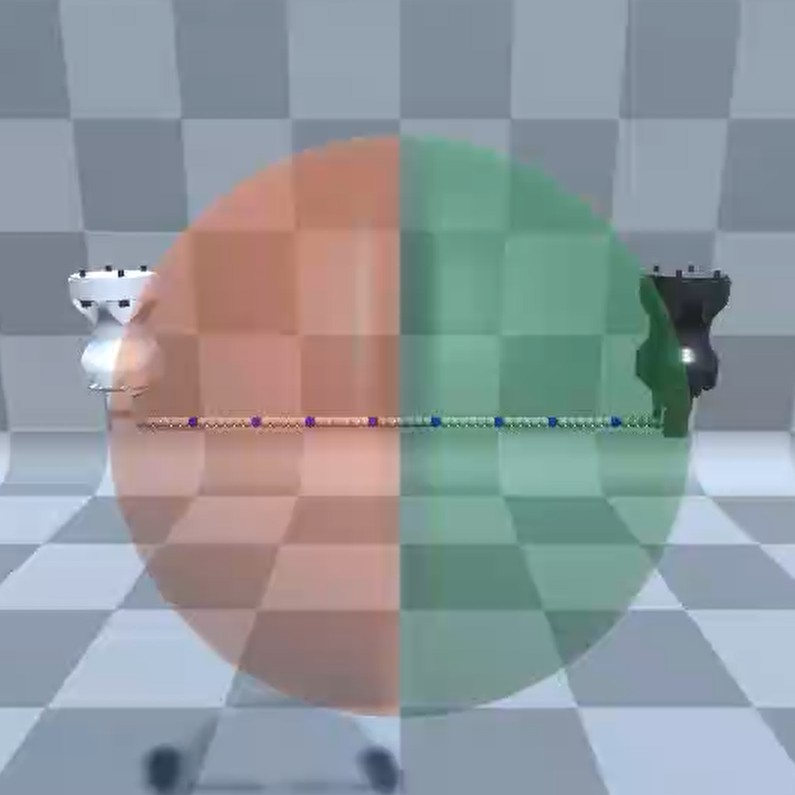} 
  } 
  \hspace{-0.4cm}
  \subfigure[]{ 
    \includegraphics[width=2.82cm]{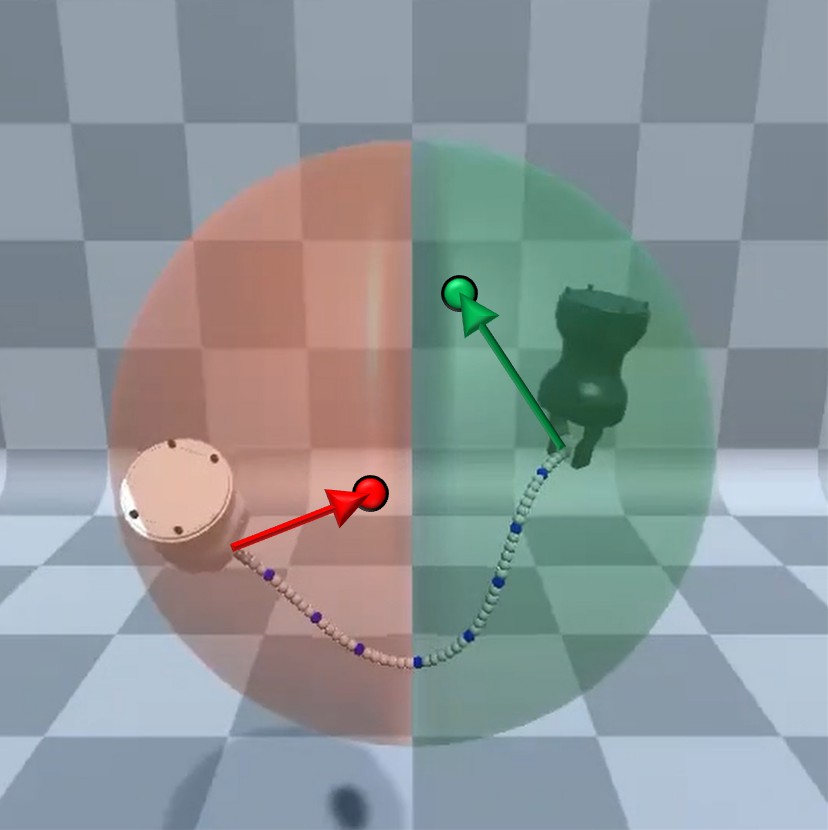} 
  }
  \hspace{-0.4cm}
  \subfigure[]{ 
    \includegraphics[width=2.82cm]{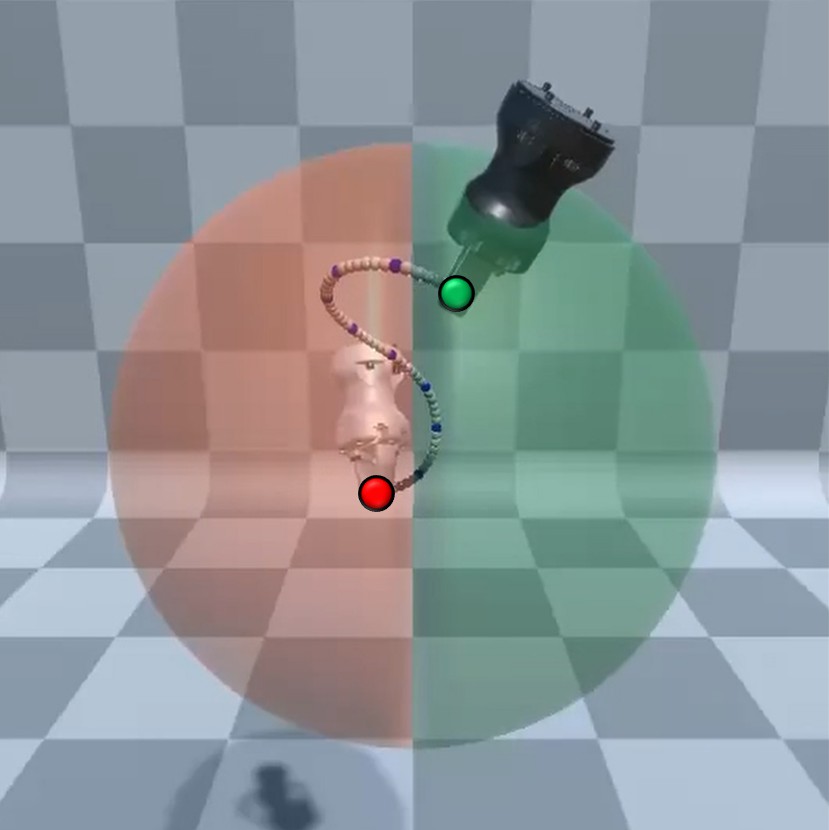} 
  }
  \caption{Illustration of the 3D offline data collection in the simulation. The red hemisphere is the workspace for the left end, and the green one is the workspace for the right end. (a) The initial straight DLO. (b) At the beginning of each time period, a destination is uniformly sampled for each end from its workspace. Then, each end is moved at a constant velocity towards its destination. (c) At the end of one time period, the ends reach the destinations.}
  \label{fig:data_collection_workspace}
    \vspace{-2mm}
\end{figure}

Our proposed data collection method has two advantages: 1) it covers as many different DLO shapes as possible; 2) it can be continuously executed in both simulation and the real world, avoiding tangling or overstretching. The method is shown in Fig. \ref{fig:data_collection_workspace}. We first define a DLO workspace as a sphere whose radius is half the DLO length. The workspace is further divided into two hemispheres by a vertical plane, in which the left one is for the left grasped end and the right one is for the right end. At the beginning of each time period, a destination for each end is uniformly sampled, whose position is in the corresponding workspace and orientation is in a preset range. During the period, each end is moved at a constant velocity to reach its destination at the end of the time period. Then, a new period starts, and new destinations are sampled.

Compared with completely random motion, our data collection method is more efficient to collect more valuable data, covering more possible shapes and avoiding tangling or overstretching. Moreover, it can be continuously executed with no need for resets, which allows the potential application in the real world.

\noindent
\textbf{Remark 2}:  
It would be better for the neural network if the offline collected data and online data have similar distribution. Thus, we separately collect 2D offline data for 2D control tasks and 3D offline data for 3D control tasks.

\subsection{Data Augmentation and Domain Randomization}

\subsubsection{Rotation Data Augmentation} \label{section:data_augmentation}
In large deformation control tasks, there may be large translations or rotations. Our model using the relative state representation is natively translation-invariant. To achieve rotation-invariance around the vertical axis, we introduce a \textit{rotation data augmentation}, which is inspired by observing the motion of the DLO and end-effectors in another coordinate that is defined by rotating the original world coordinate around the vertical axis.

It is implemented as follows. During the NN training, for each data tuple, a new coordinate $\rc$ is sampled by randomly rotating the original world coordinate $\wc$ around the vertical axis. The data are transformed from $\wc$ to $\rc$, and then sent to the NN for training.

After applying the rotation data augmentation, our model is both translation-invariant and vertical-rotation-invariant. As a result, our model can handle large translations and rotations, with no need to collect more data to guarantee generalization. It complements our data collection method which is convenient to implement but restricts the moving range of the DLO. We also find that it effectively reduces the over-fitting of the NN when the collected dataset is small, since infinite new data can be generated. This is why we do not directly consider it in the relative state representation.

\subsubsection{Domain Randomization on Different DLOs} 
To improve the generalization ability of the offline learned model on different DLOs, we apply a domain randomization method during the offline learning. That is, we train the offline model based on the combined data of several different DLOs with different lengths and thicknesses. This is for learning an offline model which is an acceptable initial estimation for different DLOs. Then, for any new DLO in the manipulation, this model can be efficiently updated via the online adaptation. 

Note that when using the domain randomization, the proposed scale normalization is still meaningful and effective, since it reveals the similarities between DLOs of different lengths and ensures that the ranges of the input values for the NN are consistent, which reduces the learning difficulty.

\section{Shape Control with Online Model Adaptation}

\subsection{General Control Problem Formulation}
The control objective is to move the target points on the DLO to the desired positions. The target points can be any subset of the features, whose indexes form set $\mathcal{C}$. Then, the shape vector, Jacobian matrix, and prediction error vector for the target points are denoted as
\begin{equation} \label{equation:all_c}
\begin{aligned}
    \bm{x}^c = \left[ \begin{array}{c}
      \vdots   \\ \bm x_k \\ \vdots \end{array} \right] , 
    \bm{J}^c({\bm s}) =  \left[ \begin{array}{c}
      \vdots   \\ \bm J_k({\bm s}) \\ \vdots \end{array} \right] ,
     \bm{e}^c = \left[ \begin{array}{c}
      \vdots   \\ \bm e_k \\ \vdots \end{array} \right]
    , \, k \in \mathcal{C}
\end{aligned}
\end{equation}
We denote the \textit{task error} as $\Delta \bm x^c = \bm x^c - \bm x^c_{\rm des}$, where $\bm x^c_{\rm des}$ is the desired position vector of the target points. 
The velocities of the end-effectors $\bm \nu$ are kinematically controlled.
 
Generally, the task can be formulated as an optimal control problem, in which the objective is to manipulate the DLO to the desired shape in a shorter time with smaller end-effector velocities while satisfying the valid-state constraints. However, since the system model is nonlinear and approximate and the constraints may be complex, exact solving of the optimal control problem is impossible.
An alternative approach is to apply model predictive control (MPC) instead. For such a nonlinear and complex problem, sampling-based MPC methods such as cross-entropy method and Model Predictive Path Integral (MPPI) \cite{williams2017information} are usually used.
To apply them, the estimated state equation needs to be approximately discretized as
\begin{equation} \label{equation:discretized_state_equation}
    \bm x(t+\delta t) = \bm x(t) + \hat{\bm J}(\bm s(t)) \, \bm \nu(t) \delta t
\end{equation}
where $\delta t$ is the time step interval. 
Then, the control problem is formulated as:
\begin{equation} \label{equation:MPC_formulation}
\begin{aligned} 
    \min_{\tiny \begin{array}{c} \bm \nu(t+i\delta t) \\0 \leq i < T_h \end{array} } 
    \, &  
    \mathcal{J}_M = \frac{1}{2} \| \Delta \bm x^c(t+T_h \delta t) \|_2^2
     + \frac{\lambda}{2} \sum_{i=0}^{T_h-1}  \| \bm \nu(t+i\delta t) \|_2^2
    \\
    \text{s.t.} \, &  \| \bm \nu(t+i\delta t) \|_2^2 \leq \nu_{\rm max}^2  , \quad 0 \leq i < T_h
    \\
    & \bm s(t+i\delta t) \in \mathcal{\bm S}_{\rm valid} , \quad 0 \leq i \leq T_h
\end{aligned}
\end{equation}
where $t$ is the current time, $T_h$ is the planning horizon, $\nu_{\rm max}$ is the maximum allowed end-effector speed, and $\mathcal{\bm S}_{\rm valid}$ is the set of valid states. This optimization problem is solved every step to update the control inputs. 

The conventional MPC commonly requires an accurate model, and may not be able to deal with huge model errors when applied on new DLOs. 
Moreover, the sampling-based MPC is usually computationally intensive because they require large amounts of sampling sequences and NN-based predictions for future states.
Therefore, we refer to the concept of MPC and propose an optimization-based adaptive controller, which can efficiently calculate the control input in the presence of an inaccurate model, handle various constraints, and further update the model online to adapt to new DLOs.

\subsection{Adaptive Controller through Online Model Adaptation} \label{section:adaptive_controller}

\begin{figure} [tb]
  \centering 
    \includegraphics[width=8.7cm]{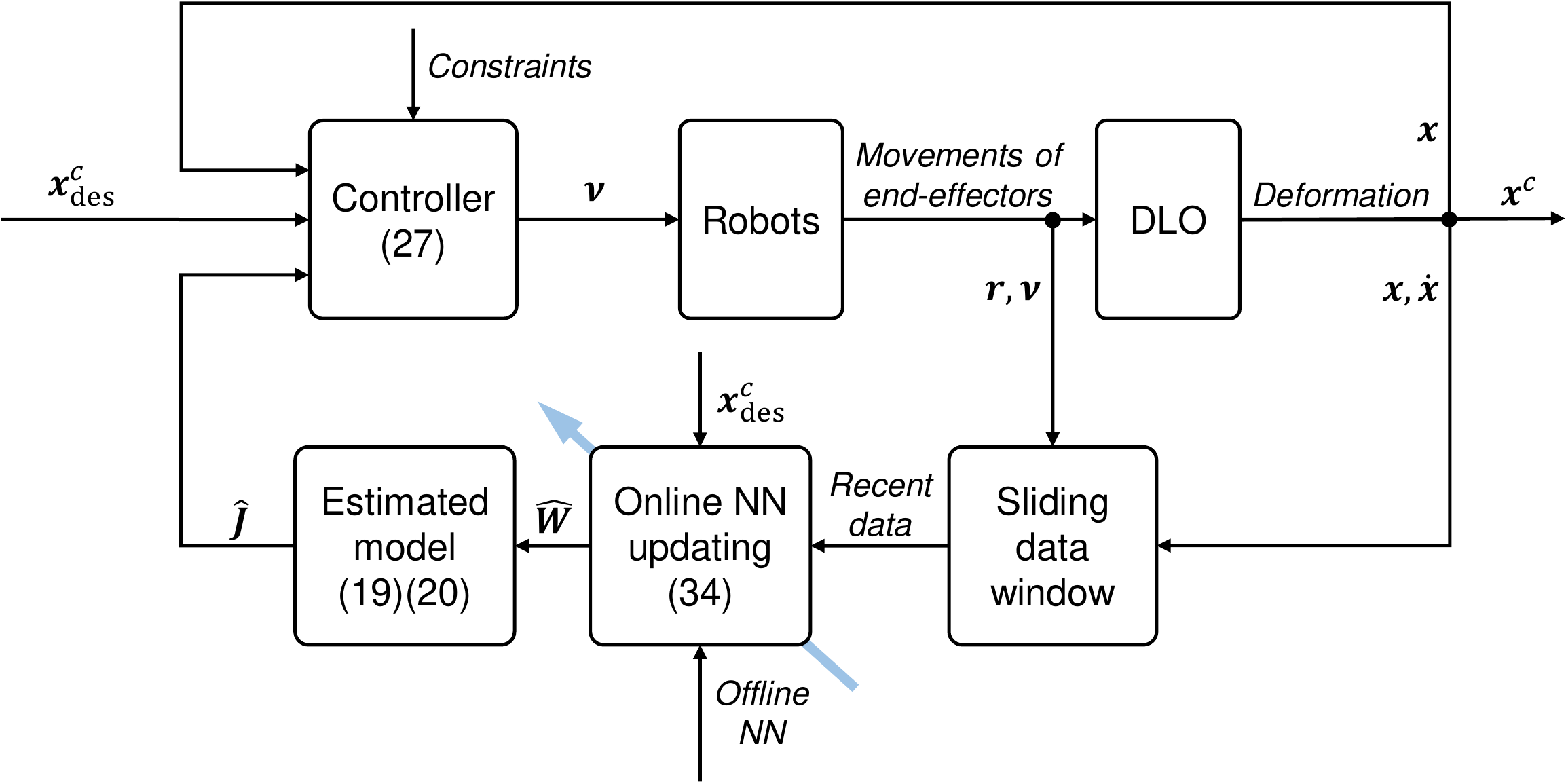} 
  \vspace{-2mm}
  \caption{Structure of the proposed control scheme (where the numbers in the brackets correspond to the equation labels).}
  \label{fig:control_block}
\end{figure}

We propose an adaptive controller to achieve the adaptivity to new DLOs through online model adaptation, whose structure is illustrated in Fig. \ref{fig:control_block}. During the manipulation, the offline-learned RBFN is further updated to compensate for the modeling errors.
The control input is efficiently calculated by solving a convex optimization problem which considers the singularity of the Jacobian and constrains the robots not to overstretch the DLO. The stability of the closed-loop system is theoretically guaranteed.

\,

\subsubsection{Control Law}
First, we define a \textit{saturated task error} as 
\begin{equation}
    \tilde{\Delta \bm x^c} 
    = 
    \left\{ \begin{array}{cl}
     \Delta \bm x^c  & , \|\Delta \bm x^c\|_2 \leq \epsilon_x \\ 
     \frac{\epsilon_x}{\|\Delta \bm x^c\|_2} \Delta \bm x^c  & , \rm{otherwise}
    \end{array}\right.
\end{equation}
where $\epsilon_x$ is the saturation threshold. Then, we define an \textit{ideal velocity vector} of the target points as 
\begin{equation} \label{equation:idea_vel}
    \dot{\bm x}^c_{\rm ide} = - \alpha \tilde{\Delta \bm x^c}
\end{equation}
where $\alpha$ is a positive gain factor. The ideal velocity vector of the target points is in the opposite direction of the task error $\Delta \bm x^c$. 
It is obvious that the target points will converge to their desired positions $\bm x_{\rm des}^c$ if they move at $\dot{\bm x}^c_{\rm ide}$. 
However, $\dot{\bm x}^c_{\rm ide}$ is unachievable in most cases, since the tight coupling between different target points makes them unable to move in arbitrary directions.
Thus, what we actually expect is to make the real velocities of the target points as close to the ideal velocities as possible. 
The benefit of the conversion from the task error to the ideal velocity vector as the control objective is that we can then formulate the controller as an optimization rather than just a feedback equation, in which essential constraints can be considered \cite{ruan2018accounting}.

The control input is specified as the optimal solution of the convex optimization problem:
\begin{equation} \label{equation:local_control_objective_estimate}
\begin{aligned}
    \min_{\bm \nu} 
    \quad &  
    \mathcal{J}_{A} =
    \frac{1}{2} \| \dot{\bm x}^c_{\rm ide} - \hat{\bm J}^c(\bm s) \bm \nu \|_2^2 + \frac{\lambda}{2} \| \bm \nu \|_2^2
    \\
    \text{s.t.} \quad &  \| \bm \nu \|_2^2 \leq \nu_{\rm max}^2
    \\
        & \bm C_{1}(\bm s) \bm \nu   \preceq \bm 0
    \\
        & \bm C_{2}(\bm s) \bm \nu = \bm 0
\end{aligned}
\end{equation}
where the linear constraints are for avoiding overstretching the DLO, which is introduced in the next section. We can also constrain the DoFs of the end-effectors for specific tasks.
Such a quadratically constrained quadratic program (QCQP) can be efficiently solved using convex optimization solvers.

Note that the form of the cost function in (\ref{equation:local_control_objective_estimate}) is similar to the damped least-squares method (DLS), which is widely used in robot inverse kinematics and control to address the singularity of the robot Jacobian \cite{chiaverini1994review}. Here we utilize it to deal with the singularity of the Jacobian of the DLO, which is very common during large deformation.
In addition, $\lambda$ is chosen as $\lambda = \lambda_0 \| \tilde{\Delta \bm x^c} \|_2$ where $\lambda_0$ is a positive constant. This means that when the current shape is far from the desired shape, a larger $\lambda$ is preferred for addressing the singularity problem; but when the current shape is near the desired shape, a smaller $\lambda$ is preferred for more precise control.

\,

\subsubsection{Constraints for Avoiding Overstretching}
In large deformation control tasks, it is important to ensure the DLO will not be overstretched during large motion in the presence of inaccurate deformation models.
To address this problem, we add linear constraints to the motion of the ends when the DLO is going to be overstretched, without using the estimated deformation model.

We define the near-overstretched states as all DLO features lying almost in a straight line, which is mathematically described as
\begin{equation}
    \frac{(\bm x_{k+1} - \bm x_{k}) \cdot (\bm x_{k} - \bm x_{k-1})}{\|\bm x_{k+1} - \bm x_{k}\|_2 \|\bm x_{k} - \bm x_{k-1}\|_2} > 1-\epsilon_s
    , \, \forall k = 2,\cdots,m-1
\end{equation}
where $\epsilon_s$ is a small threshold.
Denote the vector between the positions of the left and right ends as $\bm p_d = \bm p_2 - \bm p_1$. First, consider the linear velocities of the two ends. The constraint is that the linear velocity of the right end projected on $\bm p_d$ must be no more than that of the left end projected on $\bm p_d$:
\begin{equation}
    \bm p_d \cdot \bm v_2 - \bm p_d \cdot \bm v_1 \leq 0
\end{equation}
Then, consider the angular velocities of the two ends, whose effect on whether the DLO will be overstretched is much more difficult to model. Thus, we simply add a strong constraint that the angular velocities of the left and right ends equal to zero:
\begin{equation}
    \bm w_1 = \bm w_2 = \bm 0
\end{equation}
Therefore, the constraints are specified as follows:
\begin{equation}
    \underbrace{ \left[ \begin{array}{cccc}
            - \bm p_d^{\transpose} \quad  & \bm 0_{1 \times 3} \quad &  \bm p_d^{\transpose} \quad & \bm 0_{1 \times 3}   \\
    \end{array}  \right] }_{\bm C_l}
    \bm \nu
     \leq  0
\end{equation}
\begin{equation}
    \underbrace{ \left[ \begin{array}{cccc}
            \bm 0_{3\times 3} \quad &  \bm I_{3\times 3}  \quad & \bm 0_{3\times 3} \quad & \bm 0_{3\times 3} \\
            \bm 0_{3\times 3} \quad &  \bm 0_{3\times 3}  \quad & \bm 0_{3\times 3} \quad & \bm I_{3\times 3} \\
    \end{array}  \right] }_{\bm C_a}
    \bm \nu
    = \bm 0
\end{equation}

Considering both normal and near-overstretched states, in (\ref{equation:local_control_objective_estimate}) we set
$\bm C_1(\bm s) \hspace{-0.5mm} = \hspace{-0.5mm}\bm 0, \bm C_2(\bm s) = \bm 0$ for normal states; and $\bm C_1(\bm s) = \bm C_l, \bm C_2(\bm s) = \bm C_a$ for near-overstretched states.

\,

\subsubsection{Online Model Adaptation} \label{section:online_learning}
Modeling errors may exist because of insufficient offline training or different properties between the manipulated DLO and the trained DLOs. 
Thus, we further update the model migrated from the offline phase while carrying out the shape control task.

We maintain a sliding window of length $T_w$ to store the recent motion data whose timestamps form set $\mathcal{T}_w$:
\begin{equation}
    \mathcal{T}_w = 
    \left\{ \tau | \tau = t - i \delta t, 0 \leq i < T_w
    \right\}
\end{equation}
where $t$ is the current time.

The online updating law of the $j^{\rm th}$ row of $\hat{\bm W}_{ki}$ of the RBFN is specified as
\begin{equation} \label{equation:online_updating_law}
\begin{aligned}
     \dot{\hat{\bm W}}_{kij}^{\transpose}(t) = &  
    \, \eta 
    [ 
    \bm \theta(\tilde{\bm s}(t))  T_i \nu_i(t)  \tilde{\Delta x_{kj}}(t)
    \\
    & + \gamma \frac{1}{T_w} \sum_{\tau \in \mathcal{T}_w} \bm \theta(\tilde{\bm s}(\tau)) T_i \frac{\nu_i(\tau)}{n_v(\tau)} \frac{e_{kj}(\tau, t)}{n_v(\tau)} ]
\end{aligned}
\end{equation}
where $\tilde{\Delta x_{kj}}$ is the $j^{\rm th}$ element of the saturated task error $\tilde{\Delta \bm{x}_k}$ and $e_{kj}$ is the $j^{\rm th}$ element of the prediction error $\bm e_k$. Note that $\bm e_k(\tau, t)$ represents the prediction error of the stored data at time $\tau$ using the updated estimated Jacobian model at time $t$. In addition, $\eta$ is the positive online learning rate, $\gamma$ is a positive weight coefficient, and $n_v(\tau)$ is a normalization factor specified as:
\begin{equation}
    n_v(\tau) = 
    \left\{ \begin{array}{cl}
     \| \dot{\bm x}(\tau) \|_2  & , \| \dot{\bm x}(\tau) \|_2 \geq \epsilon_v \\ 
     \epsilon_v  & , \text{otherwise}
    \end{array}\right.
\end{equation}
where $\epsilon_v$ is a positive threshold to avoid amplifying the noise when the velocities are very small. Such updating is done for all $k \in \mathcal{C}$, $i = 1, \cdots, 12$, and $j = 1, \cdots, 3$.

The proposed online updating law (\ref{equation:online_updating_law}) has several advantages: 
1) considering all recent data in the sliding window instead of only the latest data can reduce the effect of sensing noise; 
2) the updating is driven by both the task error and prediction error, which enables faster and more stable updating; 
3) the combination of the model updating law and control law theoretically guarantees the stability of the closed-loop system; 4) the online updating starts from the pre-trained results obtained in the offline learning stage, so that the advantages of the learning in both phases are fully explored and combined.

\,

\noindent\textbf{Theorem 2}: When the adaptive control scheme described by (\ref{equation:local_control_objective_estimate}) and (\ref{equation:online_updating_law}) is applied to the robot system for shape control of DLOs, the closed-loop system is stable and the task error $\Delta \bm x^c$ is bounded in the presence of modeling errors. Furthermore, the task error will converge to zero as $t \rightarrow \infty$, unless the prediction errors $\bm e^c(\tau, t)$ for all data in the sliding window are zero as well as the optimal solution of (\ref{equation:local_control_objective_estimate}) is zero at a configuration on the path.

\noindent\textbf{Proof}: See Appendix \ref{appendix:stablity_analysis}.

\,

\noindent
\textbf{Remark 3}: The proposed control method assumes the desired positions are achievable. Thus, the desired shapes are set as pre-recorded shapes of the manipulated DLO in our experiments. In future works, we will study the determination of achievable desired shapes for specific tasks from the perspective of planning.

\section{Simulation Results}

\subsection{Evaluation Metrics}
The evaluation metrics used in the simulation and also the real-world experiments are introduced as follows.

\subsubsection{Deformation Magnitude} \label{subsubsection:deformation_magnitude}
We divide the generalized deformation of a DLO between time $t_1$ and $t_2$ into two parts: \textit{translation} and \textit{relative deformation}. The \textit{translation} refers to the translation of the centroid of the DLO (approximated by the average position of all features), which is specified as
\begin{equation}
    D_{\rm t}(\bm x(t_1), \bm x(t_2)) 
    = \| \bar{\bm x}(t_1) - \bar{\bm x}(t_2)   \|_2
\end{equation}
where $\bar{\bm x} = \frac{1}{m} \sum_{k=1}^{m}{\bm x_k}$. Then, the \textit{relative deformation} is defined as the average of the movement of each feature relative to the centroid, which is specified as
\begin{equation} \label{equation:relative_deformation}
\begin{aligned}
    D_{\rm rd}(\bm x(t_1), & \bm x(t_2))   =
    \\
    \frac{1}{m} \sum_{k=1}^m  & \| (\bm x_k(t_1)  - \bar{\bm x}(t_1)) 
    - (\bm x_k(t_2) - \bar{\bm x}(t_2)) \|_2
\end{aligned}
\end{equation}
Note that the \textit{relative deformation} describes changes of the overall shape while ignoring translations.

\subsubsection{Metrics for Modeling Accuracy} 
The shape prediction error is defined as
\begin{equation}
    e_{\rm shape} = \| \bm x_{\rm pred} - \bm x_{\rm groundtruth}   \|_2
\end{equation}
The relative prediction error of the feature velocity vector using our Jacobian model is defined as
\begin{equation} \label{equation:relative_vel_pred_error}
    e_{\rm vel} = \frac{\| \dot{\bm{x}} - \hat{\bm{J}}(\bm{s}) \bm\nu \|_2 }{\| \dot{\bm{x}} \|_2} \times 100\%
\end{equation}

\subsubsection{Metrics for Shape Control} \label{subsubsection:metrics_shape_control}
Criteria for evaluating shape control performance: 
(1) final task error: the final Euclidean distance between the desired position vector and final position vector within 30 seconds:
\begin{equation}
    e_{\rm control} = \|  \bm x^c(t_f)  -  \bm x^c_{\rm des} \|_2, \quad t_f = 30{\rm s}
\end{equation}
(2) average task error of all cases: the average final task error over all cases;
(3) success rate: if the final task error is less than 5cm, this case is regarded successful; 
(4) average task error of successful cases: the average final task error over all successful cases; 
(5) average task time: the average time used to achieve success over all successful cases. The task time is for reference only, since it depends on the control gain in servo methods or the bound of control inputs in MPC.

Note that the above $e_{\rm shape}$, $e_{\rm vel}$, and $e_{\rm control}$ contain the errors of all features / target points without averaging.

\begin{figure*} [tb]
  \centering 
    \includegraphics[width=\textwidth]{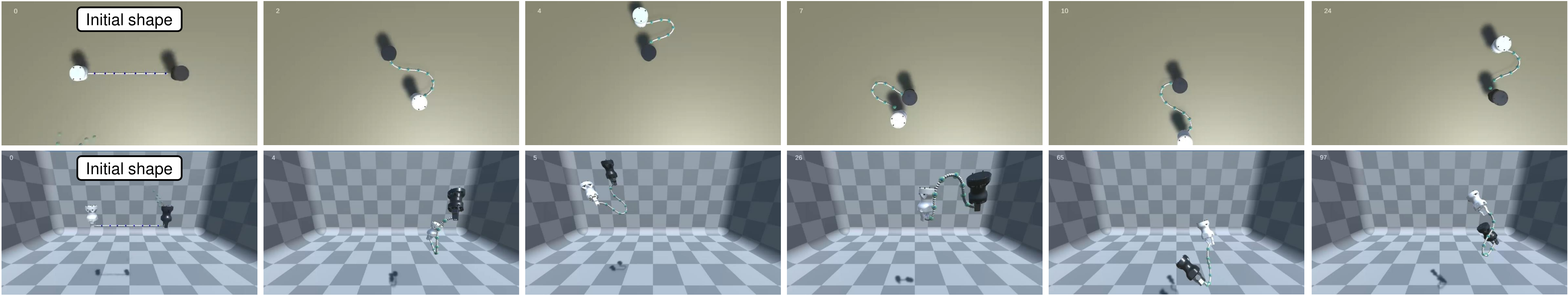} 
  \caption{The simulation environment and some of the shape control tasks accomplished using our method. The top row is from the 2D tasks, and the bottom row is from the 3D tasks. The blue points represent the target points (also the features) along the DLO. The translucent green points represent the desired positions of the target points. In each case, the DLO starts from a straight line, as shown in the leftmost figure of each scenario.}
  \label{fig:sim_control_snapshots}
\end{figure*}

\subsection{Simulation Setup}
The simulation environment is shown in Fig. \ref{fig:sim_control_snapshots}. The simulation of DLOs is based on Obi \cite{obi}, a unified particle physics engine for deformable objects in Unity3D \cite{unity}. Unity ML-Agents Toolkit \cite{juliani2018unity} is used for the communication between Unity and Python scripts. The two ends of the DLO are grasped by two grippers, which can translate and rotate.
The DLO shape is represented by 8 features ($m\hspace{-1mm}=\hspace{-1mm}8$). The data collection frequency, control frequency, and online learning frequency are 10Hz, 10Hz, and 50Hz, respectively ($\delta t = 0.1$). 

The hyper-parameters for the controller are set as: $\alpha = 1.0$, $\lambda_0 = 0.1$, $\epsilon_x = 0.2$, and $\epsilon_s = 0.002$; those for the online model adaptation are set as: $T_w = 20$, $\gamma=10$, $\epsilon_v=0.01$, and $\eta=1.0$.

\subsection{Offline Learning of the Deformation Model}
The offline data of DLOs are collected in simulation, using the method introduced in Section \ref{section:offline_data_collection}. An RBFN with 256 hidden neurons ($q=256$) is first trained offline to learn the initial model. We perform a series of quantitative comparative studies to validate the proposed Jacobian model and offline training methods.

\,
\subsubsection{Our Jacobian Model v.s. Forward Kinematics Model} \label{subsubsection:jaco_fkm_compare}

\begin{figure} [tb]
  \centering 
  \subfigure[2D]{ 
    \includegraphics[width=4.3cm]{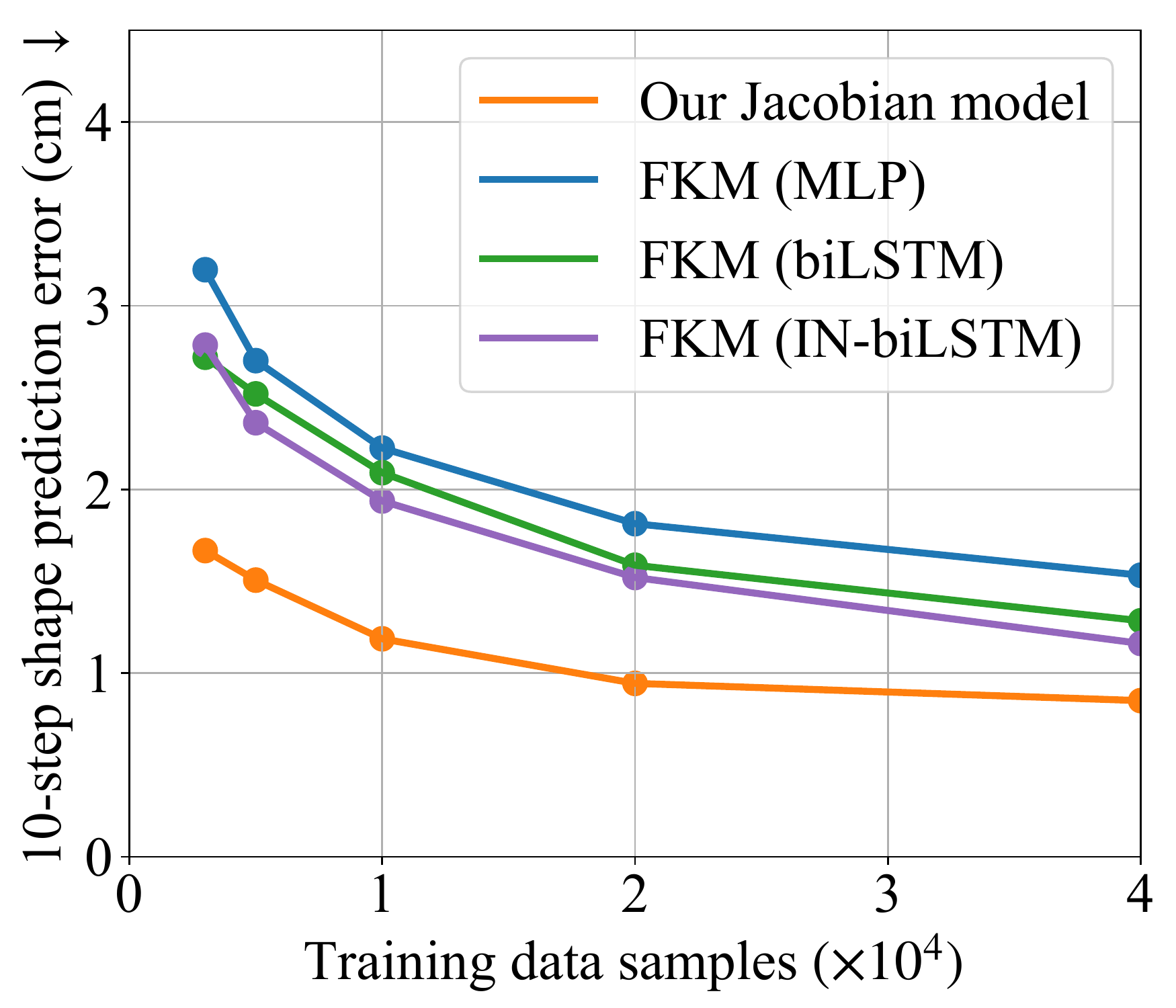} 
  } 
  \hspace{-0.5cm}
  \subfigure[3D]{ 
    \includegraphics[width=4.3cm]{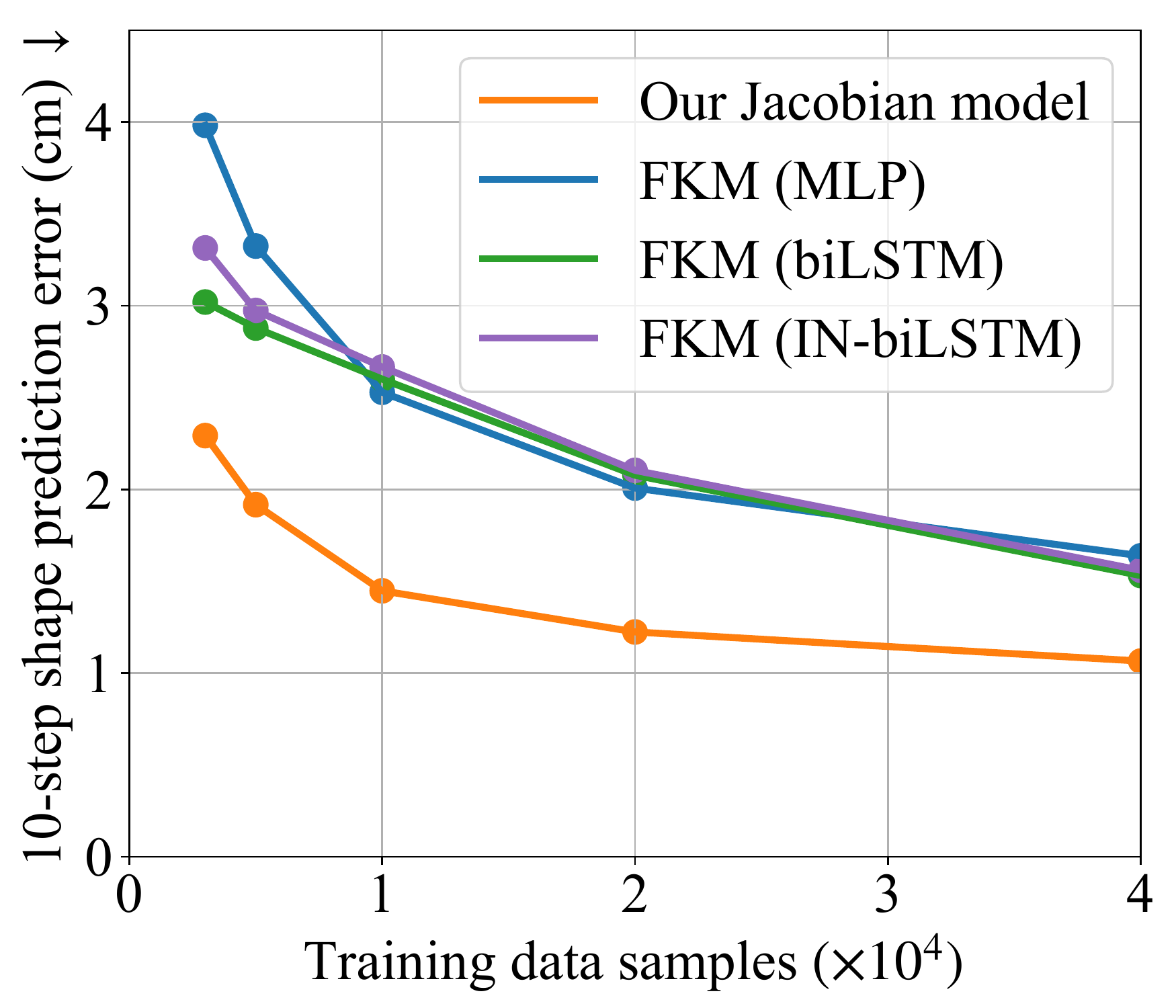} 
  }
  \caption{Comparison between our Jacobian model and the forward kinematics models (FKMs), and the relationship between the offline modeling accuracy and the amount of training data. All training data and test data are from the same DLO. The error is the average Euclidean distance between the prediction and ground truth of the shape after 10 steps.}
  \label{fig:exp_jaco_fkm_compare}
\end{figure}

\begin{figure*} [tb]
  \centering 
  \subfigure{ 
    \includegraphics[width=2.8cm]{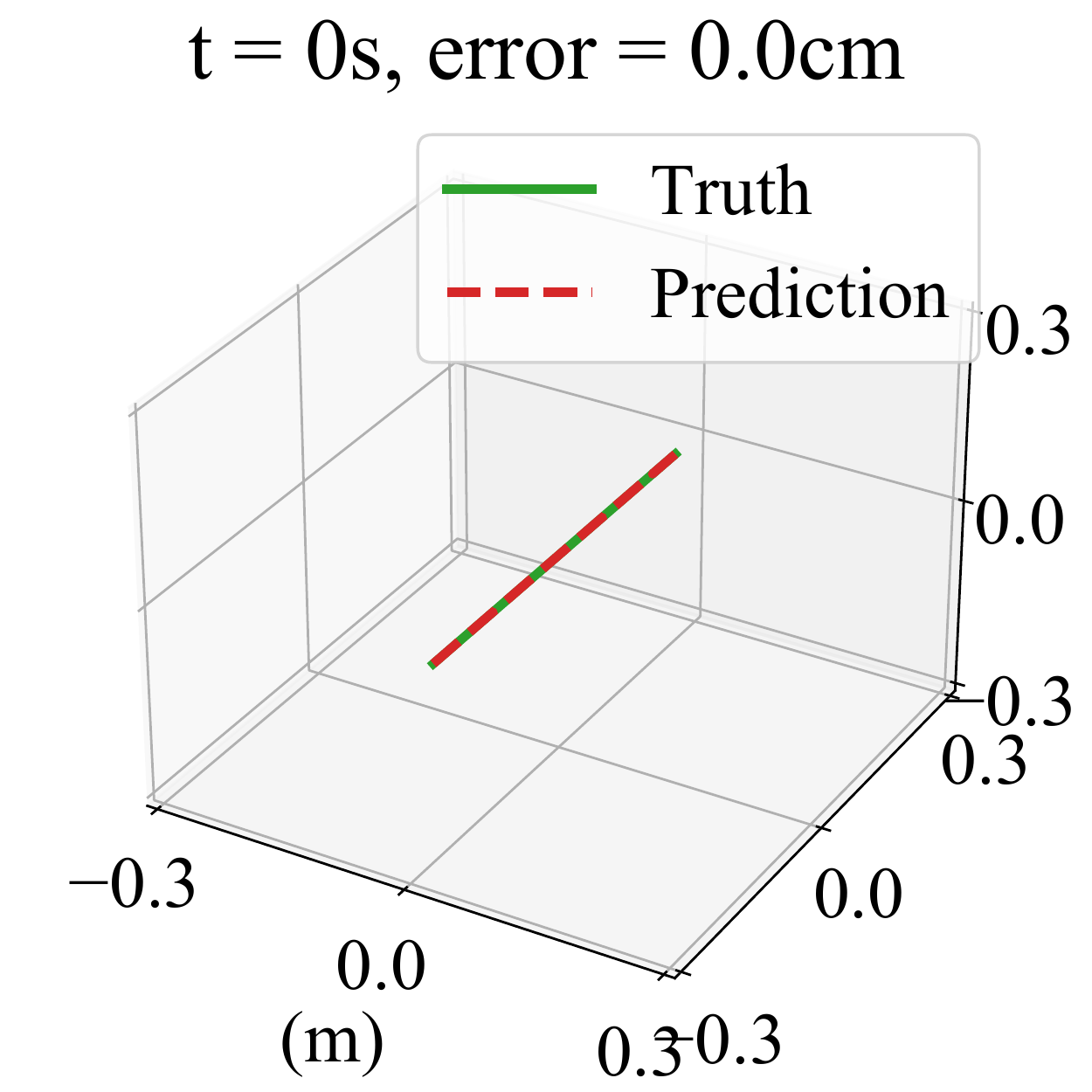} 
  }
  \hspace{-0.5cm}
  \subfigure{ 
    \includegraphics[width=2.8cm]{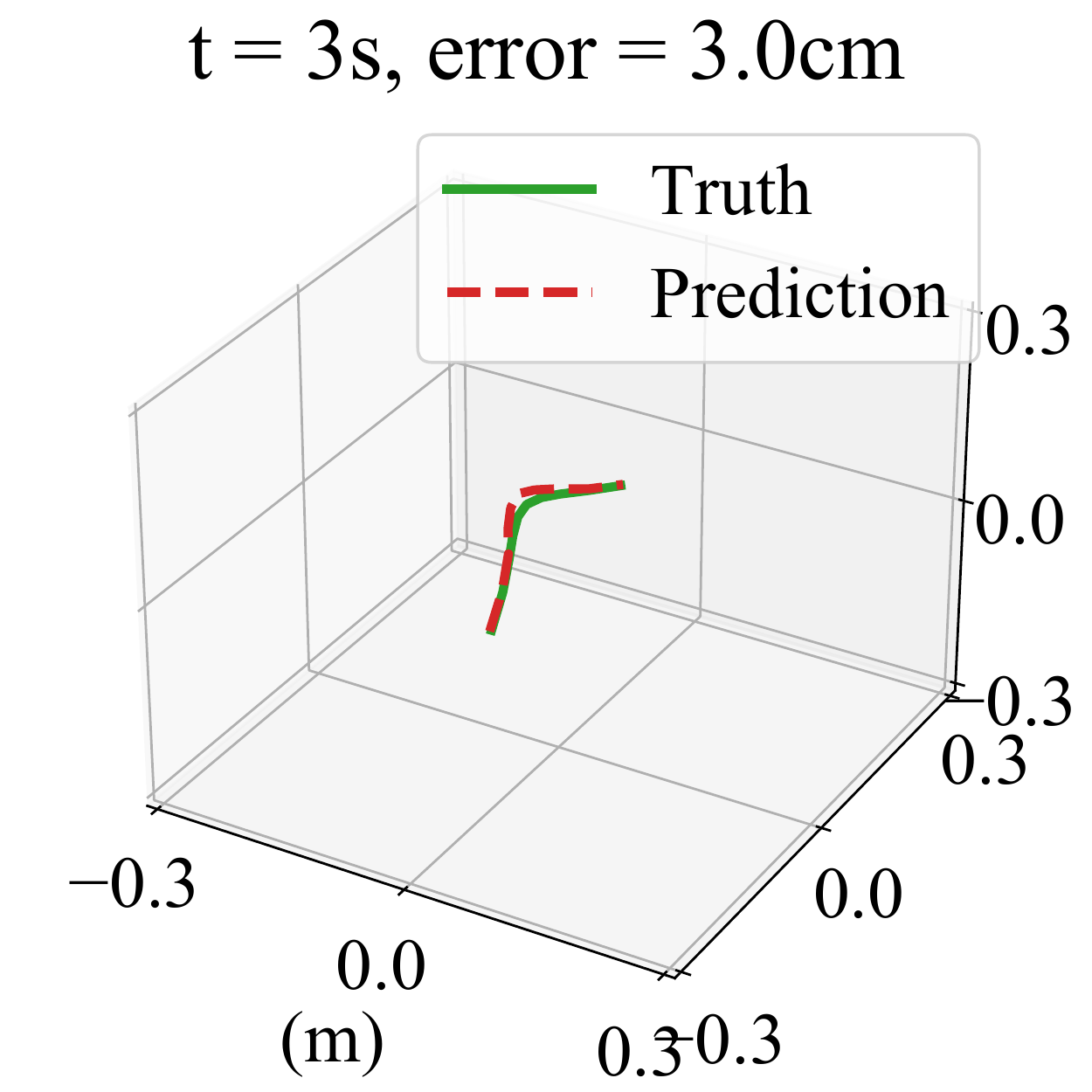} 
  } 
  \hspace{-0.5cm}
  \subfigure{ 
    \includegraphics[width=2.8cm]{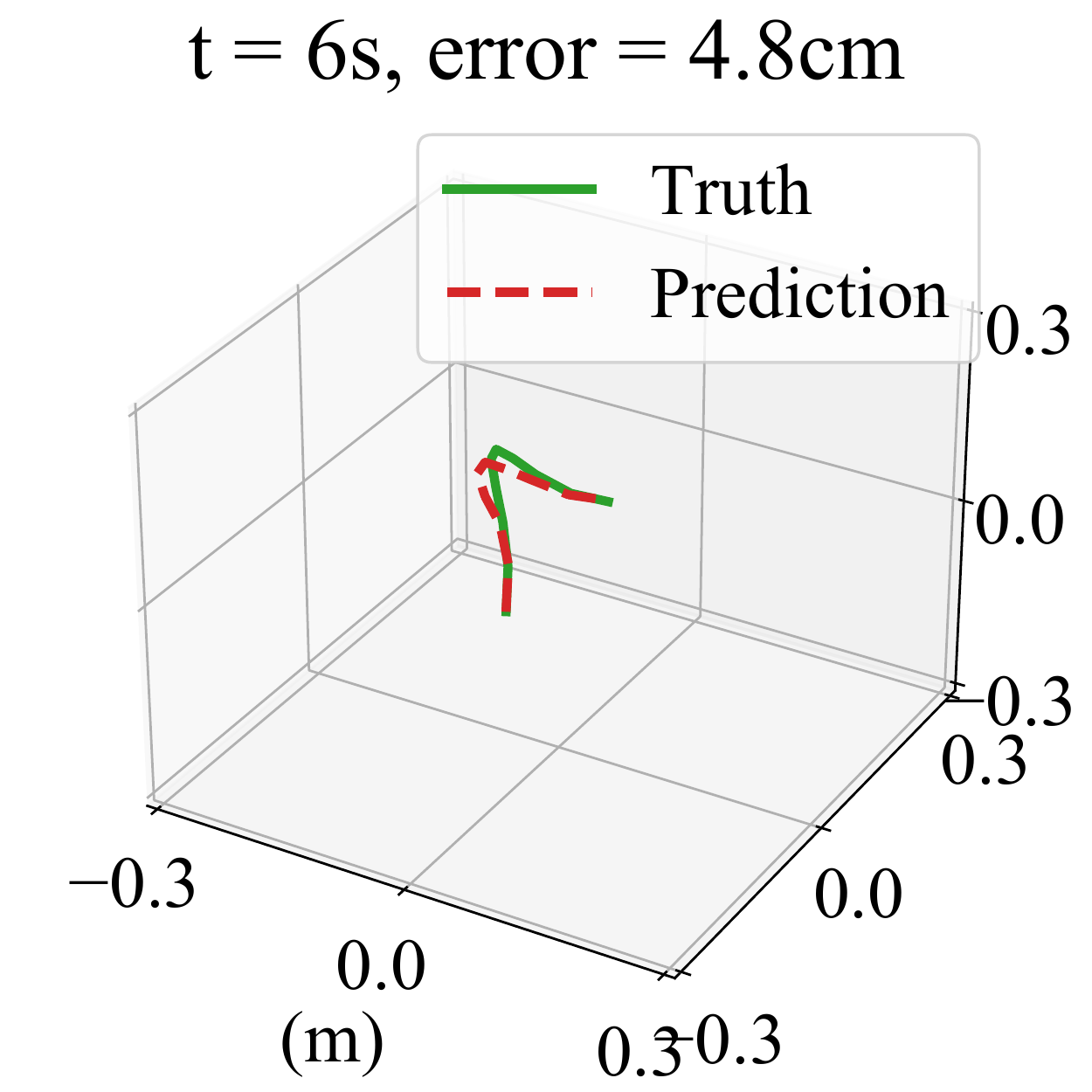} 
  }
  \hspace{-0.5cm}
  \subfigure{ 
    \includegraphics[width=2.8cm]{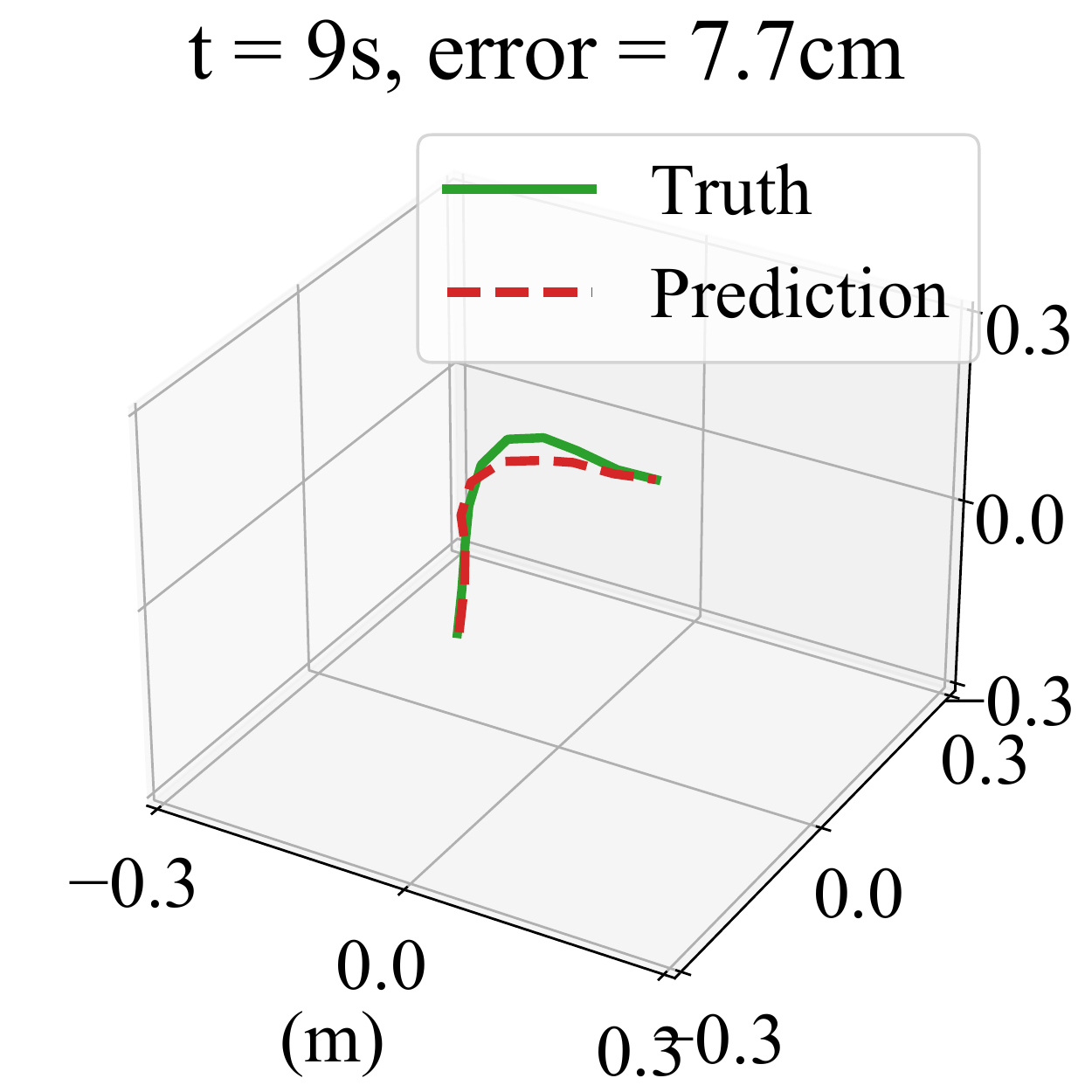} 
  } 
  \hspace{-0.5cm}
  \subfigure{ 
    \includegraphics[width=2.8cm]{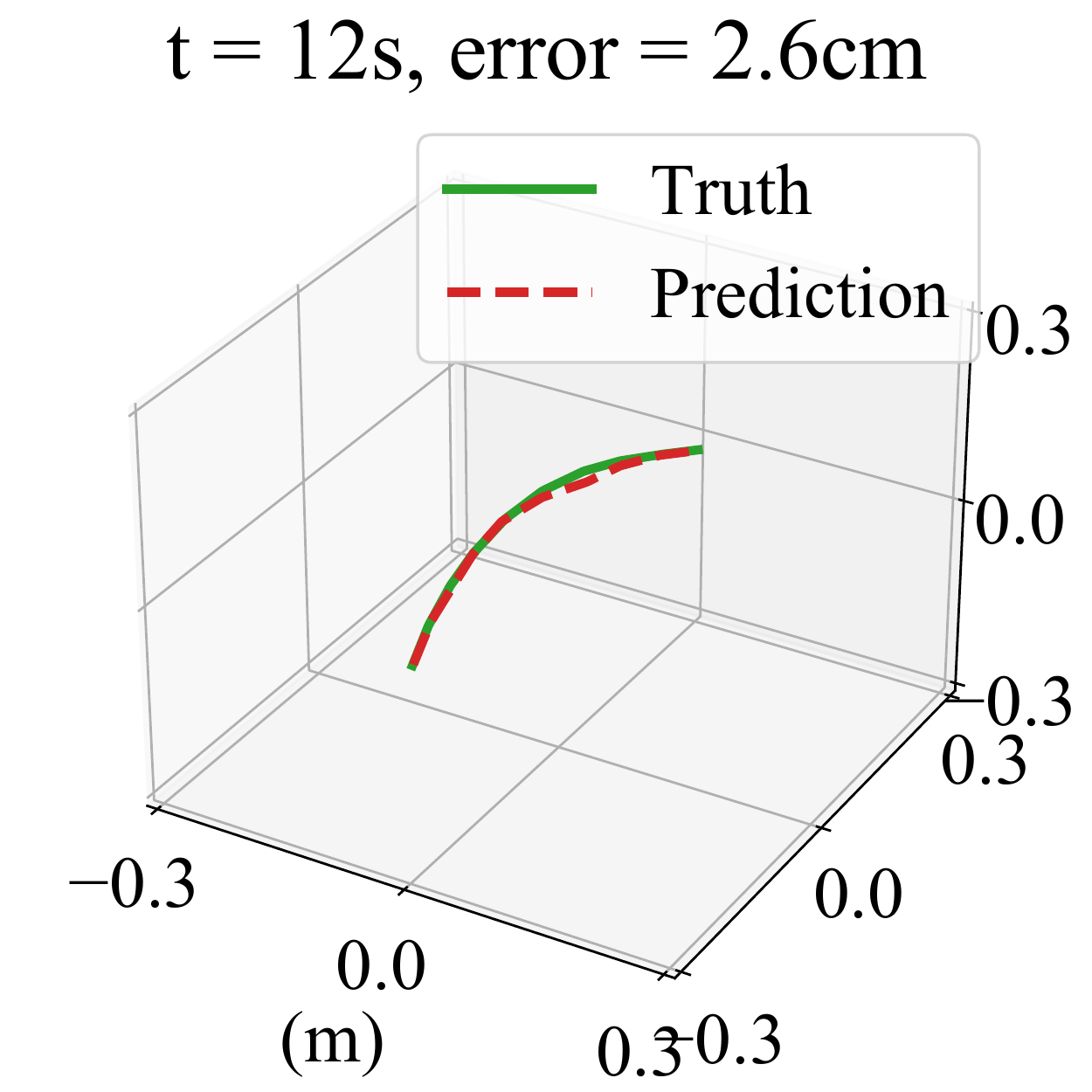} 
  } 
  \hspace{-0.5cm}
  \subfigure{ 
    \includegraphics[width=2.8cm]{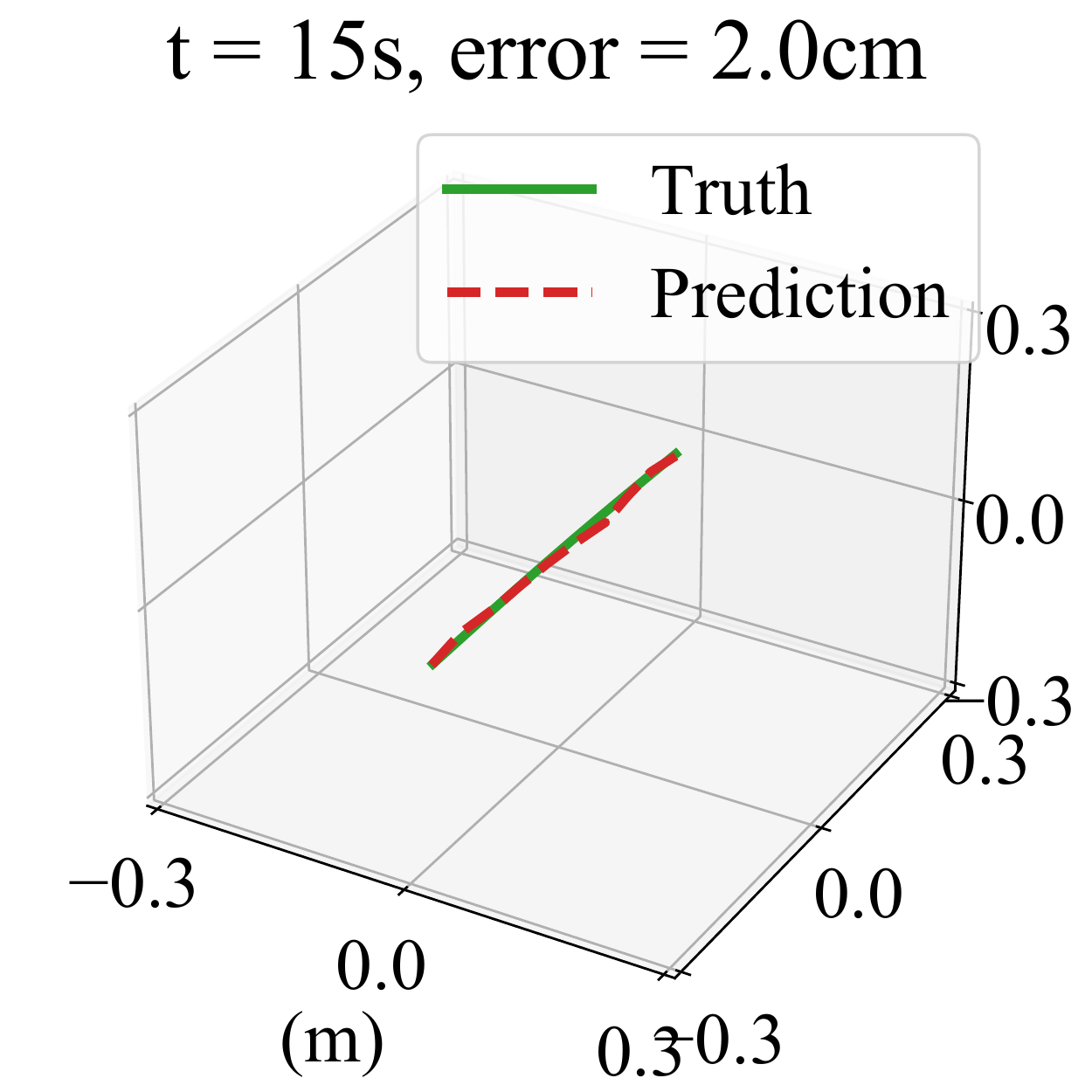} 
  } 
  \caption{Visualization of the long-term shape prediction results on a 3D test sequence, using our offline learned Jacobian model trained on 60k data.}
  \label{fig:exp_shape_prediction_vis}
\end{figure*}

We compare the offline modeling accuracy of the forward kinematics model (FKMs) and our Jacobian model on a certain DLO, by using the trained models to predict the DLO shape after 10 steps (the forward prediction by our Jacobian model is by (\ref{equation:discretized_state_equation})). 
FKMs based on different network architectures are implemented as baselines, including the multi-layer perceptron (MLP) \cite{mitrano2021learning}, the bidirectional LSTM (biLSTM) \cite{yan_self-supervised_2020}, and the combination of the interaction network and biLSTM (IN-biLSTM) \cite{yang2021learning}.
We also test the relationship between their modeling accuracy and the amount of training data. All training data and additional 10k test data are collected on the same DLO. Shown in Fig. \ref{fig:exp_jaco_fkm_compare}, the results indicate that our Jacobian model can achieve higher prediction accuracy with less training data than the FKMs. Compared with the general nonlinear FKMs, our Jacobian model implies a strong local linear prior, which is theoretically and practically reasonable. Hence, the learning efficiency is considerably improved.

We also visualize the shape prediction result of a 15s test sequence in Fig \ref{fig:exp_shape_prediction_vis}, using our offline learned Jacobian model trained on 60k data. The model iteratively predicts the positions of the features based on the initial state and the following motion of the end-effectors. It is shown that if there are enough offline data, the offline learned model can achieve long-term shape predictions with acceptable accuracy and be able to apply in manipulation planning.

\,
\subsubsection{Model Generalization Improvement}

\begin{figure} [tb]
  \centering 
    \includegraphics[width=8.7cm]{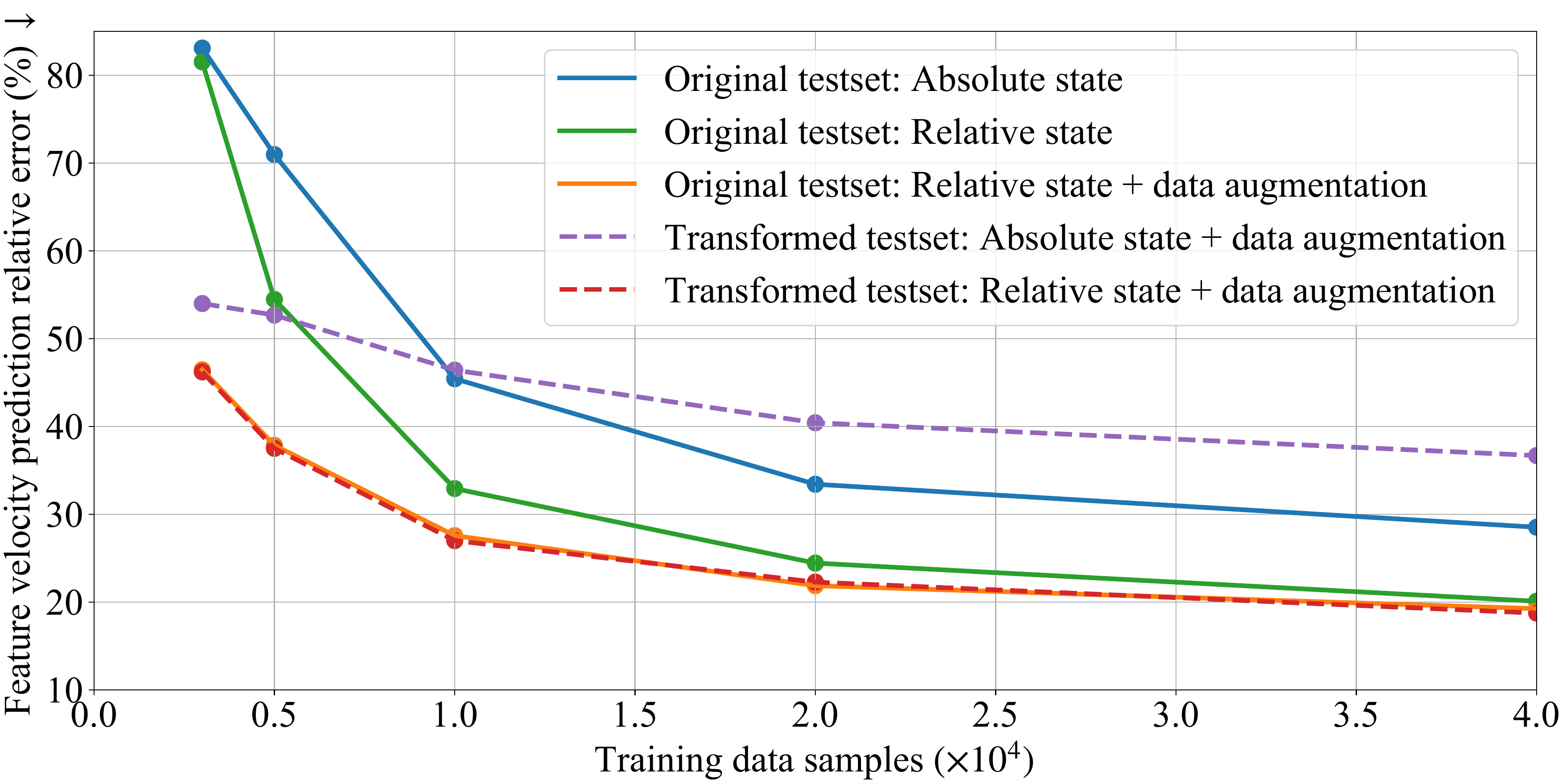} 
  \caption{Validation of the translation-invariant relative state representation and rotation data augmentation. The transformed testset is obtained by adding random translation and rotation transformation to the original testset. The used data augmentation includes both translation and rotation augmentation. All the training data and test data are from the same DLO.}
  \label{fig:exp_trans_rot_invariance}
\end{figure}

\begin{table}
\centering
\caption{Parameters of the 11 DLOs used in the simulation.}
\label{tab:dlos_params}
\begin{threeparttable}
\begin{tabular}{cccc} 
\toprule
\begin{tabular}[c]{@{}c@{}}DLO Index\end{tabular} & \begin{tabular}[c]{@{}c@{}}Length (m)\end{tabular} & \begin{tabular}[c]{@{}c@{}}Diameter (mm)\end{tabular} & \begin{tabular}[c]{@{}c@{}}Collected largest\\relative deformation\tnote{a}\\2D / 3D (m)\end{tabular} \\ 
\hline
0 & 0.5 & 10 & 0.14 / 0.14 \\
1 & 0.3 & 16 & 0.08 / 0.08 \\
2 & 0.4 & 6 & 0.10 / 0.10 \\
3 & 0.5 & 18 & 0.13 / 0.15 \\
4 & 0.6 & 8 & 0.14 / 0.15 \\
5 & 0.7 & 20 & 0.18 / 0.19 \\
6 & 0.8 & 10 & 0.20 / 0.22 \\
7 & 0.9 & 22 & 0.24 / 0.27 \\
8 & 1.0 & 12 & 0.28 / 0.26 \\
9 & 1.1 & 24 & 0.33 / 0.33 \\
10 & 1.2 & 14 & 0.33 / 0.34 \\
\bottomrule
\end{tabular}
\begin{tablenotes}
     \item[a] The average of the largest 10\% relative deformations (\ref{equation:relative_deformation}) between the straight DLO and deformed DLO  in the offline collected dataset.
\end{tablenotes}
\end{threeparttable}
\end{table}

We introduce the model modification considering the translation-invariance and approximate scale-invariance in Section \ref{section:model_modification} and the rotation data augmentation in Section \ref{section:data_augmentation}. All these are to improve the model's generalization ability on different DLOs and different deformation shapes. To validate these methods, we design the following tests in 3D scenarios, in which the evaluation criterion is the relative prediction error of the feature velocity vector as (\ref{equation:relative_vel_pred_error}).

First, we validate the influence of the translation-invariant relative state representation and the rotation data augmentation on the data of a certain DLO. We respectively train models using absolute state or relative state, and whether using the translation and rotation data augmentation or not. Then, we respectively test the models on an original collected testset, and a testset with random translation and rotation transformation. From the results shown in Fig. \ref{fig:exp_trans_rot_invariance}, we find that: 1) both the relative state representation and rotation data augmentation contribute to the improvement of the modeling accuracy; 2) the rotation data augmentation can significantly reduce the model's over-fitting when the training dataset is small; 3) the proposed methods make the model data-efficient and perform equally well on the original collected testset and randomly transformed testset.

\begin{figure} [tb]
  \centering 
    \includegraphics[width=8.7cm]{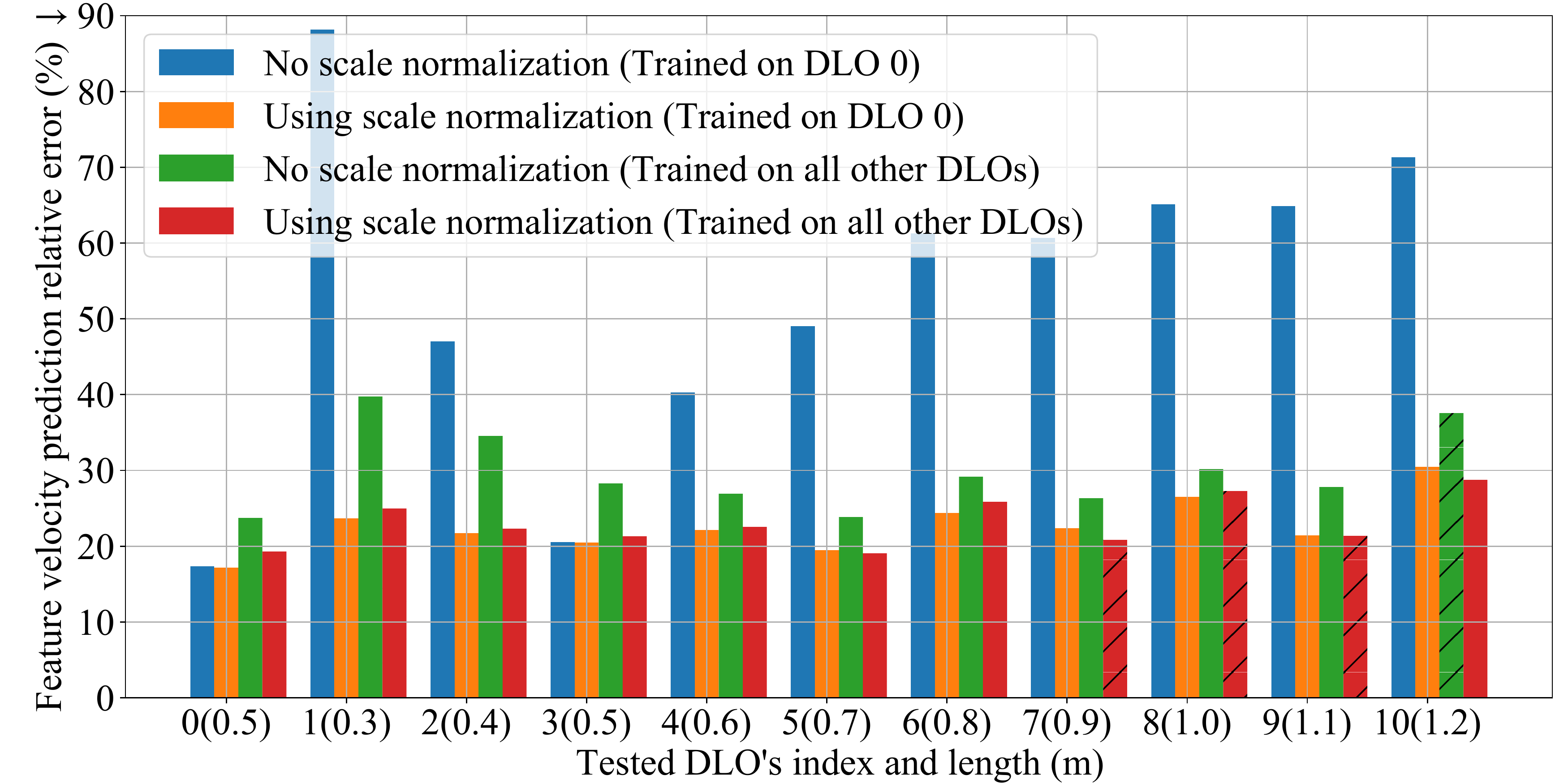} 
  \caption{Validation of the scale normalization. First, the models are trained on DLO 0 and then tested on all DLOs respectively (the histograms without slashes). Second, the models are trained using the combined data of 10 DLOs and then tested on the remaining DLO (the histograms with slashes).}
  \label{fig:exp_scale_invariance}
\end{figure}

Second, we validate the proposed scale normalization, by training and testing the models on DLOs of different lengths. We collect data of 11 DLOs of different lengths and thicknesses (details shown in Table \ref{tab:dlos_params}), and then compare the models with and without using the scale normalization in the following two experiments. First, we train the two models using 60k data of only DLO 0, and test them on 6k data of other DLOs respectively. Second, we train the models on $10 \times 6$k data of 10 DLOs, and test them on 6k data of the remaining one. The results in Fig. \ref{fig:exp_scale_invariance} indicate that: 1) when trained with only the data of DLO 0 of length 0.5m, the model using the scale normalization directly generalizes well to other DLOs of lengths from 0.3m to 1.2m. while the model not using it performs terribly; 2) when trained with the combined data of different DLOs, the model using the scale normalization still outperforms the other one. The results demonstrate that the approximate scale-invariance is reasonable, and the proposed scale normalization is effective.

\begin{table*}[tb]
\centering
\caption{Comparison between different models and controllers in 3D shape control tasks in the simulation.}
\label{tab:sim_control_diff_model_controller}
\begin{threeparttable}[b]
\begin{tabular}{cc|cccc} 
\toprule
Model\tnote{a} & Controller\tnote{b} & \begin{tabular}[c]{@{}c@{}}Average task error \\ of all cases (cm) $\downarrow$\end{tabular} & Success rate $\uparrow$ & \begin{tabular}[c]{@{}c@{}}Average task error\\ of successful cases (cm) $\downarrow$\end{tabular} & \begin{tabular}[c]{@{}c@{}}Average task time\\ of successful cases (s) $\downarrow$\end{tabular} \\ 
\hline
FKM & MPC & 5.780 & 69/100 & 2.035 & 12.371 \\
Our Jacobian model & MPC & 3.588 & 81/100 & 1.556 & 9.567 \\
Our Jacobian model & Naive P controller & 12.163 & 52/100 & 0.380 & 10.662 \\
Our Jacobian model & Our controller & 2.680 & 80/100 & 0.422 & 5.739 \\
\bottomrule
\end{tabular}
\begin{tablenotes}
     \item[a] All offline models are trained with the same 10$\times$6k data of DLO 1 to 10.
     \item[b] The online model adaptation of our method is not executed for a fair comparison. 
\end{tablenotes}
\end{threeparttable}
\end{table*}

\begin{table*}[tb]
\centering
\caption{Comparison with the existing methods in 2D and 3D DLO shape control tasks in the simulation.}
\label{tab:sim_control_diff_methods}
\begin{threeparttable}[b]
\begin{tabular}{c|cc|cccc} 
\toprule
Scenario & Methods & \begin{tabular}[c]{@{}c@{}}Offline \\ training data\end{tabular} & \begin{tabular}[c]{@{}c@{}}Average task error \\ of all cases (cm) $\downarrow$\end{tabular} & Success rate $\uparrow$ & \begin{tabular}[c]{@{}c@{}}Average task error\\ of successful cases (cm) $\downarrow$\end{tabular} & \begin{tabular}[c]{@{}c@{}}Average task time\\ of successful cases (s) $\downarrow$\end{tabular} \\ 
\hline
\multirow{4}{*}{2D} & WLS & - & 9.016 & 63/100 & 0.186 & 6\tnote{a} + 5.500 \\
 & FKM+MPC & 10$\times$6k & 7.944 & 54/100 & 1.961 & 10.961 \\
 & Ours & 10$\times$0.2k & 1.175 & 90/100 & \textbf{0.013} & 5.370 \\
 & Ours & 10$\times$6k & \textbf{0.185} & \textbf{95/100} & \textbf{0.013} & 5.397 \\ 
\hline
\multirow{4}{*}{3D} & WLS & - & 9.133 & 64/100 & 0.953 & 12\tnote{b} + 7.317 \\
 & FKM+MPC & 10$\times$6k & 5.780 & 69/100 & 2.035 & 12.371 \\
 & Ours & 10$\times$0.2k & 4.996 & 69/100 & 0.626 & 10.793 \\
 & Ours & 10$\times$6k & \textbf{1.223} & \textbf{94/100} & \textbf{0.175} & 5.777 \\
\bottomrule
\end{tabular}
\begin{tablenotes}
     \item[a,b] Time for initializing the Jacobian matrix estimation.
\end{tablenotes}
\end{threeparttable}
\end{table*}

\subsection{Shape Control with Online Model Adaptation} \label{subsection:sim_shape_control}

We detailedly analyze the proposed method for DLO shape control in 3D simulation tasks, and compare it with the existing methods in both 2D and 3D tasks. We conduct all tasks on DLO 0, in which 100 different feasible desired shapes are randomly chosen for testing. In most cases, there is a large deformation between the desired shape and initial shape. All DLO features are set as the target points. The performance criteria are introduced in Section \ref{subsubsection:metrics_shape_control}.

\,
\subsubsection{Effect of Online Model Adaptation}
In section \ref{section:adaptive_controller}, we propose an adaptive controller through online model updating to compensate for the offline modeling error owing to insufficient offline training or changes of the DLO properties. We validate the effect of the online adaptation for different initial offline models, and with different online learning rates.

\begin{figure} [tb]
  \centering 
    \includegraphics[width=8.7cm]{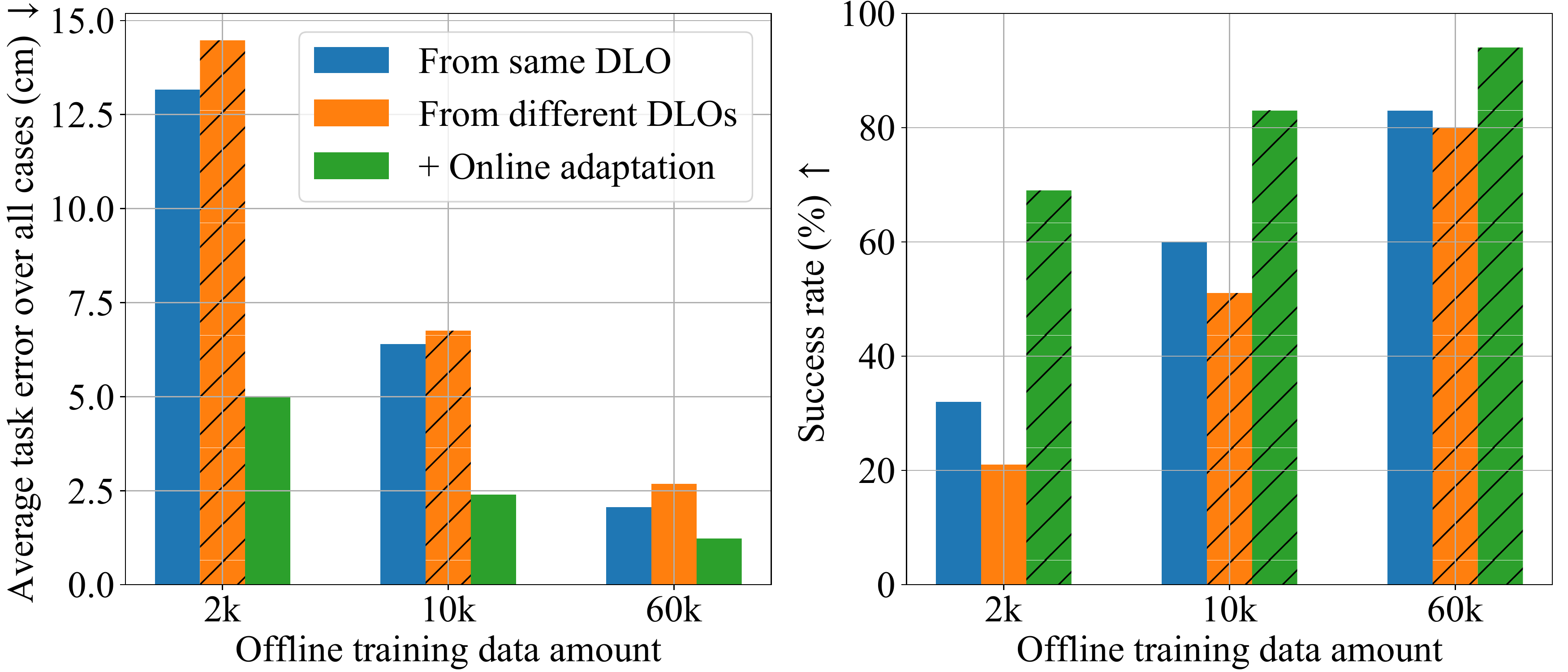} 
  \caption{Validation of the online model adaptation for different offline models. The histograms without slashes use the offline model trained on DLO 0 (the manipulated DLO); those with slashes use the offline model trained on the combined data of DLO 1 to 10.}
  \label{fig:exp_sim_control_offline_data}
\end{figure}

We first train six models based on different offline data, in which three are trained on 2k / 10k / 60k data from the DLO 0 (the manipulated DLO), and the other three are trained on 10$\times$0.2k / 10$\times$1k / 10$\times$6k data from the other 10 DLOs. The control performances of three settings are compared: 1) directly using the offline models trained on the same DLO; 2) directly using the offline models trained on different DLOs; 3) using the offline models trained on different DLOs and using the online adaptation (online learning rate $\eta = 1.0$). For each setting, the models using different amounts of offline data are tested. The results shown in Fig. \ref{fig:exp_sim_control_offline_data} indicate that: 1) the models trained on more offline data achieve better control performance; 2) the models trained on the same DLO are better than those trained on different DLOs; 3) the online adaptation can effectively compensate for the effect of insufficient offline data or different properties between the trained DLOs and the manipulated DLO.

\begin{figure} [tb]
  \centering 
    \includegraphics[width=8.7cm]{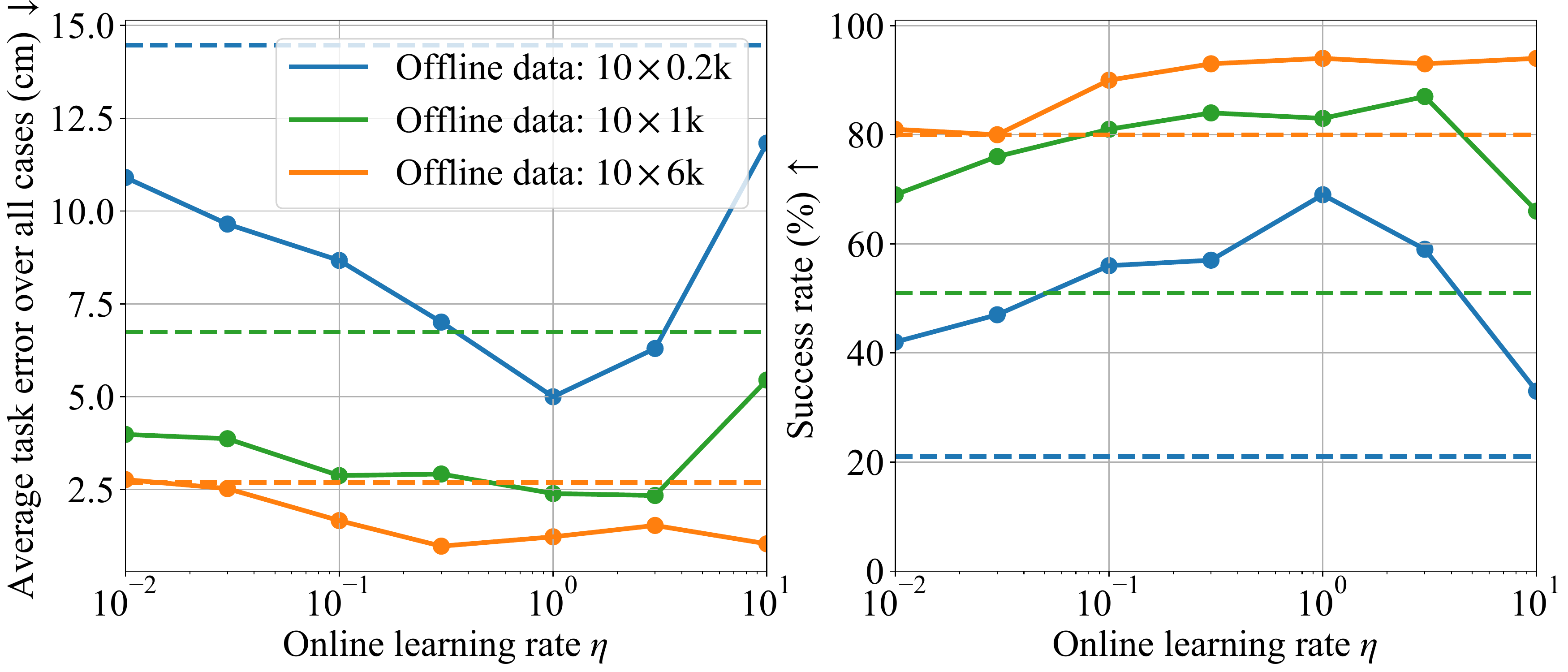}
  \caption{Sensitivity Analysis of the online learning rate $\eta$. The straight dash lines refer to the performances without the online model adaptation (their colors correspond to the used offline models). All the offline models are trained on DLO 1 to 10, but the data amounts are different.}
  \label{fig:exp_sim_control_online_lr}
\end{figure}

We further test the method's sensitivity to the online learning rate $\eta$, as shown in Fig. \ref{fig:exp_sim_control_online_lr}. It indicates that the online adaptation performs relatively stably with a large range of learning rates from $10^{-2}$ to $10^{1}$. We finally choose $\eta = 1.0$ as the most proper learning rate and use it in other simulation and real-world experiments.

\,
\subsubsection{Comparison between Different Models and Controllers}

We further analyze the effect of different models (FKM v.s. our Jacobian model) and different controllers (MPC v.s. our controller) in shape control.
The MPPI \cite{williams2017information} is used as the specific MPC.
Both models are offline trained on the same 10$\times$6k data of 10 DLOs different from the manipulated DLO. For a fair comparison, the online  adaptation is not executed.

First, we compare the two models by using the same MPC. Then, we compare the different controllers by using the same Jacobian model, in which we also test a naive P controller specified as 
\begin{equation}
    \bm \nu
    = - \alpha 
    \left(\hat{\bm J}^c(\bm s)\right)^{\dagger}
    \tilde{\Delta \bm x^c}
\end{equation}
The results in Table \ref{tab:sim_control_diff_model_controller} show that: 1) using the same MPC, our Jacobian model greatly outperforms the FKM; 2) using the same Jacobian model, our controller achieves a success rate similar to the MPC's but higher control accuracy; 3) the naive P controller performs poorly because of the possible singularity problem of the Jacobian matrix.

\,
\subsubsection{Comparison with Existing Methods}

We choose two representative classes of existing methods for comparison: 1) the offline method: learning a forward kinematics model of DLOs offline and using the MPC for shape control (FKM+MPC); 2) the online method: estimating the Jacobian matrix online using weighted least square estimation (WLS) and using the same control law as (\ref{equation:local_control_objective_estimate}). Compared with them, our method benefits from both the offline learning and online adaptation. All the offline models are trained on the data from the other 10 DLOs. Both 2D and 3D tasks are tested.

As shown in Table \ref{tab:sim_control_diff_methods}, our method significantly outperforms the compared methods on both the success rate and average task error.
Compared with the online WLS, our method performs better using only $10\times0.2$k (3.3 min) offline data, and much better using more offline data. 
Since the WLS online estimates a local model which only utilizes the limited local data, its performance on cases with large deformation is poor. Besides, it costs the longest time since it needs to initialize the Jacobian by moving the DLO ends in each DoF every time it starts. 
Compared with the offline FKM+MPC, our method performs significantly better using the same $10\times6$k (100 min) offline data, and performs comparably using only 1/30 of the data.
The poor performance of the FKM+MPC is due to the FKM's lower offline modeling accuracy and the lack of further updating for the untrained manipulated DLO, which also causes its highest average task error over the successful cases. We visualize some of the tasks accomplished using our method in Fig. \ref{fig:sim_control_snapshots}.

\section{Real-world Experiment Results}

\begin{figure} [tb]
  \centering 
    \includegraphics[width=8.0cm]{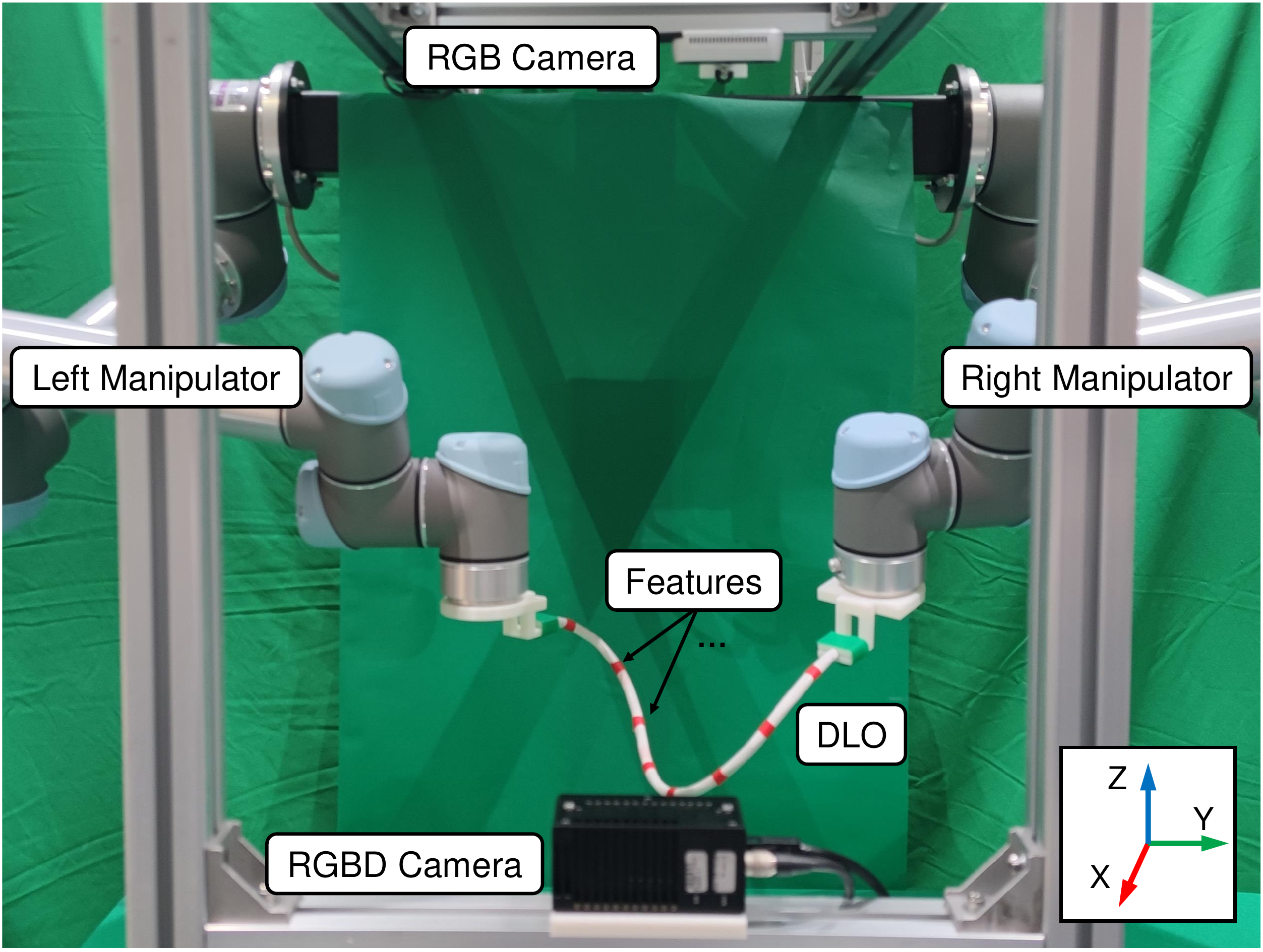} 
  \caption{Setup of the real-world experiments. The DLO is rigidly grasped by the two robot manipulators. For the 2D tasks, the DLO is placed on the table and tracked by the top RGB camera; for the 3D tasks, the DLO is tracked by the front structured-light RGBD camera.}
  \label{fig:exp_setting}
\end{figure}

\begin{figure}[tb]
  \centering 
    \includegraphics[width=8.7cm]{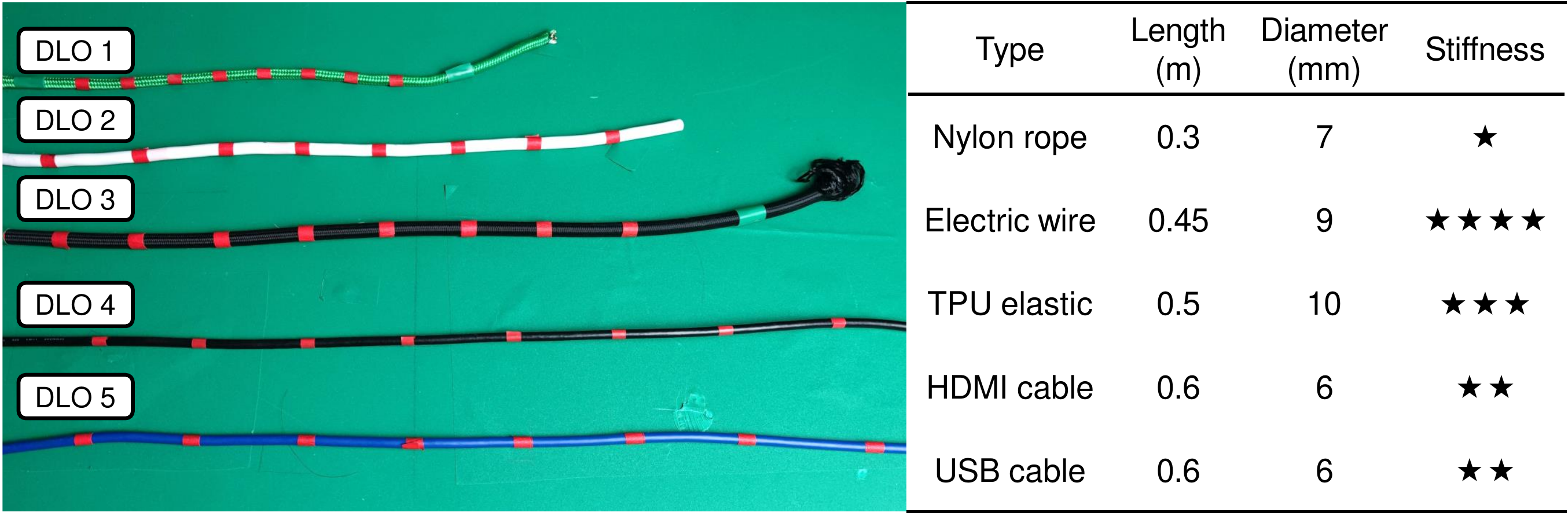} 
  \caption{The DLOs used in the real-world experiments
  and their parameters.}
  \label{fig:real_DLOs}
\end{figure}

\begin{figure} [tb]
  \centering 
  \subfigure[a case on DLO 1]{ 
    \includegraphics[width=4.2cm]{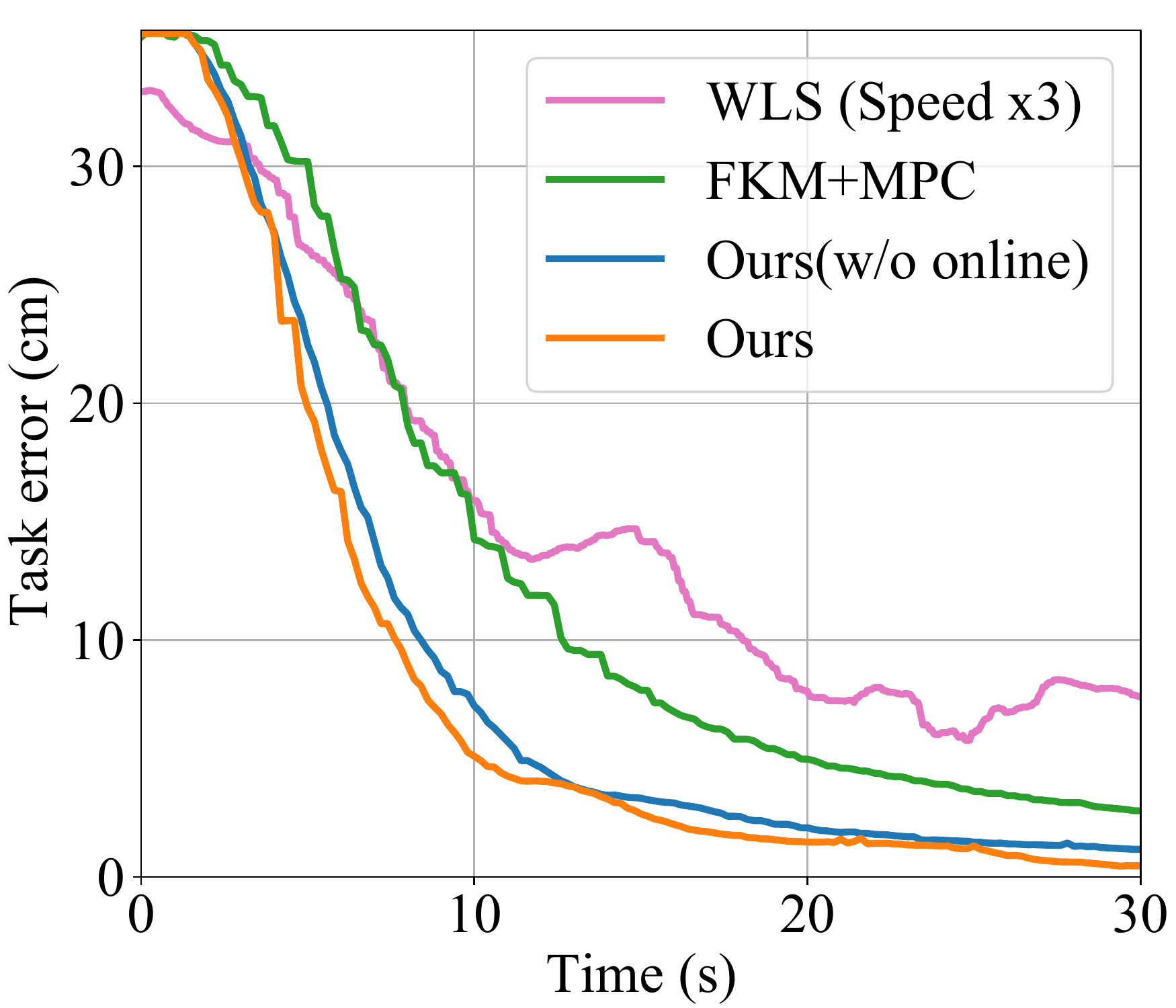} 
  } 
  \hspace{-0.4cm}
  \subfigure[a case on DLO 2]{ 
    \includegraphics[width=4.2cm]{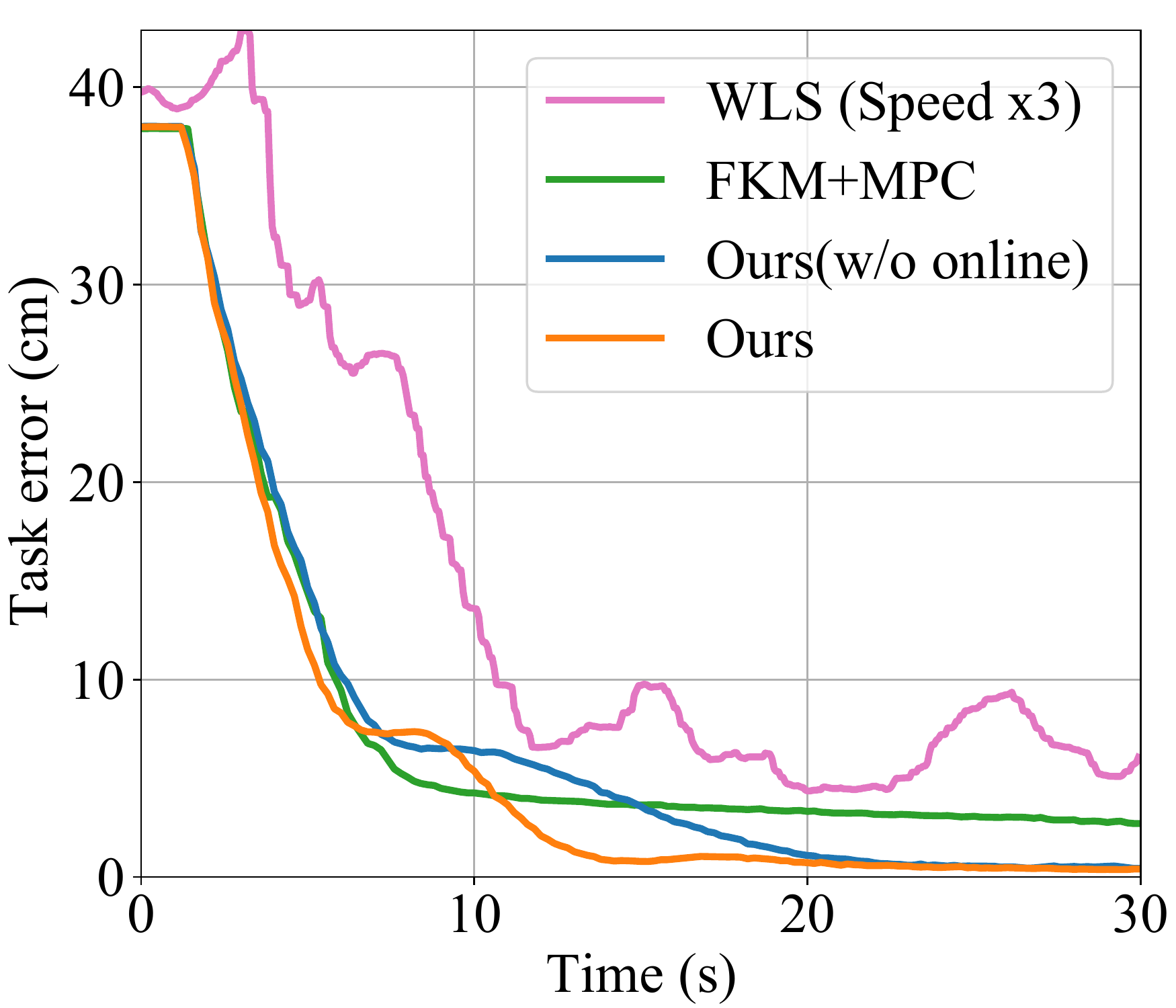} 
  }
  \hspace{-0.4cm}
  \subfigure[a case on DLO 3]{ 
    \includegraphics[width=4.2cm]{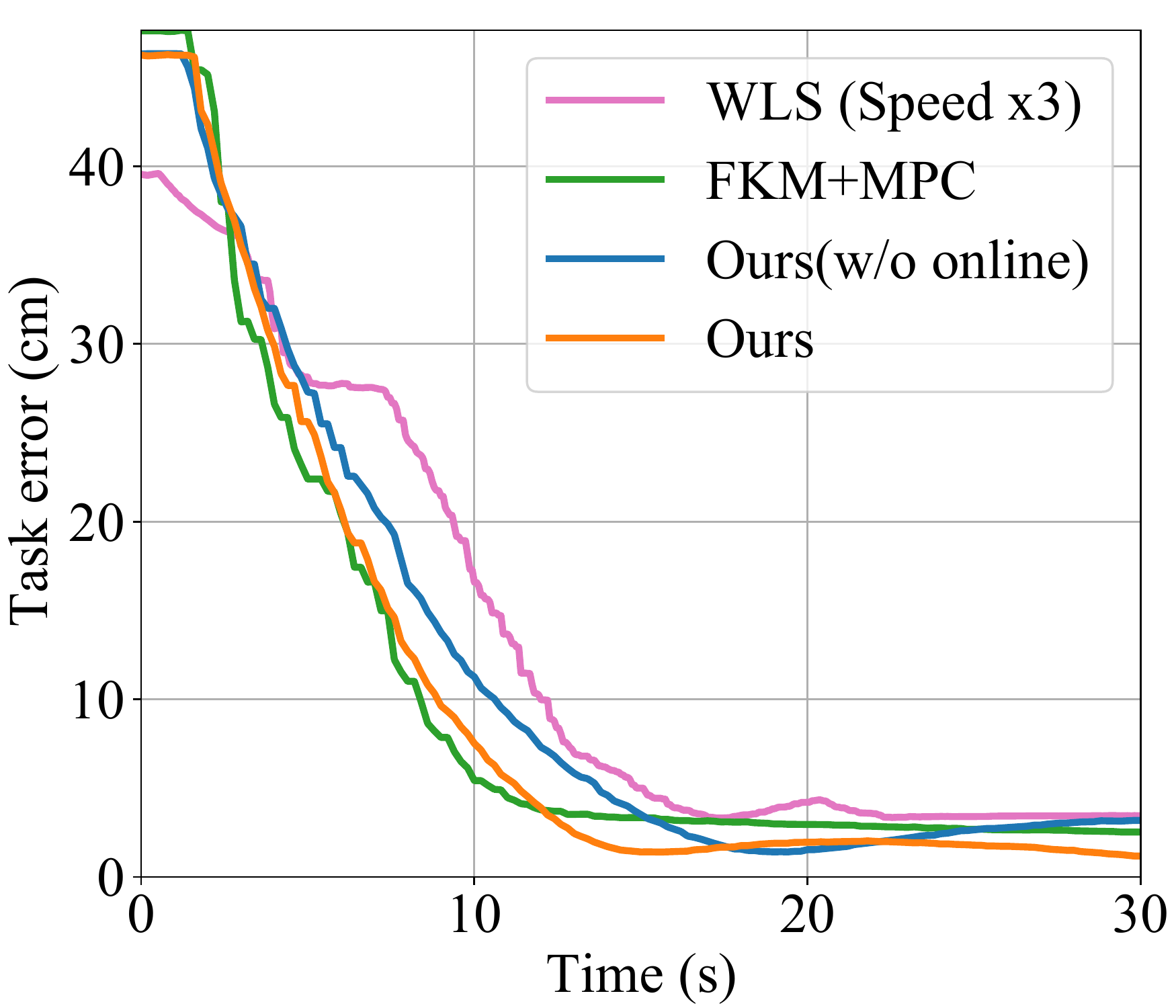} 
  }
  \hspace{-0.4cm}
  \subfigure[a case on DLO 5]{ 
    \includegraphics[width=4.2cm]{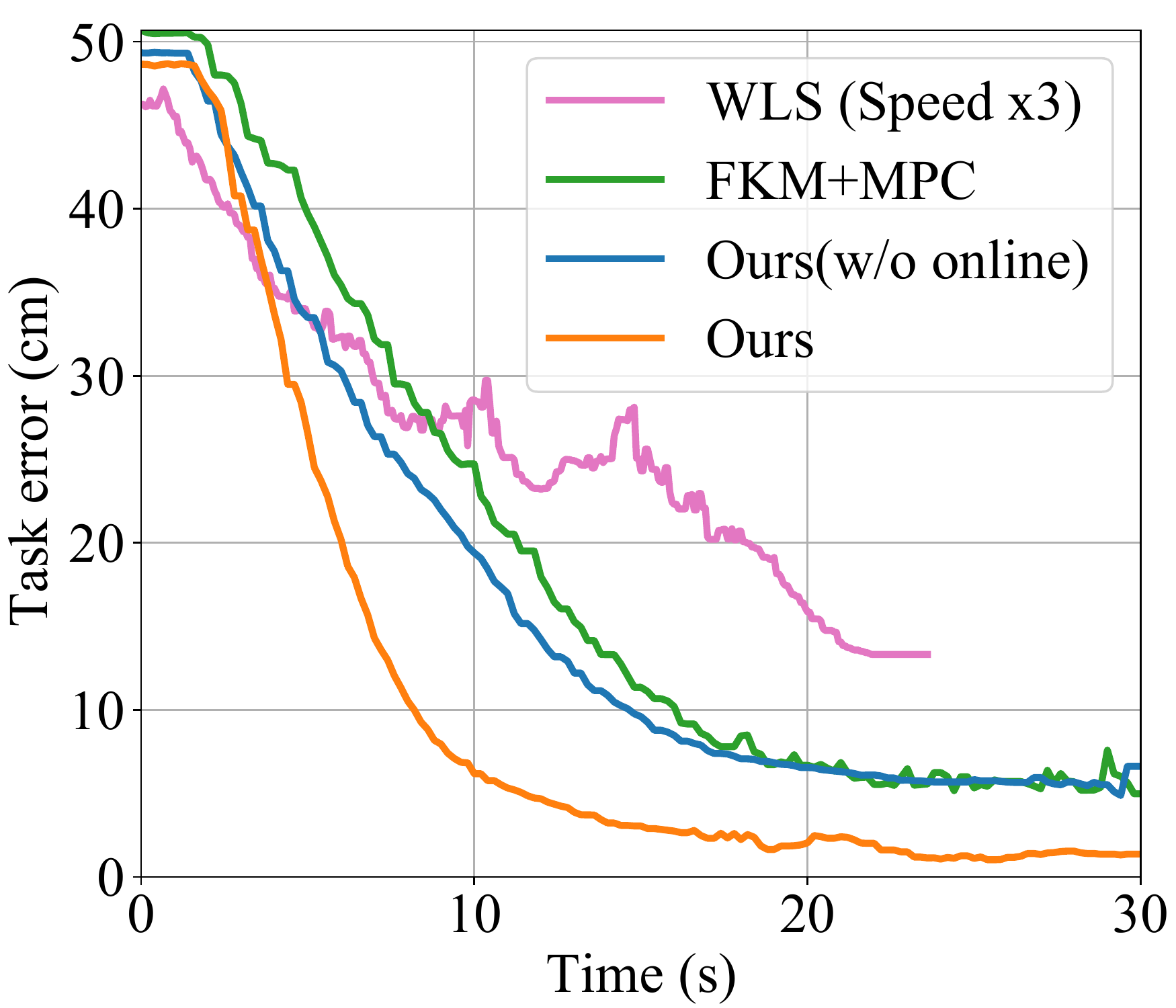} 
  }
  \caption{Control processes of four cases in the 3D real-world experiments. For the WLS, the process of initializing the Jacobian is not shown (so the starting point of the plotted line may be a little different from those of other methods), and the time is scaled for better plotting. In (d), the WLS stops because the unreasonable robot motion causes the occlusion of the DLO.}
  \label{fig:exp_real_control_process}
\end{figure}

\subsection{Experiment Setup}

The experiment setup is shown in Fig. \ref{fig:exp_setting}. 
The two ends of the DLO are rigidly grasped by dual UR5 robots.
The positions and velocities of the DLO features are obtained by applying Kalman filters on the measurement of the positions of the red markers.
For the 2D tasks, the DLO is placed on a table and tracked by a top calibrated RGB camera; for the 3D tasks, the DLO is tracked by a front calibrated RGBD camera. The top RGB camera is a Realsense D435 (because the precision of its depth map is not enough for sensing the thin DLO, we just use it as an RGB camera), and the RGBD camera is a Percipio FS820-E1 which is a structured-light camera. 
The communication between the devices is based on ROS \cite{quigley2009ros}. 
Limited by the maximum frame rate of the RGBD camera, the control frequency is set as 5Hz ($\delta t = 0.2$). 

The hyper-parameters for the controller are set as: $\alpha = 0.3$, $\epsilon_s = 0.02$, and $\lambda_0 = 0.3 / 1.0$ for the 2D/3D tasks; those for the online model adaptation are set as: $T_w = 10$, $\gamma=10$, $\epsilon_v=0.01$, and $\eta=1.0$.

\subsection{Comparison with Existing Methods on Various DLOs}

\begin{table*}[tb]
\centering
\caption{Comparison with the existing methods in 2D and 3D DLO shape control tasks in the real-world experiments.}
\label{tab:real_control_diff_methods}
\begin{threeparttable}
\begin{tabular}{c|c|cccc} 
\toprule
Scenario & Methods & \begin{tabular}[c]{@{}c@{}}Average task error \\ of all cases (cm) $\downarrow$\end{tabular} & Success rate $\uparrow$ & \begin{tabular}[c]{@{}c@{}}Average task error\\ of successful cases (cm) $\downarrow$\end{tabular} & \begin{tabular}[c]{@{}c@{}}Average task time\\ of successful cases (s) $\downarrow$\end{tabular} \\ 
\hline
\multirow{4}{*}{2D} & WLS & 5.328 & 9/12 & 2.146 & 12\tnote{a} + 39.822 \\
 & FKM+MPC & 3.000 & 11/12 & 1.673 & 13.527 \\
 & Ours(w/o online adaptation) & 1.082 & \textbf{12/12} & 1.082 & 12.433 \\
 & Ours & \textbf{0.475} & \textbf{12/12} & \textbf{0.475} & 11.767 \\ 
\hline
\multirow{4}{*}{3D} & WLS & 7.475 & 4/12 & 3.461 & 24\tnote{b} + 63.2 \\
 & FKM+MPC & 3.584 & 9/12 & 2.911 & 14.311 \\
 & Ours(w/o online adaptation) & 1.680 & 11/12 & 1.318 & 11.491 \\
 & Ours & \textbf{0.757} & \textbf{12/12} & \textbf{0.757} & 9.283 \\
\bottomrule
\end{tabular}
\begin{tablenotes}
     \item[a,b] Time for initializing the Jacobian matrix estimation.
\end{tablenotes}
\end{threeparttable}
\end{table*}

\begin{figure*} [tb]
  \centering 
    \includegraphics[width=\textwidth]{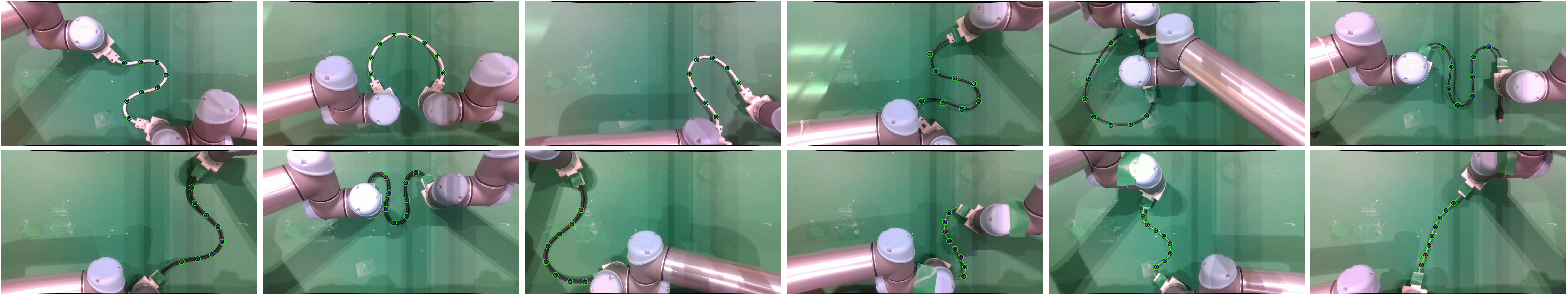} 
  \caption{2D shape control results using our method in the real-world experiments. Each picture shows a completed task. The blue points represent the tracking results of the red features (also represent the target points). The green+black circles represent the desired positions of the target points. In all cases, the DLO starts from a straight line in the center of the camera's field of view. (Please refer to our video for the full control processes.)}
  \label{fig:real_control_snapshots_2D}
\end{figure*}

\begin{figure*} [tb]
  \centering 
    \includegraphics[width=\textwidth]{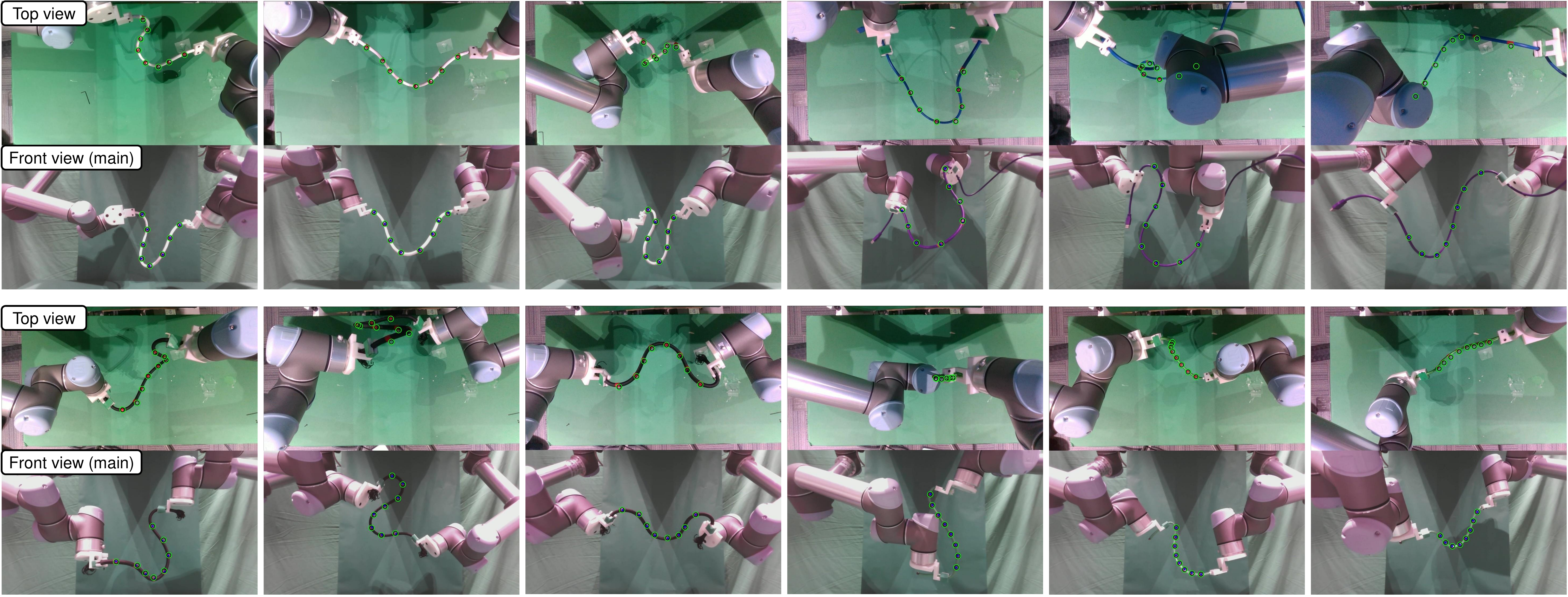} 
  \caption{3D shape control results using our method in real-world experiments. Each picture shows a completed task: the bottom sub-picture shows the front view captured by the RGBD camera, which is actually used for the shape control; the top sub-picture shows the top view captured by an RGB camera, which is only for a better illustration of the 3D DLO shapes to readers but not used for tracking or control. Note that the task error shown in the top view looks a little larger owing to the small calibration error between the two cameras. (Please refer to our video for the full control processes.)}
  \label{fig:real_control_snapshots_3D}
\end{figure*}

We validate the proposed method on several different DLOs with different materials, lengths, and thicknesses in the real world, and compare it with the WLS and FKM+MPC. The used DLOs and their detailed parameters are shown in Fig. \ref{fig:real_DLOs}.
In the 2D tests, DLO 1-4 are used; while in the 3D tests, DLO 5 is used instead of DLO 4, because DLO 4 cannot be sensed precisely by the structured-light camera since it is black and very thin.
We conduct three tests with different feasible desired shapes on each DLO. 
All the used offline models are trained using $10\times6$k data of 10 DLOs in the simulation, which means no real-world data are collected for offline training.
All DLO features are set as the target points.
For the WLS, we reduce the control gain $\alpha$ from 0.3 to 0.1 and the control frequency to 1Hz, because otherwise it performs significantly unstable and unsmooth. For a fair comparison, we run the WLS for 90s on each test, while we run the other methods for 30s.

The results are summarized in Table \ref{tab:real_control_diff_methods}, and the control processes of four cases are shown in Fig. \ref{fig:exp_real_control_process}.  
With the help of the online model adaptation, our method accomplishes all 24 tasks in 2D and 3D scenarios, and achieves the fastest and most precise control. 
The FKM+MPC accomplishes most of the tests but with larger task errors.
The WLS has the lowest success rate and the highest task error, demonstrating that the WLS is unsuitable for large deformation control of DLOs. In the previous research, the good performance of the WLS (or other online Jacobian estimation methods) relies on a good initialization of the Jacobian matrix by moving the DLO ends in each DoF. However, when the shape is far from the initial shape, it is hard to estimate a sufficiently accurate Jacobian matrix using only local online data in the presence of measuring errors, especially in 3D scenarios with more DoFs.

All the shape control results using our method are visualized in Fig. \ref{fig:real_control_snapshots_2D} and Fig. \ref{fig:real_control_snapshots_3D}. 
In addition, Table \ref{tab:deformation_magnitude} lists the deformation magnitudes (defined in Section \ref{subsubsection:deformation_magnitude}) between the initial shapes and desired shapes.
To the best of our knowledge, none of the existing works have achieved shape control tasks of elastic DLOs with such large deformations in the real world.
For instance, in \cite{wang2022offline}, the largest deformation on a 0.75m DLO in the real world is 0.14m in translation and 0.04m in relative deformation; in \cite{lagneau_automatic_2020}, the largest deformation on a 0.15m DLO is 0.018m (combination of translation and relative deformation).

\begin{table}
\centering
\caption{Deformation magnitudes of the shape control tasks in the simulation and real-world experiments.}
\label{tab:deformation_magnitude}
\begin{tabular}{cc|c|cc} 
\toprule
\multicolumn{2}{c|}{Scenario} & \begin{tabular}[c]{@{}c@{}}DLO\\length (m)\end{tabular} & \begin{tabular}[c]{@{}c@{}}Translation\\max / mean~ (m)\end{tabular} & \begin{tabular}[c]{@{}c@{}}Relative Deformation\\max / mean~(m)\end{tabular} \\ 
\hline
\multirow{2}{*}{Sim} & 2D & 0.5 & 0.51 / 0.24 & 0.19 / 0.11 \\ 
\cline{2-5}
 & 3D & 0.5 & 0.54 / 0.31 & 0.20 / 0.11 \\ 
\hline
\multirow{8}{*}{Real} & \multirow{4}{*}{2D} & 0.3 & 0.14 / 0.09 & 0.11 / 0.08 \\
 &  & {0.45} & {0.18 / 0.10} & {0.14 / 0.10} \\
 &  & {0.5} & {0.35 / 0.21} & {0.14 / 0.13} \\
 &  & {0.6} & {0.27 / 0.14} & {0.16 / 0.14} \\ 
\cline{2-5}
 & \multirow{4}{*}{{3D}} & {0.3} & {0.10 / 0.06} & {0.11 / 0.08} \\
 &  & {0.45} & {0.08 / 0.07} & {0.10 / 0.09} \\
 &  & {0.5} & {0.11 / 0.08} & {0.15 / 0.11} \\
 &  & {0.6} & {0.12 / 0.08} & {0.13 / 0.11} \\
\bottomrule
\end{tabular}
\end{table}

\subsection{Cases of Study}

\begin{figure} [tb]
  \centering 
    \includegraphics[width=8.7cm]{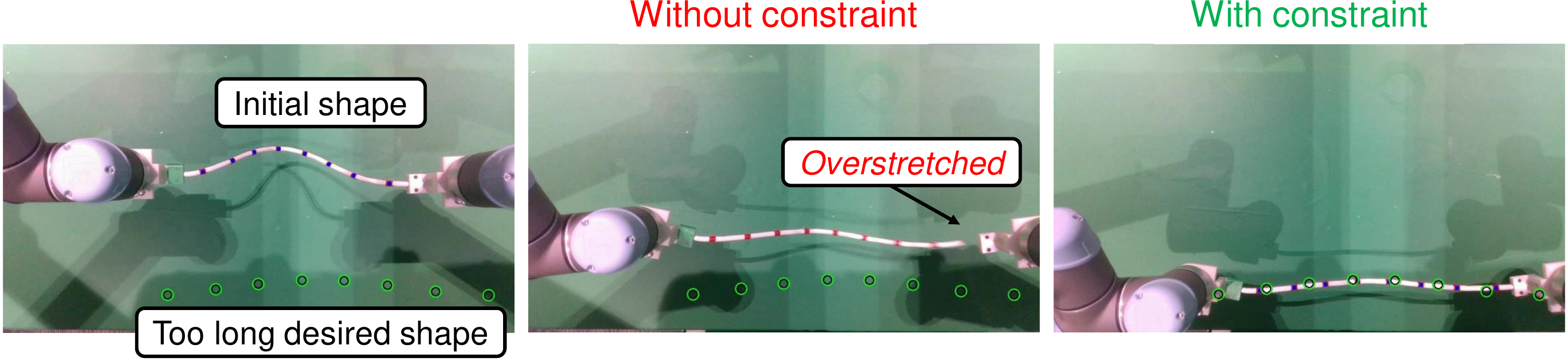} 
  \caption{Case to show the effect of the constraints for avoiding overstretching. The DLO is manipulated to an infeasible desired shape whose length is 1.3 times the length of the DLO.}
  \label{fig:exp_over-stretch}
\end{figure}

\begin{figure} [tb]
  \centering 
    \includegraphics[width=8.7cm]{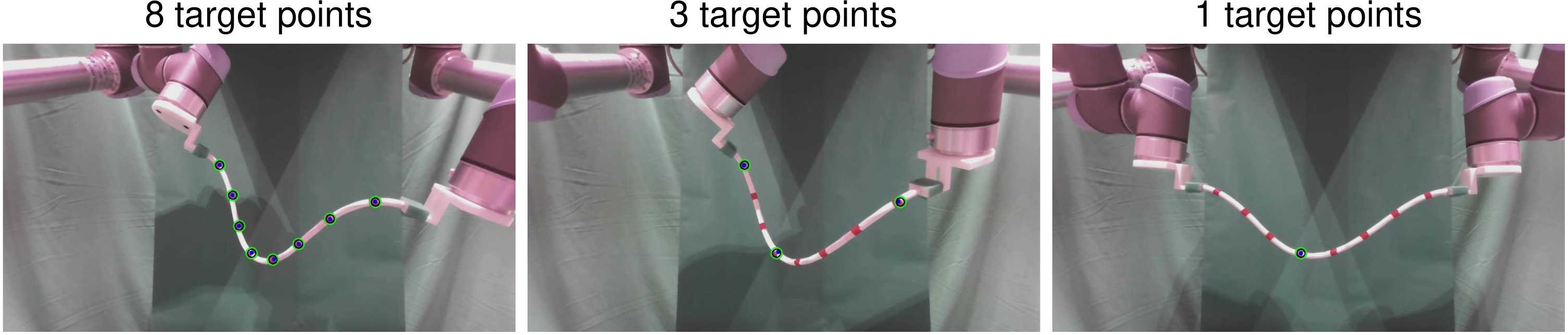} 
  \caption{Case of different choices of the target points. From left to right: all features / three features / one feature are chosen as the target points.}
  \label{fig:exp_diff_target_points}
\end{figure}

\begin{figure} [tb]
  \centering 
    \includegraphics[width=8.8cm]{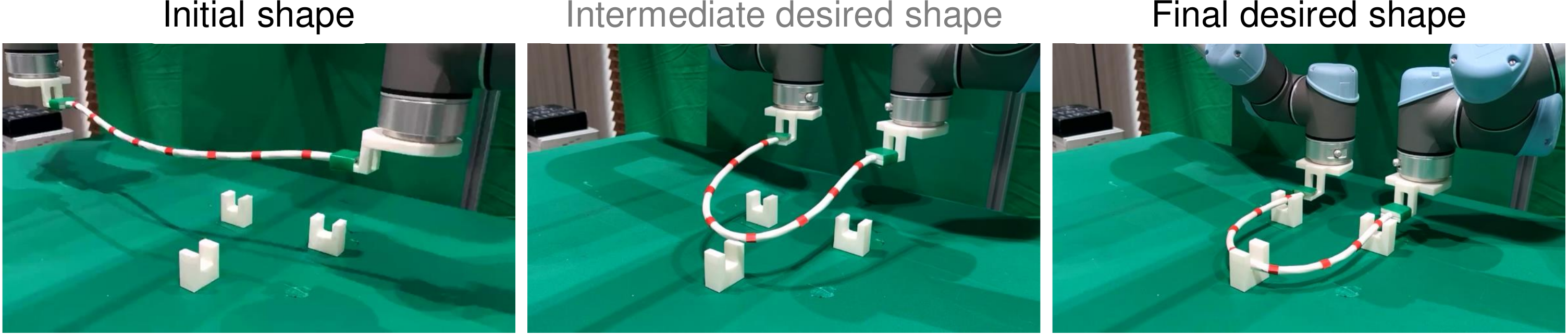}
  \caption{Case of applications: embedding the DLO into the grooves. The DLO is first manipulated to an intermediate desired shape, and then the final desired shape, to complete the whole task.}
  \label{fig:exp_groove}
\end{figure}

\begin{figure} [tb]
  \centering 
    \includegraphics[width=8.8cm]{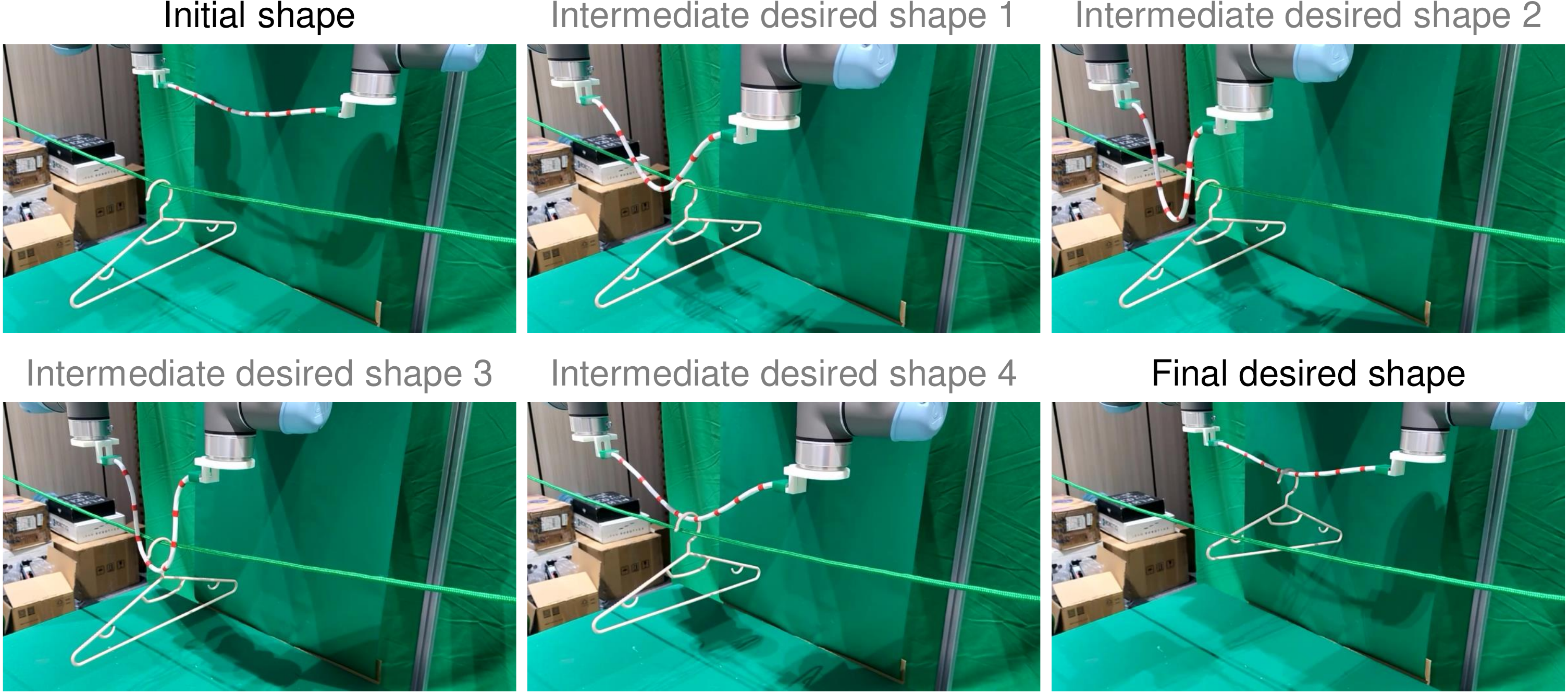} 
  \caption{Case of applications: hooking up a hanger. The robot end-effectors are not allowed to reduce their height, so they can reach the hanger only by deforming the DLO.}
  \label{fig:exp_hook}
\end{figure}

\begin{figure} [tb]
  \centering 
    \includegraphics[width=8.8cm]{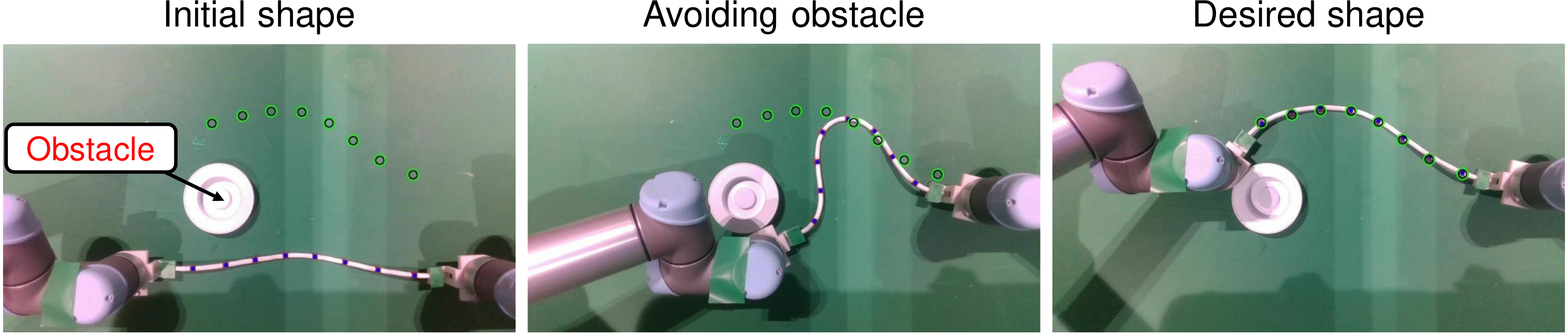} 
  \caption{Case of using the MPC with our Jacobian model for obstacle avoidance. Note that such an approach requires an accurate offline model.}
  \label{fig:exp_obstacle}
\end{figure}

\subsubsection{Avoiding Overstretching} \label{section:avoid_over-stretch}
We design a case to validate the proposed constraints for avoiding overstretching the DLO. As shown in Fig. \ref{fig:exp_over-stretch}, we set an infeasible desired shape whose length is 1.3 times the length of the DLO. The results show that, if the constraint is not used, the DLO is overstretched and falls out of the gripper; in contrast, if the constraint is used, {the DLO is manipulated to a shape as close to the infeasible desired shape as possible but not overstretched}.

\subsubsection{Different Choices of Target Points}
We show that any subset of the features can be set as the target points in Fig. \ref{fig:exp_diff_target_points}. It is found that the final shapes of choosing all eight features or only three features as the target points are similar, which indicates the coupling of the features. When choosing only one target point, there are infinite solutions, but the robots will achieve the task with minimal movements of the end-effectors.

\subsubsection{Application: embedding the DLO into grooves}
We show two cases of the potential applications of the proposed method. In the first case, we apply the method to embed the DLO into the grooves, as shown in Fig. \ref{fig:exp_groove}. The $1^{\rm st}, 4^{\rm th}, 5^{\rm th}$ and $8^{\rm th}$ features are set as the target points, and the 3-DoF translation and the rotation around the Z-axis are allowed for each robot end-effector. Besides the final desired shape in the grooves, we also define an intermediate desired shape above the grooves. The DLO is first controlled to the intermediate desired shape, and then the final desired shape, to complete the whole embedding tasks.

\subsubsection{Application: hooking up a hanger}
In the second case, we apply the proposed method to control the DLO to hook up a hanger which is initially hanging on a rope, as shown in Fig. \ref{fig:exp_hook}. Only the $4^{\rm th}$ and $5^{\rm th}$ features are set as the target points. Only the translation along the X and Y axis is allowed for the robot end-effectors, which means that the robots can reach the hanger only by deforming the DLO, but not by just reducing the height of the end-effectors. The whole complicated task is divided into several parts, and an intermediate desired shape is defined for each part. 
These two cases also show that our method can be used with different choices of the target points and the allowed DoFs of the end-effectors.

\subsubsection{Our Jacobian Model + MPC for Obstacle Avoidance}
In this case, we show that our Jacobian model can be used with the MPC for more complicated tasks such as obstacle avoidance. As shown in Fig. \ref{fig:exp_obstacle}, there is an obstacle between the initial shape and the desired shape. The robots must manipulate the DLO to the desired shape while avoiding the collision between the DLO and the obstacle. We formulate the control problem as (\ref{equation:MPC_formulation}), and apply the MPPI to calculate the control input. For the MPPI, the prediction horizon is set as $T_h=20$ (4 seconds), and the number of samples is set as 1000. The result demonstrates the long-term prediction ability of our Jacobian model. Note that such an approach is effective only if a sufficiently accurate offline model is available.

\section{Discussion and Conclusion}

\subsection{Limitations}
The limitations of our method include: 
1) Our method is designed for slow (quasi-static) manipulations and cannot be applied in dynamic manipulations where the inertia should be considered, because the derivation of our Jacobian model is based on the quasi-static assumption. 
{2) This work focuses on control but not planning. As a result, the moving path of the DLO may not be globally optimal. In addition, collisions between robot arms and self-intersection of the DLO may happen. In more complicated tasks, a global planner should be introduced to roughly plan a proper path or necessary intermediate desired shapes ahead.
3) We cannot determine whether a desired shape is feasible before manipulation. In our method, if the desired shape is infeasible, the DLO will stop as close to the desired shape as possible, such as the case in \ref{section:avoid_over-stretch} and Fig. \ref{fig:exp_over-stretch}.
One possible way to roughly study the feasibility of a desired shape is to use an energy-based DLO model to decide whether the shape is at a local minimum of the deformation energy.}

In addition, the perception of DLOs is a very challenging research topic; however, it is out of the scope of this work. In the real-world experiments, we simplify the sensing of DLO features by putting markers on the DLO, and manually remove the cases where most methods fail owing to occlusions. Recently, some researches have preliminarily investigated marker-free perception approaches. These approaches track the virtual points along the DLO from the point cloud using the Gaussian mixture model and other constraints, and can even handle slight occlusions \cite{tang2018track,chi2019occlusion,wang2021tracking}. These perception methods can be applied as the front end of our method in marker-free scenarios, in which the DLO features in our method are a subset of the virtual tracking points.

\subsection{Conclusion}
This work proposes a new scheme for large deformation control of DLOs with coupled offline and online learning of the unknown global deformation model. The combination of offline and online learning enables both accurate global modeling and further updating for new DLOs during actual manipulation, which allows our method to handle large deformation tasks and adapt well to new DLOs.
In the offline phase, an offline model is trained on random motion data of DLOs of different properties, to obtain an estimation with good generalization performance. Then, the offline model is seamlessly migrated to the online phase as an initial estimation. Finally, in the online phase, the shape control task is executed, while the model is concurrently updated based on online motion data to compensate for offline modeling errors.

In detail, we describe the global deformation model by a nonlinear mapping from the DLO configuration to a local Jacobian matrix, and prove its rationality. We also introduce several strategies to improve the model's training efficiency and generalization ability, including the scale normalization and rotation data augmentation. As for the controller, we formulate it as the optimal solution of a convex optimization problem, which considers the singularity of the Jacobian matrix and constrains the robots not to overstretch the DLO. We use the Lyapunov method to analyze the stability and convergence of the whole system.

We conduct a series of simulations and real-world experiments to demonstrate that our method can stably and precisely achieve large deformation control of DLOs, and greatly outperforms the existing data-driven methods. We demonstrate that our Jacobian model is more data-efficient, and the online adaptation effectively compensates for offline model errors owing to insufficient training or changes of DLO properties. Using the offline model trained with only simulation data, our method accomplishes all the 2D and 3D tasks on different DLOs in the real-world experiments with the highest accuracy and within roughly ten seconds only.

In terms of future work, we would like to study the marker-free perception method to make our method more practical in reality. {We would also like to introduce a high-level planner, which can be combined with this control method to achieve complex tasks that require not only accurate final control results but also proper moving paths.}




{\appendices

\section{Proof of Theorem 2} \label{appendix:stablity_analysis}

The stability of the system using the control law (\ref{equation:local_control_objective_estimate}) and the online updating law (\ref{equation:online_updating_law}) is analyzed as follows. 

First, since $\bm \nu = \bm 0$ is a possible solution of the optimization problem (\ref{equation:local_control_objective_estimate}), the optimal must be no more than $\frac{1}{2} \| \dot{\bm x}^c_{\rm ide} \|_2^2$. Thus, for the optimal solution $\bm \nu$, we have
\begin{equation}
    \frac{1}{2} \| \dot{\bm x}^c_{\rm ide} - \hat{\bm J}^c(\bm s) \bm \nu \|_2^2 + \frac{\lambda}{2} \| \bm \nu \|_2^2 \leq \frac{1}{2} \| \dot{\bm x}^c_{\rm ide} \|_2^2
\end{equation}
\begin{equation}
    -(\dot{\bm x}^c_{\rm ide})^{\transpose} \hat{\bm J}^c(\bm s) \bm \nu
    + \frac{1}{2} (\bm \nu)^{\transpose} (\hat{\bm J}^c(\bm s))^{\transpose} \hat{\bm J}^c(\bm s) \bm \nu
    + \frac{\lambda}{2} (\bm \nu)^{\transpose} \bm \nu
    \leq \bm 0
\end{equation}
\begin{equation} \label{equation:stability_analysis_0}
\begin{aligned}
    (\dot{\bm x}^c_{\rm ide})^{\transpose} \hat{\bm J}^c(\bm s) \bm \nu 
    \geq \frac{1}{2} (\bm \nu)^{\transpose} (\hat{\bm J}^c(\bm s))^{\transpose} \hat{\bm J}^c(\bm s) \bm \nu
    + \frac{\lambda}{2} (\bm \nu)^{\transpose} \bm \nu
    \geq \bm 0
\end{aligned}
\end{equation}
Subscribing (\ref{equation:idea_vel}) into it yields
\begin{equation} \label{equation:stability_analysis_1}
    (\tilde{\Delta \bm x^c})^{\transpose} \hat{\bm J}^c(\bm s) \bm \nu 
    \leq \bm 0
\end{equation}

Next, subscribing (\ref{equation:ew}) and (\ref{equation:all_c}) into (\ref{equation:jaco4}) and noticing that the desired position vector $\bm x^c_{\rm des}$ is fixed, we have:
\begin{equation} \label{equation:stability_analysis_2}
\begin{aligned}
    \Delta \dot{\bm x}^c
    &= \dot{\bm x}^c 
    = \bm J^c(\bm s) \bm \nu \\
    &= \hat{\bm J}^c(\bm s) \bm \nu - \hat{\bm J}^c(\bm s) \bm \nu + \bm J^c(\bm s) \bm \nu
    = \hat{\bm J}^c(\bm s) \bm \nu + \bm e^c
\end{aligned}
\end{equation}

Then, define a potential function of $\Delta\bm x^c$ as
\begin{equation} 
    P(\Delta\bm x^c) = 
    \left\{ \begin{array}{cl}
    \frac{1}{2}(\Delta \bm x^c)^{\transpose} \Delta \bm x^c  & , \quad \|\Delta \bm x^c\|_2 < \epsilon_x \\ 
     \epsilon_x \|\Delta \bm x^c\|_2 - \frac{\epsilon_x^2}{2}  & , \quad \rm{otherwise}
    \end{array}\right.
\end{equation}
and we have 
\begin{equation} \label{equation:stability_analysis_3}
    \frac{{\rm d} P(\Delta\bm x^c)}{{\rm d} \Delta\bm x^c}
    =  
    \left\{ \begin{array}{cl}
    \Delta \bm x^c  & , \quad \|\Delta \bm x^c\|_2 < \epsilon_x \\ 
    \epsilon_x \frac{\Delta \bm x^c}{\|\Delta \bm x^c\|_2}  & , \quad \rm{otherwise}
    \end{array}\right.
    = \tilde{\Delta \bm x^c}
\end{equation}

A Lyapunov-like candidate is given as 
\begin{equation} \label{equation:V}
    V = P(\Delta\bm x^c)
    + \frac{1}{2\eta} \sum_{k\in \mathcal{C}}\sum\limits_{i=1}^{12} \sum\limits_{j=1}^3 \Delta\bm W_{kij} \Delta\bm W_{kij}^{\transpose}
\end{equation}
Differentiating (\ref{equation:V}) with respect to time:
\begin{equation} \label{equation:dotV_2}
\begin{aligned}
    & \dot{V} 
    = \frac{{\rm d} P(\Delta\bm x^c)}{{\rm d} (\Delta\bm x^c)^\transpose} \Delta\dot{\bm x}^c
    + \frac{1}{\eta}  \sum_{k\in \mathcal{C}}\sum\limits_{i=1}^{12}\sum\limits_{j=1}^3 \Delta\bm W_{kij}  \Delta \dot{\hat{\bm W}}_{kij}^{\transpose}
\end{aligned}
\end{equation}
Subscribing (\ref{equation:stability_analysis_3}) (\ref{equation:stability_analysis_2}) (\ref{equation:ew}) and (\ref{equation:online_updating_law}):
\begin{equation}
\begin{aligned}
    & \dot{V} 
    = (\tilde{\Delta\bm x^c})^{\transpose} \left( \hat{\bm J}^c(\bm s) \bm \nu + \bm e^c \right)
    - \frac{1}{\eta}  \sum_{k\in \mathcal{C}}\sum\limits_{i=1}^{12}\sum\limits_{j=1}^3 \Delta\bm W_{kij}  \dot{\hat{\bm W}}_{kij}^{\transpose}
    \\
    &= (\tilde{\Delta\bm x^c})^{\transpose}  \hat{\bm J}^c(\bm s) \bm \nu
    + (\tilde{\Delta\bm x^c})^{\transpose}  \bm e^c
    \\
    & - \sum_{k\in \mathcal{C}}\sum\limits_{i=1}^{12}\sum\limits_{j=1}^3 \Delta\bm W_{kij}  \bm\theta(\tilde{\bm s})  T_i \nu_i  \tilde{\Delta x_{kj}}
    \\
    &  - \frac{\gamma}{T_w}  \sum_{k\in \mathcal{C}}\sum\limits_{i=1}^{12}\sum\limits_{j=1}^3 
    \Delta\bm W_{kij} \sum_{\tau \in \mathcal{T}_w}  \bm\theta(\tilde{\bm s}(\tau)) T_i \frac{\nu_i(\tau)}{n_v(\tau)}  \frac{e_{kj}(\tau, t)}{n_v(\tau)}
\end{aligned}
\end{equation}
Subscribing (\ref{equation:ew}) and (\ref{equation:all_c}):
\begin{equation}
\begin{aligned}
    & \dot{V} 
    = (\tilde{\Delta\bm x^c})^{\transpose}  \hat{\bm J}^c(\bm s) \bm \nu
    + (\tilde{\Delta\bm x^c})^{\transpose}  \bm e^c
    - (\bm e^c)^{\transpose} \tilde{\Delta\bm x^c}
    \\
    &  - \frac{\gamma}{T_w}  \sum_{\tau \in \mathcal{T}_w} 
    \sum_{k\in \mathcal{C}}\sum\limits_{i=1}^{12}\sum\limits_{j=1}^3 
    \Delta\bm W_{kij}  \bm\theta(\tilde{\bm s}(\tau)) T_i \frac{\nu_i(\tau)}{n_v(\tau)}  \frac{e_{kj}(\tau, t)}{n_v(\tau)}
    \\
    &= (\tilde{\Delta\bm x^c})^{\transpose}  \hat{\bm J}^c(\bm s) \bm \nu
    - \frac{\gamma}{T_w} \sum_{\tau \in \mathcal{T}_w}   \frac{(\bm e^c(\tau, t))^{\transpose}}{n_v(\tau)} \frac{\bm e^c(\tau, t)}{n_v(\tau)}
\end{aligned}
\end{equation}
Since 
\begin{equation}
    \frac{\gamma}{T_w} \sum_{\tau \in \mathcal{T}_w}   \frac{(\bm e^c(\tau, t))^{\transpose}}{n_v(\tau)} \frac{\bm e^c(\tau, t)}{n_v(\tau)}
    \geq 0
\end{equation}
and (\ref{equation:stability_analysis_1}), we finally derive that $\dot{V} \leq 0$.

Since $V > 0$ and $\dot V \leq 0$, the closed-loop system is stable, and the boundedness of $V$ ensures the boundedness of task error $\Delta \bm x^c$ from (\ref{equation:V}).

Note that $\dot{V} = 0$ only if
\begin{equation}
\frac{\gamma}{T_w} \sum_{\tau \in \mathcal{T}_w}   \frac{(\bm e^c(\tau, t))^{\transpose}}{n_v(\tau)} \frac{\bm e^c(\tau, t)}{n_v(\tau)} = 0
\, \text{ and } \,
(\tilde{\Delta\bm x^c})^{\transpose}  \hat{\bm J}^c(\bm s) \bm \nu = 0 
\end{equation}

First, consider the first term, which will equal zero only if the prediction error $\bm e^c(\tau, t) = \bm 0$ for all data in the sliding window $\tau \in \mathcal{T}_w$. Such situations are very rare before the task is completed, where the approximate model should be absolutely accurate and all prediction errors should be zero.

Then, consider the second term, which is equivalent to $(\dot{\bm x}^c_{\rm ide})^{\transpose} \hat{\bm J}^c(\bm s) \bm \nu = 0$ from (\ref{equation:idea_vel}). According to (\ref{equation:stability_analysis_0}), we have $(\dot{\bm x}^c_{\rm ide})^{\transpose} \hat{\bm J}^c(\bm s) \bm \nu = 0$ only if the optimal solution of the problem (\ref{equation:local_control_objective_estimate}) is $\bm \nu = \bm 0$. While the task is not completed ($\Delta \bm x^c \neq \bm 0$), such situations may happen when the system is trapped into a local minimum point. It means that there are huge conflicts between the current task errors of different target points, so the controller based on the current estimated model thinks that any local robot movement cannot reduce the overall task error at the current configuration.
In such an underactuated system, the local minimum is theoretically inevitable, but the experimental results demonstrate that such huge conflicts happen rarely as long as the desired position vector is feasible. 

Only when these two conditions are met at the same time will $\dot{V}$ equal zero. From a practical point of view, this is almost impossible. 
As a result, it is reasonable that $\dot{V} < 0$ always holds before the task is completed. Since $V$ is positive definite, $\dot{V}$ is negative definite, and $V \rightarrow \infty$ as $\| \Delta \bm x^c \|_2 \rightarrow \infty$ from (\ref{equation:V}), the convergence of $\Delta \bm x^c \rightarrow \bm 0$ as $t \rightarrow \infty$ is ensured, following \cite{slotine1991applied}.


\section{{Additional Results}}

\subsection{{Snapshots of Full Control Process}}

{
Fig. \ref{fig:snapshots_of_a_case} provides snapshots of a case in real-world 3D experiments to better illustrate the manipulation process to readers. Please refer to our video for the full control processes of other cases.
}

\begin{figure*} [tb]
  \centering 
    \includegraphics[width=\textwidth]{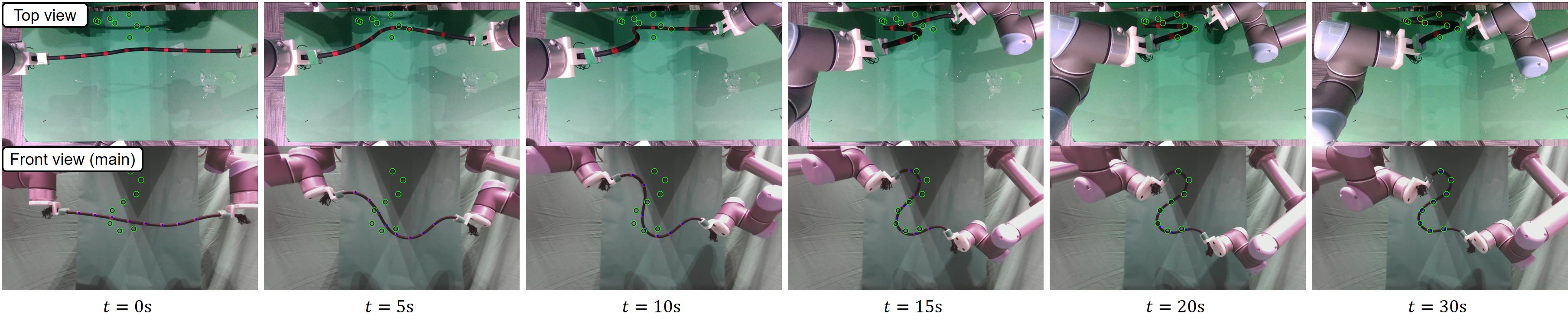} 
  \caption{{Snapshots of a case in the 3D experiments to illustrate the full shape control process. 
  }}
  \label{fig:snapshots_of_a_case}
\end{figure*}

\subsection{{Manipulating Very Short or Very Long DLOs}}

{
Generally speaking, very short DLOs are easier to model and manipulate, since their motion is usually more like rigid objects. 
Very long DLOs will easily get out of the dual-arm workspace and the camera's field of view (FOV). In addition, they may be more deformable, and their inertial effect may be greater, making the modeling and manipulation harder.
}

{
Fig. \ref{fig:10cm_2m_DLO} shows simulated shape control on a very short DLO (0.1m) and a very long DLO (2.0m). It indicates that our method can be applied to DLOs whose lengths are beyond the range of the offline trained DLOs' lengths (0.3m-1.2m). 
}

\begin{figure} [tb]
  \centering 
  \subfigure{ 
    \includegraphics[width=8.7cm]{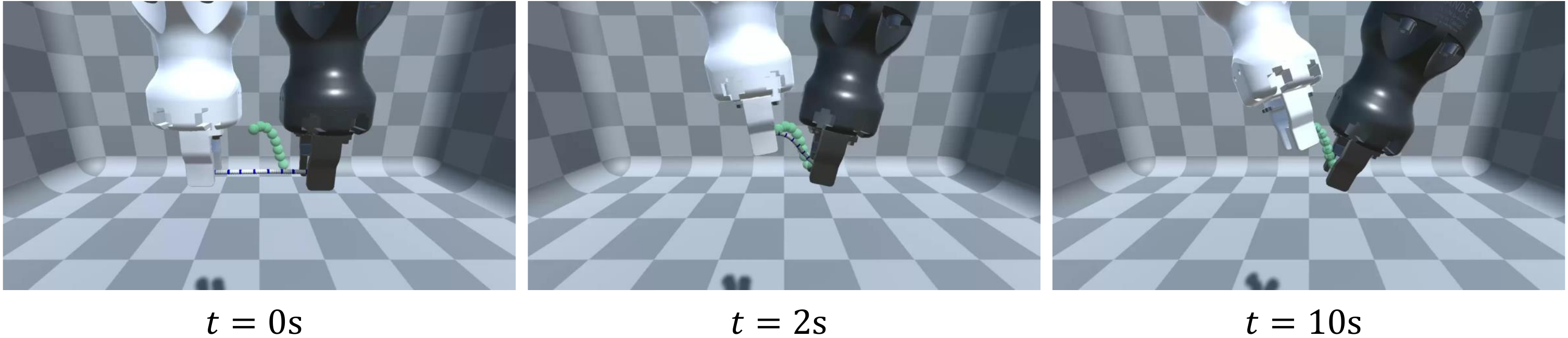} 
  } 
  \subfigure{ 
    \includegraphics[width=8.7cm]{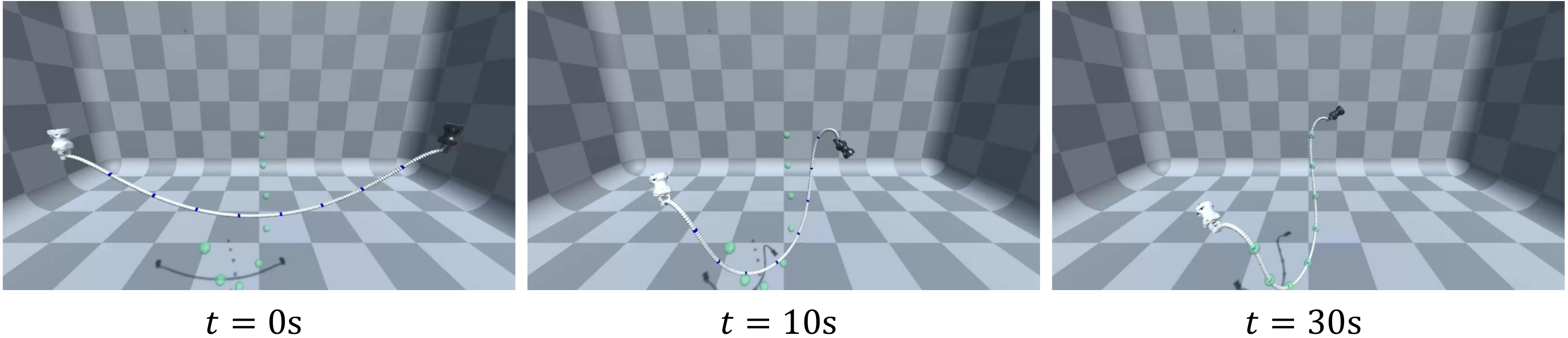}
  }
  \caption{{Shape control of very short (0.1m, top) and very long (2.0m, bottom) DLOs in the simulation.}}
  \label{fig:10cm_2m_DLO}
\end{figure}

\subsection{{Impact of Sensing Noise}}

{
We conduct simulation tests to quantitatively study the impact of the sensing error on the overall control performance. We add Gaussian noise whose distribution is $\mathcal{N}(0, \sigma^2)$ to each dimension of each feature's position, with zero mean and the variance of $\sigma^2$. Hence, the noise increases when $\sigma$ becomes larger. The feature velocity is simply calculated by $(\bm x(t) - \bm x(t-\delta t)) / \delta t$, where $\delta t$ is the time step. Note that in this way, the feature velocity noise distribution is $\mathcal{N}(0, 2(\sigma / \delta t)^2)$. In our simulation, $\delta t$ is 0.1s, so the standard deviation of the velocity noise is $10\sqrt{2}$ times that of the position noise. Other settings are the same as those in Section \ref{subsection:sim_shape_control}.
}

\begin{figure} [tb]
  \centering 
  \subfigure{ 
    \includegraphics[width=4.3cm]{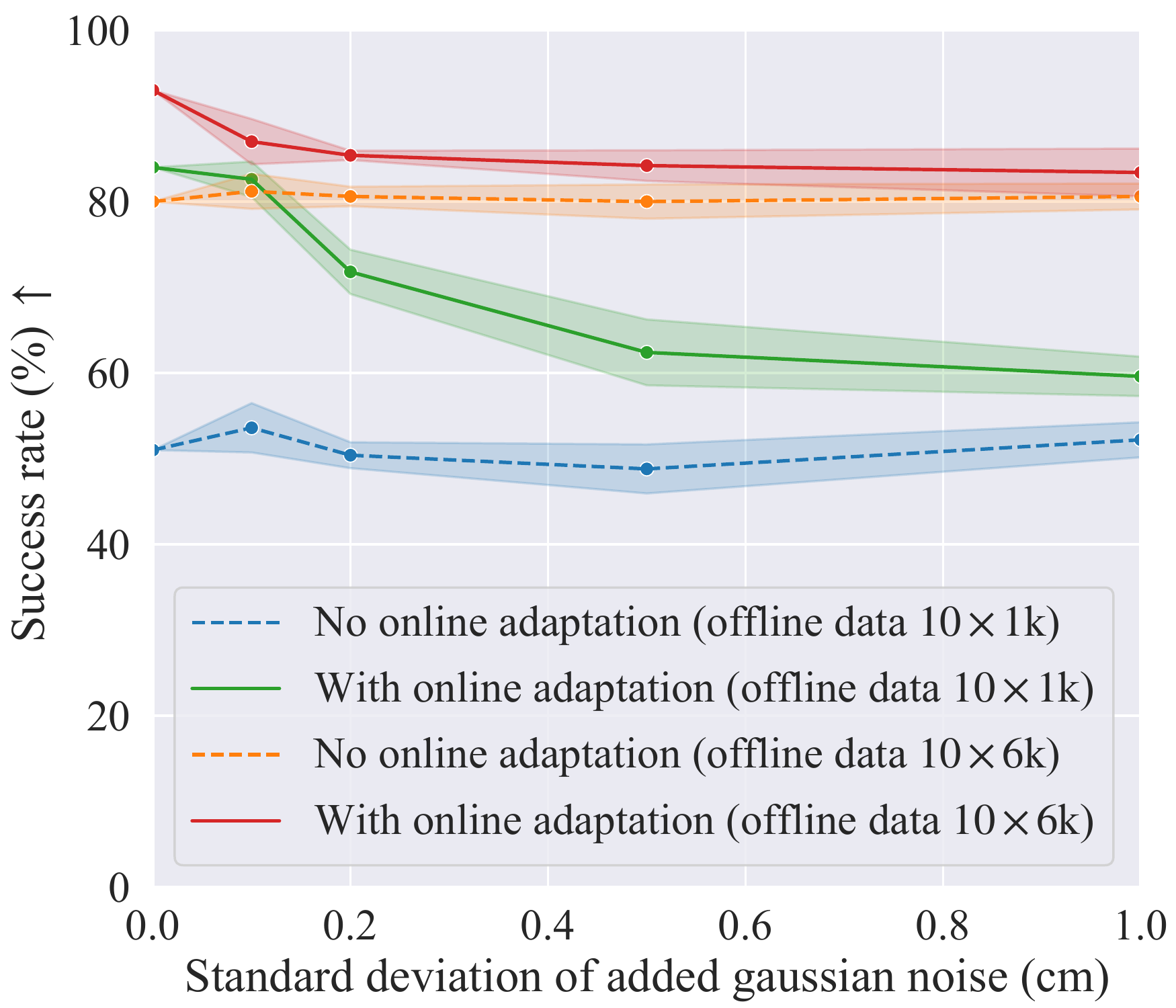} 
  } 
  \hspace{-5mm}
  \subfigure{ 
    \includegraphics[width=4.3cm]{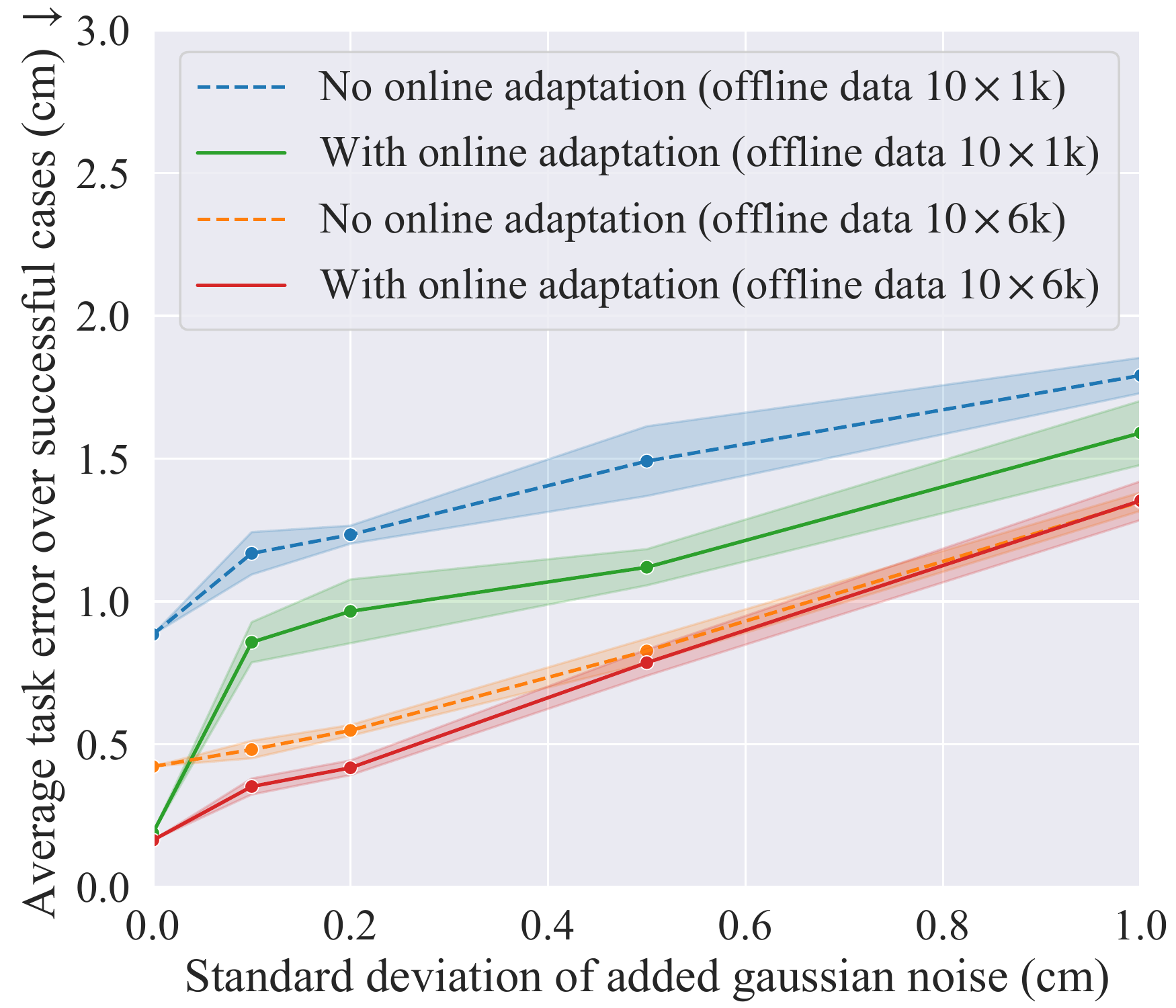}
  }
  \caption{{Impact of the sensing noise on the overall control performance. The position noise leads to velocity noise, affecting the online model adaptation. The test is repeated 5 times, and the mean and standard derivation are plotted.}}
  \label{fig:sensing_noise}
\end{figure}

The test results are shown in Fig. \ref{fig:sensing_noise}. In our method, the Jacobian calculation and feedback control law require the feature positions, and the online model adaptation is driven by both the feature positions and velocities. 
As the added noise increases, it is seen that:
1) In general, our method is robust to the noise, where the success rate changes little.
2) The average task error over successful cases increases, since the position noise causes oscillation around the desired shapes. 
3) The performance of the online model adaptation drops, since small position noise may cause very large velocity noise; to alleviate the issue, a front-end filter is preferred in noisy scenarios for more reliable velocity perception.

} 

\bibliographystyle{IEEEtran}
\bibliography{IEEEabrv, ref}

\begin{thebibliography}{10}
\providecommand{\url}[1]{#1}
\csname url@samestyle\endcsname
\providecommand{\newblock}{\relax}
\providecommand{\bibinfo}[2]{#2}
\providecommand{\BIBentrySTDinterwordspacing}{\spaceskip=0pt\relax}
\providecommand{\BIBentryALTinterwordstretchfactor}{4}
\providecommand{\BIBentryALTinterwordspacing}{\spaceskip=\fontdimen2\font plus
\BIBentryALTinterwordstretchfactor\fontdimen3\font minus
  \fontdimen4\font\relax}
\providecommand{\BIBforeignlanguage}[2]{{%
\expandafter\ifx\csname l@#1\endcsname\relax
\typeout{** WARNING: IEEEtran.bst: No hyphenation pattern has been}%
\typeout{** loaded for the language `#1'. Using the pattern for}%
\typeout{** the default language instead.}%
\else
\language=\csname l@#1\endcsname
\fi
#2}}
\providecommand{\BIBdecl}{\relax}
\BIBdecl

\bibitem{jose2018robotic}
J.~Sanchez, J.-A. Corrales, B.-C. Bouzgarrou, and Y.~Mezouar, ``Robotic
  manipulation and sensing of deformable objects in domestic and industrial
  applications: a survey,'' \emph{The International Journal of Robotics
  Research}, vol.~37, no.~7, pp. 688--716, 2018.

\bibitem{yin2021modeling}
H.~Yin, A.~Varava, and D.~Kragic, ``Modeling, learning, perception, and control
  methods for deformable object manipulation,'' \emph{Science Robotics},
  vol.~6, no.~54, 2021.

\bibitem{nadon2018multi}
F.~Nadon, A.~J. Valencia, and P.~Payeur, ``Multi-modal sensing and robotic
  manipulation of non-rigid objects: A survey,'' \emph{Robotics}, vol.~7,
  no.~4, p.~74, 2018.

\bibitem{herguedas2019survey}
R.~Herguedas, G.~López-Nicolás, R.~Aragüés, and C.~Sagüés, ``Survey on
  multi-robot manipulation of deformable objects,'' in \emph{2019 24th IEEE
  International Conference on Emerging Technologies and Factory Automation
  (ETFA)}, 2019, pp. 977--984.

\bibitem{8460694}
X.~{Li}, X.~{Su}, Y.~{Gao}, and Y.~{Liu}, ``Vision-based robotic grasping and
  manipulation of usb wires,'' in \emph{2018 IEEE International Conference on
  Robotics and Automation (ICRA)}, 2018, pp. 3482--3487.

\bibitem{jin2021trajectory}
S.~Jin, D.~Romeres, A.~Ragunathan, D.~K. Jha, and M.~Tomizuka, ``Trajectory
  optimization for manipulation of deformable objects: Assembly of belt drive
  units,'' in \emph{2021 IEEE International Conference on Robotics and
  Automation (ICRA)}, 2021, pp. 10\,002--10\,008.

\bibitem{cao2020sewing}
L.~{Cao}, X.~{Li}, P.~T. {Phan}, A.~M.~H. {Tiong}, H.~L. {Kaan}, J.~{Liu},
  W.~{Lai}, Y.~{Huang}, H.~M. {Le}, M.~{Miyasaka}, K.~Y. {Ho}, P.~W.~Y. {Chiu},
  and S.~J. {Phee}, ``Sewing up the wounds: A robotic suturing system for
  flexible endoscopy,'' \emph{IEEE Robotics Automation Magazine}, vol.~27,
  no.~3, pp. 45--54, 2020.

\bibitem{rita2021reform}
R.~Laezza, R.~Gieselmann, F.~T. Pokorny, and Y.~Karayiannidis, ``Reform: A
  robot learning sandbox for deformable linear object manipulation,'' in
  \emph{2021 IEEE International Conference on Robotics and Automation (ICRA)},
  2021, pp. 4717--4723.

\bibitem{wakamatsu2006knotting}
H.~Wakamatsu, E.~Arai, and S.~Hirai, ``Knotting/unknotting manipulation of
  deformable linear objects,'' \emph{The International Journal of Robotics
  Research}, vol.~25, no.~4, pp. 371--395, 2006.

\bibitem{saha2007manipulation}
M.~Saha and P.~Isto, ``Manipulation planning for deformable linear objects,''
  \emph{IEEE Transactions on Robotics}, vol.~23, no.~6, pp. 1141--1150, 2007.

\bibitem{mcconachie2020manipulating}
D.~McConachie, A.~Dobson, M.~Ruan, and D.~Berenson, ``Manipulating deformable
  objects by interleaving prediction, planning, and control,'' \emph{The
  International Journal of Robotics Research}, vol.~39, no.~8, pp. 957--982,
  2020.

\bibitem{mitrano2021learning}
P.~Mitrano, D.~McConachie, and D.~Berenson, ``Learning where to trust
  unreliable models in an unstructured world for deformable object
  manipulation,'' \emph{Science Robotics}, vol.~6, no.~54, p. eabd8170, 2021.

\bibitem{she2021cable}
Y.~She, S.~Wang, S.~Dong, N.~Sunil, A.~Rodriguez, and E.~Adelson, ``Cable
  manipulation with a tactile-reactive gripper,'' \emph{The International
  Journal of Robotics Research}, vol.~40, no. 12-14, pp. 1385--1401, 2021.

\bibitem{zhu2021challenges}
J.~Zhu, A.~Cherubini, C.~Dune, D.~Navarro-Alarcon, F.~Alambeigi, D.~Berenson,
  F.~Ficuciello, K.~Harada, J.~Kober, X.~LI, J.~Pan, W.~Yuan, and M.~Gienger,
  ``Challenges and outlook in robotic manipulation of deformable objects,''
  \emph{IEEE Robotics and Automation Magazine}, pp. 2--12, 2022.

\bibitem{arriola2020modeling}
V.~E. Arriola-Rios, P.~Guler, F.~Ficuciello, D.~Kragic, B.~Siciliano, and J.~L.
  Wyatt, ``Modeling of deformable objects for robotic manipulation: A tutorial
  and review,'' \emph{Frontiers in Robotics and AI}, vol.~7, p.~82, 2020.

\bibitem{yan2020learning}
W.~Yan, A.~Vangipuram, P.~Abbeel, and L.~Pinto, ``Learning predictive
  representations for deformable objects using contrastive estimation,'' in
  \emph{4th Conference on Robot Learning (CoRL)}, 2020.

\bibitem{wenbo2021deformable}
W.~Zhang, K.~Schmeckpeper, P.~Chaudhari, and K.~Daniilidis, ``Deformable linear
  object prediction using locally linear latent dynamics,'' in \emph{2021 IEEE
  International Conference on Robotics and Automation (ICRA)}, 2021, pp.
  13\,503--13\,509.

\bibitem{yan_self-supervised_2020}
M.~{Yan}, Y.~{Zhu}, N.~{Jin}, and J.~{Bohg}, ``Self-supervised learning of
  state estimation for manipulating deformable linear objects,'' \emph{IEEE
  Robotics and Automation Letters}, vol.~5, no.~2, pp. 2372--2379, 2020.

\bibitem{yang2021learning}
Y.~Yang, J.~A. Stork, and T.~Stoyanov, ``Learning to propagate interaction
  effects for modeling deformable linear objects dynamics,'' in \emph{2021 IEEE
  International Conference on Robotics and Automation (ICRA)}, 2021, pp.
  1950--1957.

\bibitem{lee2022sample}
R.~Lee, M.~Hamaya, T.~Murooka, Y.~Ijiri, and P.~Corke, ``Sample-efficient
  learning of deformable linear object manipulation in the real world through
  self-supervision,'' \emph{IEEE Robotics and Automation Letters}, vol.~7,
  no.~1, pp. 573--580, 2022.

\bibitem{lin2020softgym}
X.~Lin, Y.~Wang, J.~Olkin, and D.~Held, ``Softgym: Benchmarking deep
  reinforcement learning for deformable object manipulation,'' in \emph{4th
  Conference on Robot Learning (CoRL)}, 2020.

\bibitem{rita2021learning}
R.~Laezza and Y.~Karayiannidis, ``Learning shape control of elastoplastic
  deformable linear objects,'' in \emph{2021 IEEE International Conference on
  Robotics and Automation (ICRA)}, 2021, pp. 4438--4444.

\bibitem{david2013modelfree}
D.~Navarro-Alarcón, Y.-H. Liu, J.~G. Romero, and P.~Li, ``Model-free visually
  servoed deformation control of elastic objects by robot manipulators,''
  \emph{IEEE Transactions on Robotics}, vol.~29, no.~6, pp. 1457--1468, 2013.

\bibitem{navarro2016Automatic}
D.~{Navarro-Alarcon}, H.~M. {Yip}, Z.~{Wang}, Y.~{Liu}, F.~{Zhong}, T.~{Zhang},
  and P.~{Li}, ``Automatic 3-d manipulation of soft objects by robotic arms
  with an adaptive deformation model,'' \emph{IEEE Transactions on Robotics},
  vol.~32, no.~2, pp. 429--441, 2016.

\bibitem{zhu_dual-arm_2018}
J.~{Zhu}, B.~{Navarro}, P.~{Fraisse}, A.~{Crosnier}, and A.~{Cherubini},
  ``Dual-arm robotic manipulation of flexible cables,'' in \emph{2018 IEEE/RSJ
  International Conference on Intelligent Robots and Systems (IROS)}, 2018, pp.
  479--484.

\bibitem{jin2019robust}
S.~Jin, C.~Wang, and M.~Tomizuka, ``Robust deformation model approximation for
  robotic cable manipulation,'' in \emph{2019 IEEE/RSJ International Conference
  on Intelligent Robots and Systems (IROS)}, 2019, pp. 6586--6593.

\bibitem{lagneau_automatic_2020}
R.~{Lagneau}, A.~{Krupa}, and M.~{Marchal}, ``Automatic shape control of
  deformable wires based on model-free visual servoing,'' \emph{IEEE Robotics
  and Automation Letters}, vol.~5, no.~4, pp. 5252--5259, 2020.

\bibitem{zhu2021vision}
J.~Zhu, D.~Navarro-Alarcon, R.~Passama, and A.~Cherubini, ``Vision-based
  manipulation of deformable and rigid objects using subspace projections of 2d
  contours,'' \emph{Robotics and Autonomous Systems}, vol. 142, p. 103798,
  2021.

\bibitem{yu2022shape}
M.~Yu, H.~Zhong, and X.~Li, ``Shape control of deformable linear objects with
  offline and online learning of local linear deformation models,'' in
  \emph{2022 International Conference on Robotics and Automation (ICRA)}, 2022,
  pp. 1337--1343.

\bibitem{bretl2014quasi}
T.~Bretl and Z.~McCarthy, ``Quasi-static manipulation of a kirchhoff elastic
  rod based on a geometric analysis of equilibrium configurations,'' \emph{The
  International Journal of Robotics Research}, vol.~33, no.~1, pp. 48--68,
  2014.

\bibitem{roussel2015manipulation}
O.~Roussel, A.~Borum, M.~Taïx, and T.~Bretl, ``Manipulation planning with
  contacts for an extensible elastic rod by sampling on the submanifold of
  static equilibrium configurations,'' in \emph{2015 IEEE International
  Conference on Robotics and Automation (ICRA)}, 2015, pp. 3116--3121.

\bibitem{lv2022dynamic}
N.~Lv, J.~Liu, and Y.~Jia, ``Dynamic modeling and control of deformable linear
  objects for single-arm and dual-arm robot manipulations,'' \emph{IEEE
  Transactions on Robotics}, pp. 1--13, 2022.

\bibitem{duenser2018interactive}
S.~Duenser, J.~M. Bern, R.~Poranne, and S.~Coros, ``Interactive robotic
  manipulation of elastic objects,'' in \emph{2018 IEEE/RSJ International
  Conference on Intelligent Robots and Systems (IROS)}, 2018, pp. 3476--3481.

\bibitem{koessler2021anefficient}
A.~Koessler, N.~R. Filella, B.~Bouzgarrou, L.~Lequièvre, and J.-A.~C. Ramon,
  ``An efficient approach to closed-loop shape control of deformable objects
  using finite element models,'' in \emph{2021 IEEE International Conference on
  Robotics and Automation (ICRA)}, 2021, pp. 1637--1643.

\bibitem{rambow2012autonomous}
M.~Rambow, T.~Schauß, M.~Buss, and S.~Hirche, ``Autonomous manipulation of
  deformable objects based on teleoperated demonstrations,'' in \emph{2012
  IEEE/RSJ International Conference on Intelligent Robots and Systems (IROS)},
  2012, pp. 2809--2814.

\bibitem{nair2017combining}
A.~Nair, D.~Chen, P.~Agrawal, P.~Isola, P.~Abbeel, J.~Malik, and S.~Levine,
  ``Combining self-supervised learning and imitation for vision-based rope
  manipulation,'' in \emph{2017 IEEE International Conference on Robotics and
  Automation (ICRA)}, 2017, pp. 2146--2153.

\bibitem{tang2018aframework}
T.~Tang, C.~Wang, and M.~Tomizuka, ``A framework for manipulating deformable
  linear objects by coherent point drift,'' \emph{IEEE Robotics and Automation
  Letters}, vol.~3, no.~4, pp. 3426--3433, 2018.

\bibitem{wang2022offline}
C.~Wang, Y.~Zhang, X.~Zhang, Z.~Wu, X.~Zhu, S.~Jin, T.~Tang, and M.~Tomizuka,
  ``Offline-online learning of deformation model for cable manipulation with
  graph neural networks,'' \emph{IEEE Robotics and Automation Letters}, vol.~7,
  no.~2, pp. 5544--5551, 2022.

\bibitem{berenson2013manipulation}
D.~Berenson, ``Manipulation of deformable objects without modeling and
  simulating deformation,'' in \emph{2013 IEEE/RSJ International Conference on
  Intelligent Robots and Systems (IROS)}, 2013, pp. 4525--4532.

\bibitem{girshick2015fast}
R.~Girshick, ``Fast {R-CNN},'' in \emph{Proceedings of the IEEE International
  Conference on Computer Vision (ICCV)}, 2015.

\bibitem{kingma2014adam}
D.~P. Kingma and J.~Ba, ``{Adam}: A method for stochastic optimization,'' in
  \emph{Proceedings of the 3rd International Conference on Learning
  Representations (ICLR)}, 2014.

\bibitem{yu2011advantages}
H.~Yu, T.~Xie, S.~Paszczynski, and B.~M. Wilamowski, ``Advantages of radial
  basis function networks for dynamic system design,'' \emph{IEEE Transactions
  on Industrial Electronics}, vol.~58, no.~12, pp. 5438--5450, 2011.

\bibitem{williams2017information}
G.~Williams, N.~Wagener, B.~Goldfain, P.~Drews, J.~M. Rehg, B.~Boots, and E.~A.
  Theodorou, ``Information theoretic mpc for model-based reinforcement
  learning,'' in \emph{2017 IEEE International Conference on Robotics and
  Automation (ICRA)}, 2017, pp. 1714--1721.

\bibitem{ruan2018accounting}
M.~Ruan, D.~McConachie, and D.~Berenson, ``Accounting for directional rigidity
  and constraints in control for manipulation of deformable objects without
  physical simulation,'' in \emph{2018 IEEE/RSJ International Conference on
  Intelligent Robots and Systems (IROS)}, 2018, pp. 512--519.

\bibitem{chiaverini1994review}
S.~Chiaverini, B.~Siciliano, and O.~Egeland, ``Review of the damped
  least-squares inverse kinematics with experiments on an industrial robot
  manipulator,'' \emph{IEEE Transactions on Control Systems Technology},
  vol.~2, no.~2, pp. 123--134, 1994.

\bibitem{obi}
\BIBentryALTinterwordspacing
V.~M. Studio. (2019) Obi - {Unified} particle physics for {Unity 3D}. [Online].
  Available: \url{http://obi.virtualmethodstudio.com/}
\BIBentrySTDinterwordspacing

\bibitem{unity}
\BIBentryALTinterwordspacing
U.~Technologies. (2021) Unity real-time development platform. [Online].
  Available: \url{https://unity.com/}
\BIBentrySTDinterwordspacing

\bibitem{juliani2018unity}
A.~Juliani, V.-P. Berges, E.~Teng, A.~Cohen, J.~Harper, C.~Elion, C.~Goy,
  Y.~Gao, H.~Henry, M.~Mattar \emph{et~al.}, ``Unity: A general platform for
  intelligent agents,'' \emph{arXiv preprint arXiv:1809.02627}, 2018.

\bibitem{quigley2009ros}
M.~Quigley, K.~Conley, B.~Gerkey, J.~Faust, T.~Foote, J.~Leibs, R.~Wheeler,
  A.~Y. Ng \emph{et~al.}, ``{ROS}: an open-source robot operating system,'' in
  \emph{ICRA Workshop on Open Source Software}, vol.~3, no. 3.2.\hskip 1em plus
  0.5em minus 0.4em\relax Kobe, Japan, 2009, p.~5.

\bibitem{tang2018track}
T.~Tang and M.~Tomizuka, ``Track deformable objects from point clouds with
  structure preserved registration,'' \emph{The International Journal of
  Robotics Research}, 2019.

\bibitem{chi2019occlusion}
C.~Chi and D.~Berenson, ``Occlusion-robust deformable object tracking without
  physics simulation,'' in \emph{2019 IEEE/RSJ International Conference on
  Intelligent Robots and Systems (IROS)}, 2019, pp. 6443--6450.

\bibitem{wang2021tracking}
Y.~Wang, D.~McConachie, and D.~Berenson, ``Tracking partially-occluded
  deformable objects while enforcing geometric constraints,'' in \emph{2021
  IEEE International Conference on Robotics and Automation (ICRA)}, 2021, pp.
  14\,199--14\,205.

\bibitem{slotine1991applied}
J.-J.~E. Slotine, W.~Li \emph{et~al.}, \emph{Applied nonlinear control}.\hskip
  1em plus 0.5em minus 0.4em\relax Prentice hall Englewood Cliffs, NJ, 1991,
  vol. 199, no.~1.

\end{thebibliography}

\vfill

\end{document}